\documentclass[10pt,twocolumn,letterpaper]{article}

\usepackage[table,x11names,dvipsnames]{xcolor}
\usepackage{amsmath}
\usepackage{amssymb}
\usepackage{amsthm}
\usepackage{makecell}
\usepackage{color,soul}
\usepackage{booktabs}
\usepackage{multirow}
\usepackage{caption}
\usepackage{subcaption}
\usepackage{graphicx}
\usepackage{adjustbox}
\usepackage{dblfloatfix}
\usepackage{float}
\usepackage{slashbox}
\newtheorem{definition}{Definition}

\newtheorem{prop}{Proposition}

\DeclareMathOperator*{\argmin}{arg\,min}
\usepackage[pagenumbers]{cvpr} % To force page numbers, e.g. for an arXiv version

\usepackage[dvipsnames]{xcolor}

\definecolor{cvprblue}{rgb}{0.21,0.49,0.74}
\usepackage[pagebackref,breaklinks,colorlinks,citecolor=cvprblue]{hyperref}

\def\paperID{14087} 
\def\confName{CVPR\xspace}
\def\confYear{2024\xspace}

\title{Geometry Depth Consistency in RGBD Relative Pose Estimation}

\author{Sourav Kumar \hspace{6em} Chiang-Heng Chien \hspace{6em} Benjamin Kimia \\
{\tt\small sourav\_kumar@brown.edu} \hspace{2em} {\tt\small chiang-heng\_chien@brown.edu} \hspace{2em} {\tt\small benjamin\_kimia@brown.edu} \vspace{0.5em} \\ School of Engineering, Brown University
}

\begin{document}
\maketitle
\begin{abstract}
Relative pose estimation for RGBD cameras is crucial in a number of applications. Previous approaches either rely on the RGB aspect of the images to estimate pose thus not fully making use of depth in the estimation process or estimate pose from the 3D cloud of points that each image produces, thus not making full use of RGB information. This paper shows that if one pair of correspondences is hypothesized from the RGB-based ranked-ordered correspondence list, then the space of remaining correspondences is restricted to corresponding pairs of curves nested around the hypothesized correspondence, implicitly capturing depth consistency. This simple {\em Geometric Depth Constraint (GDC) } significantly reduces potential matches. In effect this becomes a filter on possible correspondences that helps reduce the number of outliers and thus expedites RANSAC significantly. As such, the same budget of time allows for more RANSAC iterations and therefore additional robustness and a significant speedup. In addition, the paper proposed a Nested RANSAC approach that also speeds up the process, as shown through experiments on TUM, ICL-NUIM, and  RGBD Scenes v2 datasets. 
\end{abstract} 

\section{Introduction}
\label{sec:intro}
Relative pose estimation from image pairs is a fundamental and ubiquitous problem for many computer vision tasks, \emph{e.g.} visual odometry~\cite{fontan2020information, yuan2021rgb, zhao2023robust}, SLAM~\cite{fontan2023sid, zhu2022nice}, 3D scene reconstruction~\cite{zhang2023go} and completion~\cite{yang2019extreme, wang2023robust}, \emph{etc}. A robust estimation process typically follows a three-step paradigm~\cite{el2023self}, namely, ({\em i}) detect and extract features, \emph{e.g.}, SIFT~\cite{lowe2004distinctive} or SuperPoint~\cite{detone2018superpoint}; ({\em ii}) measure pairwise feature similarity and form a rank-ordered list of potential matches; 
({\em iii}) apply RANSAC by selecting a certain number of matches from the top $M$ rank-ordered list that is large enough to support the formation of hypotheses but small enough to have a small rate of outliers, \emph{e.g.}, $M$ = 150~\cite{fabbri2022trifocal}, or a ratio of the number of matches such as 0.2~\cite{moulon2017openmvg} and 1 in~\cite{snavely2008modeling, OpenSfM} (taking all matches). The selected matches are used to calculate a camera pose as a competing hypothesis, and iterate $N$ loops to achieve a certain level of success $p$. The output is a hypothesis approximately consistent with inliers which is a comparably large subset of all the matches.\\
\begin{figure}[t]
    \centering
    \includegraphics[width=0.4\textwidth]{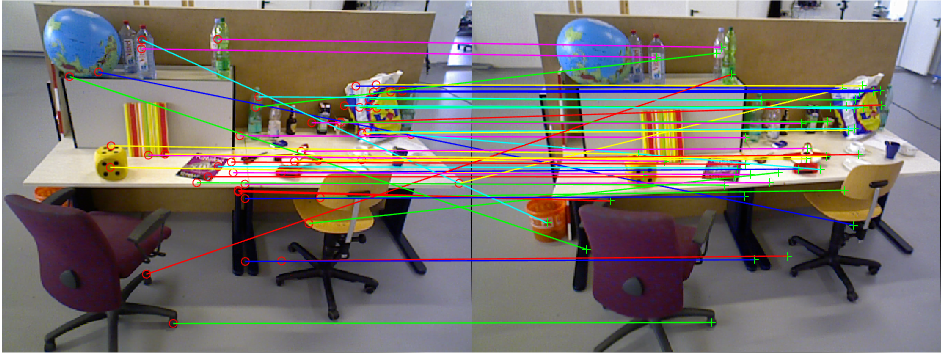}\\
    \includegraphics[width=0.2\textwidth]{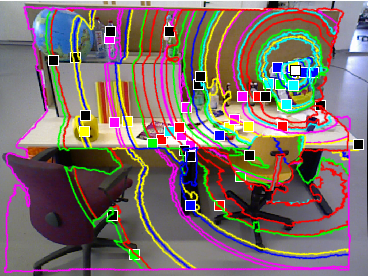}
    \includegraphics[width=0.2\textwidth]{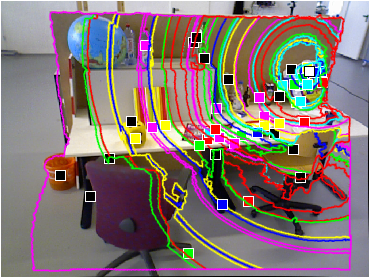}
    \caption{(Top) 50 potential matches selected from a rank-ordered list of correspondences between a pair of RGBD images. (Bottom) A pair of correspondence which is manually determined to be veridical is selected (white square tokens). Each remaining correspondence is probed as to whether the pair falls on corresponding curves using the proposed geometric depth consistency constraint. Those potential matches that fail this test are shown in black tokens and excluded as nonviable correspondences. }
    \label{fig:radial_maps_rgbd1_12}
\end{figure}
\indent While the research community has witnessed great performances in visual odometry (VO), SLAM, or structure-from-motion (SfM) pipelines based on this paradigm, the estimation accuracy significantly drops when potential matches contain a very large fraction of outliers, \emph{e.g.}, $>90\%$, for situations where image pairs experience less overlap~\cite{sun2022global}, blurry images from drastic camera motion~\cite{charatan2022benchmarking}, repetitive textures~\cite{ge2022unsupervised}, \emph{etc.} With a sufficient number of RANSAC iterations, accurate camera pose is expected to be estimated. However, existing methods typically set a maximal RANSAC iterations as efficiency prioritizes over accuracy, \emph{e.g.}, $N_{max}$ = 300~\cite{mur2017orb}, 320~\cite{moulon2017openmvg}, 1000~\cite{OpenSfM}, 8000~\cite{snavely2008modeling}, or 10000~\cite{schonberger2016structure}. Limitation on a sufficient number of RANSAC iterations pose a high risk of giving credible, robust pose estimations, especially for very high outlier ratio scenarios.

To address this problem, some approaches~\cite{sarlin2020superglue, sun2021loftr} focus on the early stage of the paradigm which gives promising dense correspondences from a learned network before the RANSAC starts, aiming to reduce the overall outlier ratio. However, these methods work entirely in the 2D image domain, ignoring the underlying 3D geometry of the scene, which can be easily acquired from RGBD cameras or learned depths~\cite{barath2022relative}. The detachment from 3D geometry leads to poor performance in large view point changes~\cite{el2023self}. Leveraging 3D geometry, \emph{e.g.}, surface normal~\cite{yang2019extreme}, curvature~\cite{wang2023guiding}, \emph{etc.} as a cue has been demonstrated to be beneficial in guiding feature matching as well as pruning out outliers under a RANSAC loop. Methods such as~\cite{zhang2019learning} infer the probabilities of correspondences being inliers with an order-aware network. Other works improve the accuracy of correspondence pruning by applying motion coherence constraints using local-to-global consensus learning procedure~\cite{liu2021learnable, ma2019locality}. These deep learning based methods are however limited to learning from carefully captured videos that can already be constructed using standard algorithms.

\indent This paper proposes an approach where efficiency and accuracy can both be achieved using two ideas: \emph{(i)} Geometric Depth Consistency (GDC), and \emph{(ii)} Nested RANSAC.
% . favored under a RANSAC scheme using a RGBD camera using the proposed two RANSAC approaches: \emph{(i)} \textbf{filtered RANSAC} which leverages geometric depth consistency constraint on point correspondences as a filter to avoid computing camera pose hypothesis from outliers, and \emph{(ii)} \textbf{nested RANSAC} which increases the probability of picking point features with high likelihoods of being inliers for computing camera pose hypothesis. While each method contributes to some degree of time savings, a combination of both methods provides an order of magnitude speedup for high outlier ratio RGBD image pairs, preserving accurate pose estimation at the same time. The proposed technique can be easily deployed to any VO, SLAM, or SfM frameworks with depths prior from RGBD cameras or learned depths~\cite{guizilini2023towards}.

% The maximum number of RANSAC iterations is set to reach a confidence level, say $p$ = 0.95~\cite{snavely2008modeling} or $p$ = 0.99~\cite{mur2017orb, OpenSfM, schonberger2016structure}, but below a maximum of RANSAC iterations for efficiency concern, \emph{e.g.}, $N_{max}$ = 300~\cite{mur2017orb}, 320~\cite{moulon2017openmvg}, 1000~\cite{OpenSfM}, 8000~\cite{snavely2008modeling}, or 10000~\cite{schonberger2016structure}.

\section{Geometric Depth Consistency} \label{section4}
\label{sec:formulation}
This section shows that the knowledge of one veridical correspondence in a pair of RGBD images significantly constrains the set of potential correspondences. Consider two RGBD cameras with unknown relative pose $(\mathcal{R},\mathcal{T})$, where $\mathcal{R}$  is the rotation matrix and $\mathcal{T}$ is the translation vector. Consider an RGBD image point $\gamma_i=(\xi_i,\eta_i,1)^T$ with depth $\rho_i$ in the image of camera one that is in correspondence with an RGBD point $\overline{\gamma}_i=(\overline{\xi}_i,\overline{\eta}_i,1)^T$  with depth $\overline{\rho}_i$ in the image of camera two. Let $\Gamma_i=\rho_i \gamma_i$ and $\overline{\Gamma}_i=\overline{\rho}_i\overline{\gamma}_i$
% \begin{equation} 
%     \label{eq:1}
%     \Gamma_i=\rho_i \gamma_i \; \text{,}\quad \overline{\Gamma}_i=\overline{\rho}_i\overline{\gamma}_i,
% \end{equation} 
be the corresponding 3D points in each camera, respectively.
%be the expressions of the common 3D point: one to the image points in cameras one and two, respectively. 
The question is whether the veridical corresponding pair $(\gamma_i,\overline{\gamma}_i)$, contains the set of correspondences, {\em i.e.,} whether given a point $\gamma_j$ in image one the locus of the corresponding point $\overline{\gamma}_j$ in image two is constrained in anyway? It is clear that without the knowledge of this veridical correspondence and with unknown pose, the space of possible correspondences for any given $\gamma_j$ is the entire image. 
A veridical correspondence implies that
\begin{equation} 
\label{eq: 6}
\overline{\Gamma}_i=\mathcal{R}\Gamma_i+\mathcal{T}, \quad \text{or} \quad \overline{\rho}_i\overline{\gamma}_i=\mathcal{R}\rho_i\gamma_i+T.
\end{equation}

In the RGB case, two of the scalar equations are used to eliminate the unknown depths, leaving a single scale equation, which is generally known as the epipolar constraint. There is also metric ambiguity in recovering the size of $\mathcal{T}$ so there are only five unknowns in $(\mathcal{R},\mathcal{T})$, which generally require five correspondences to solve for pose. In this case, the knowledge of a single correspondence does not constrain the remaining correspondences.

%This vector equation representing three scalar equations constraining the six unknowns of pose $(\mathcal{R},\mathcal{T})$ when dealing with an RGB image, the six unknown parameters of pose $(\mathcal{R},\mathcal{T})$ are augmented with the two unknown depths $(\rho_i,\overline{\rho}_i)$. Eliminating the unknown depths from this equations leaves a single scalar equation (the epipolar constraint) constraining $(\mathcal{R},\mathcal{T})$, which due to the metric ambiguity has five unknowns, three for $\mathcal{R}$ and two for a normalized $\mathcal{T}$. Thus, five correspondences are required to solve for $(\mathcal{R},\mathcal{T})$. \\
\begin{figure}[b]
  \centering
  \includegraphics[width=.4\textwidth]{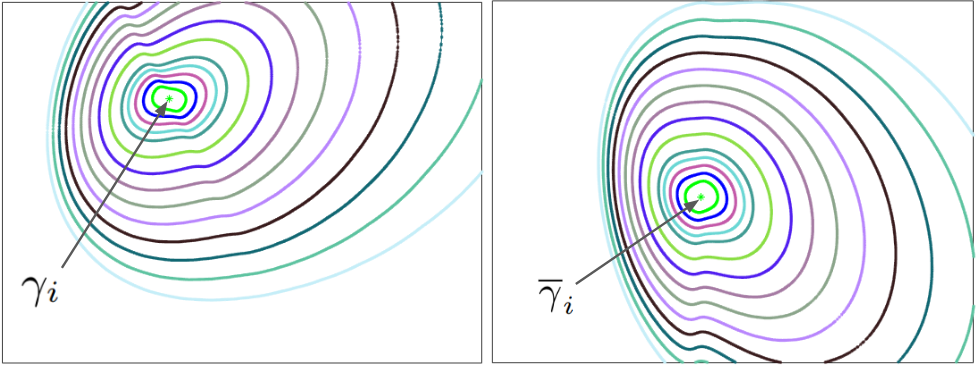}
  \caption{A veridical correspondence $(\gamma_i,\overline{\gamma}_i)$ partitions the space of correspondences $(\gamma_j,\overline{\gamma}_j)$ into a nested set of curves (identified by a common color) so that if $\gamma_j$ falls on a curve in image one, $\overline{\gamma}_j$ must fall on the corresponding curve in image two. }
  \label{fig:fig 2}
\end{figure}
The situation with RGBD data is different since the two depths $\rho_i$ and $\overline{\rho}_i$ are known, so that three equations constrain $(\mathcal{R},\mathcal{T})$. This could potentially imply a restriction on any other potential correspondence $(\gamma_j,\overline{\gamma}_j)$ with depths $\rho_j$ and $\overline{\rho}_j$ respectively. Since the two equations %thus requiring only three correspondences to solve the six unknowns of $(\mathcal{R},\mathcal{T})$. However, not all pairs of correspondences $(\gamma_i,\overline{\gamma}_i)$ and $(\gamma_j,\overline{\gamma}_j)$ with depths $(\rho_i,\overline{\rho}_i)$ and $(\rho_j,\overline{\rho}_j)$, respectively, give rise to a solution for pose. Rather, the two equations %Denoting $\Gamma_i=\rho_i\gamma_i$, $\overline{\Gamma}_i=\overline{\rho}_i\overline{\gamma}_i$, $\Gamma_j=\rho_j\gamma_j$, $\overline{\Gamma}_j=\overline{\rho}_j\overline{\gamma}_j$, The two correspondences give 
% \begin{equation} \label{eq:11}
% \left\{
% \begin{aligned}
%     \overline{\Gamma}_i &= \mathcal{R}\Gamma_i + \mathcal{T}, \\
%     \overline{\Gamma}_j &= \mathcal{R}\Gamma_j + \mathcal{T}.
% \end{aligned}
% \right.
% \end{equation}
\begin{equation} \label{eq:11}
    \overline{\Gamma}_i = \mathcal{R}\Gamma_i + \mathcal{T}, \quad
    \overline{\Gamma}_j = \mathcal{R}\Gamma_j + \mathcal{T}.
\end{equation}
must hold. Indeed, eliminating $\mathcal{T}$ by subtracting the two equations in equation~(\ref{eq:11}) gives\begin{equation} \label{eq:12}
     \overline{\Gamma}_i-\overline{\Gamma}_j=\mathcal{R}(\Gamma_i-\Gamma_j).
\end{equation} Eliminating $\mathcal{R}$  by a dot product gives 
\begin{equation}
\label{eq:1}
    (\overline{\Gamma}_i-\overline{\Gamma}_j)^T(\overline{\Gamma}_i-\overline{\Gamma}_j) = (\Gamma_i-\Gamma_j)^T\mathcal{R}^T\mathcal{R}(\Gamma_i-\Gamma_j) 
    % \\
    % & = (\Gamma_i-\Gamma_j)^T(\Gamma_i-\Gamma_j),
\end{equation}
or
\begin{equation} 
    \label{eq:2}
    |\Gamma_i-\Gamma_j |^2= |\overline{\Gamma}_i-\overline{\Gamma}_j |^2.
\end{equation}
Geometrically, the constraint is intuitive: the distance between two corresponding 3D points must be the same in the two camera coordinates. Let $|\Gamma_i-\Gamma_j|=r$ and expand this equation to reveal the constraints on $\overline{\gamma}_j$ given $\gamma_j,$%This constraint in the image domain is
\begin{equation}
    \label{eq:2.1}
    |\Gamma_j-\Gamma_i |^2= |\overline{\rho}_j\overline{\gamma}_j-\overline{\rho}_i\overline{\gamma}_i |^2=r^2.
\end{equation}
%where $r^2$ is the squared distance of $\Gamma_j$ from $\Gamma_i$. Thus, part of the vector equation~\ref{eq:11} constrains the choice of correspondences so that only two of three equations remains to constrain the pose. This is precisely why three correspondences are required in P3P, which is typically used in RGBD relative pose estimation.\\

\noindent %The paper now proposes that the constraint in Equation~\ref{eq:2.1}
This constraint, referred to here as the {\em Geometric Depth consistency (GDC) constraint}, limits the choice of correspondences and can be utilized to restrict the locus of correspondences $\overline{\gamma}_j$ for a point $\gamma_j$, given a veridical correspondence $(\gamma_i,\overline{\gamma}_i)$. Specifically, expanding Equation~\ref{eq:2.1} gives
%\begin{equation} 
%    \label{eq:3}
%    (\rho_i\gamma_i-\rho_j\gamma_j)^T(\rho_i\gamma_i-\rho_j\gamma_j)=(\overline{\rho}_i \overline{\gamma}_i-\overline{\rho}_j\overline{\gamma}_j)^T(\overline{\rho}_i\overline{\gamma}_i-\overline{\rho}_j\overline{\gamma}_j),
%\end{equation}
%or 
\begin{equation} 
    \label{eq:4}
    \begin{aligned}
   % & (\gamma_i^T\gamma_i)\rho_i^2 - 2(\gamma_i^T\gamma_j)\rho_i\rho_j +(\gamma_j^T\gamma_j)\rho_j^2 \\
    %= &
    (\overline{\gamma}_i^T\overline{\gamma}_i)\overline{\rho}_i^2 - 2(\overline{\gamma}_i^T\overline{\gamma}_j)\overline{\rho}_i\overline{\rho}_j +(\overline{\gamma}_j^T\overline{\gamma}_j)\overline{\rho}_j^2=r^2,
    \end{aligned}
\end{equation}

\noindent where $(\overline{\gamma}_i,\overline{\rho}_i)$ and $r$ are known from the first image. Thus, the only independent unknown is $\overline{\gamma_j}$, with $\overline{\rho}_j(\overline{\gamma}_j)$ being a known dependent variable. This equation then restricts the choice of $\overline{\gamma}_j$ to a curve! Conversely, for any point $\overline{\gamma}_j$, the corresponding point $\gamma_j$ lies on a curve. This partitions the correspondence space into a series of nested curves centered at $\gamma_i$ and $\overline{\gamma}_i$, respectively, parameterized by the latent variable, namely, the radius $r$, as shown in Figure~\ref{fig:fig 2}.
\begin{figure}[!htbp]
    \centering
    \includegraphics[width=.4\textwidth]{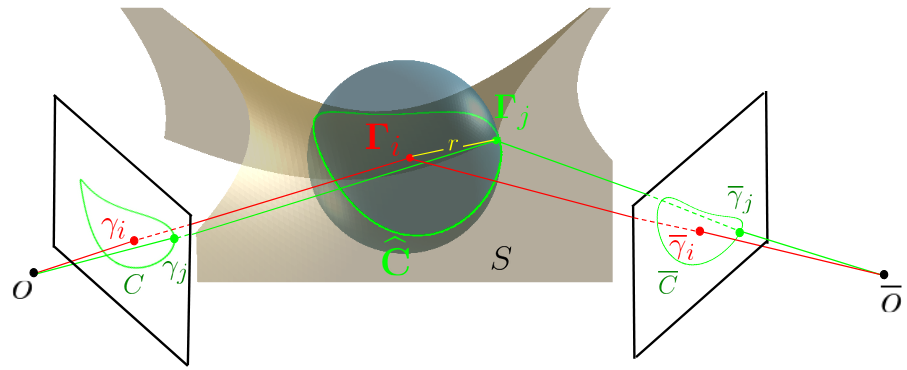}
    \caption{A scene surface $S$ viewed by two cameras. Assume the correspondence $\gamma_i$ and $\overline{\gamma}_i$ both coming from 3D point $\Gamma_i$, a sphere  of radius $r$ centered at $\Gamma_i$ (shown in red), and $S$ intersect at a curve ${\mathcal{\mathbf{\widehat{C}}}}$ (shown in green). The curve ${\mathcal{\mathbf{\widehat{C}}}}$ projects to 2D curves $C$ and $\overline{C}$ in image $i$ and image $j$, respectively. This shows any feature $\gamma_j$ lying on curve $C$ must have its correspondence on curve $\overline{C}$. }
    \label{fig:fig1}
\end{figure}
\begin{figure*}[t]
  \centering
  (a)\includegraphics[width=.15\textwidth]{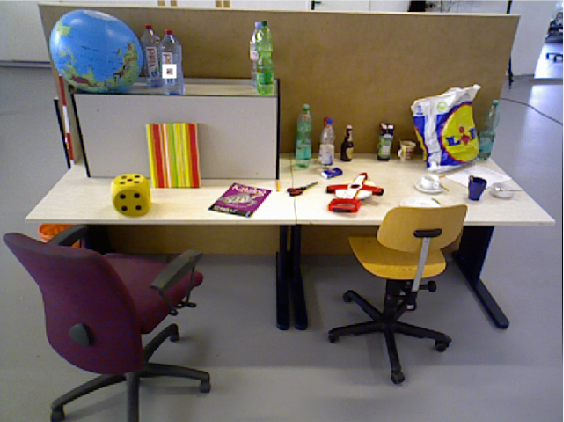}
  \includegraphics[width=.15\textwidth]{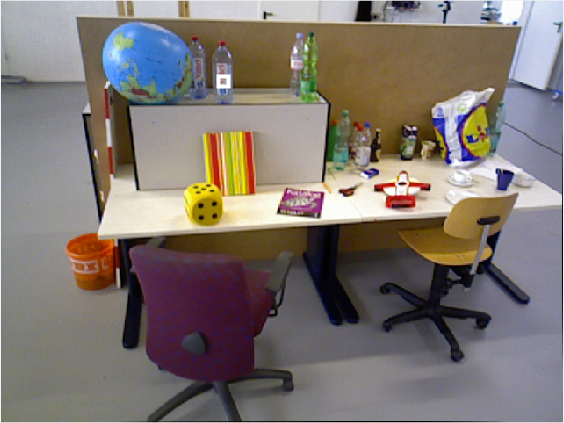}
  \centering
  (b)\includegraphics[width=.15\textwidth]{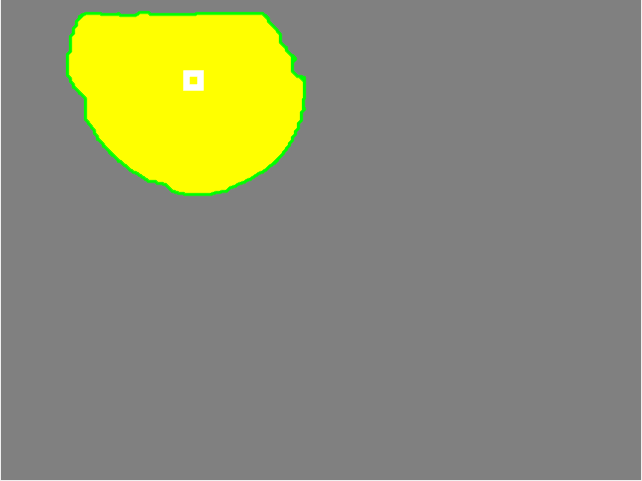}
  \includegraphics[width=.15\textwidth]{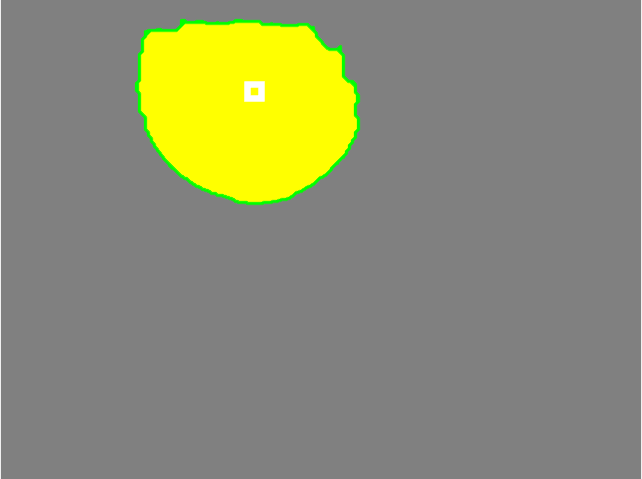}
  (c)\includegraphics[width=.15\textwidth]{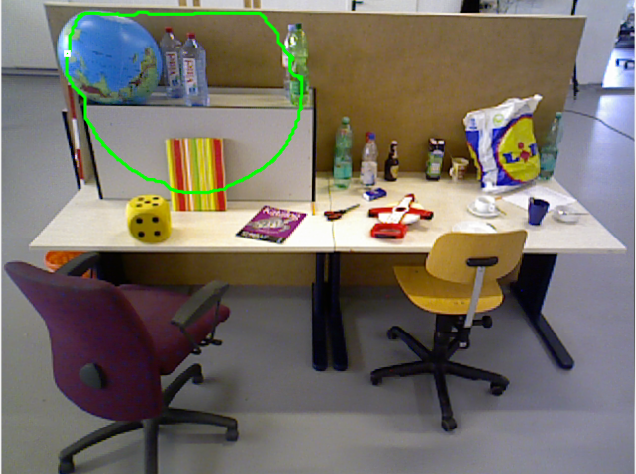}
  \includegraphics[width=.15\textwidth]{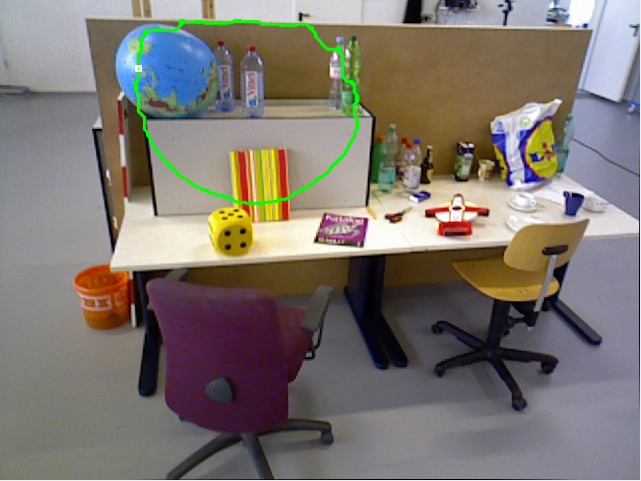}
  \caption{An example illustrating the partitioning of image space based on the geometric depth consistency (GDC) for a pair of RGBD images (a). Given an initial pair of correspondence $(\gamma_i,\overline{\gamma}_i)$ shown as white squares, a pair of corresponding curves are shown in green. (c) The same pair of curves superimposed on the image pair.}
  \label{fig:fig 3}
\end{figure*}
A geometrical examination of this constraint is illuminating. Consider a scene surface $\mathcal{S}$ which is viewed by two cameras, Figure \ref{fig:fig1}. Assume that a correspondence, say ($(\gamma_i,\rho_i$), $(\overline{\gamma}_i,\overline{\rho_i})$) arising from a common point $\Gamma_i=\rho_i\gamma_i$ expressed in camera one, and expressed as $\overline{\Gamma}_i=\overline{\rho}_i\overline{\gamma}_i$ in camera two, is known, as described earlier. Then, for any point in camera one, $\gamma_j$ with depth $\rho_j$, the 3D point $\Gamma_j=\rho_j\gamma_j$ is known, while $\overline{\gamma}_j$ is unknown. Thus, Equation (\ref{eq:2}) is effectively describing a sphere centered at $\overline{\Gamma}_i$ with radius $r=|\Gamma_j-\Gamma_i|$, as the locus of $\overline{\Gamma}_j$. Since $\overline{\Gamma}_j$ also lie on the surface $\mathcal{S}$, the locus of $\overline{\Gamma}_j$ is the intersection of the sphere and scene surface, as shown by the green curve $\mathcal{\mathbf{\widehat{C}}}$ in Figure \ref{fig:fig1}.
%The question arises: given a point $\gamma_j$ with depth $\rho_j$ in camera one, what is the locus of the corresponding point $\overline{\gamma}_j$? The 3D point corresponding to $\gamma_j$ is $\Gamma_j$ as expressed in camera one and $\overline{\Gamma}_j$ as expressed in camera two, with $\Gamma_j=\rho_j\gamma_j$, and $\overline{\Gamma}_j=\overline{\rho}_j\overline{\gamma}_j$. Let $r=|\Gamma_i-\Gamma_j|$, then we must have $|\overline{\Gamma}_i-\overline{\Gamma}_j|=|\Gamma_i-\Gamma_j|=r$. In general, given a constant $r$, the surface points ${\Gamma}_j$ with distance $r$ from ${\Gamma}_i$ are simultaneously on the sphere of radius $r$ centered at $\overline{\Gamma}_i$ and on the surface $S$, Figure~\ref{fig:fig1}. Thus, the locus of ${\Gamma}_j$ with distance $r$ from ${\Gamma}_i$ traces out a 3D curve ${\mathcal{\mathbf{\widehat{C}}}}$. 

\noindent{\bf Geometric Depth Constraint:} The projection of this 3D curve ${\mathcal{\mathbf{\widehat{C}}}}$ on the two cameras traces out 2D curves, $\mathcal{C}$ and $\mathcal{\overline{C}}$, on the first and second images, respectively, Figure \ref{fig:fig1}. Thus, any point on $\mathcal{C}$ can only have its correspondences on $\mathcal{\overline{C}}$, and conversely, any point on $\overline{C}$ can only have its correspondences on $\mathcal{C}$. This is a significant restriction on the possible correspondences for $(\gamma_j$, $\overline{\gamma}_j$). In contrast in RGB images, one correspondence $(\gamma_i,\overline{\gamma}_i)$ does not constrain any other correspondence $(\gamma_j,\overline{\gamma}_j)$ at all.

The latent parameter that partitions the space around $\gamma_i$ and $\overline{\gamma_i}$ into a set of nested corresponding curves, Figure \ref{fig:fig 2}, is the radius $r$. Specifically, given a pair of veridical correspondences $(\gamma_i,\overline{\gamma}_i)$, a radial map is constructed for each image to facilitate this partition.

\begin{definition}
   The squared radial map \footnote{The squared radius is maintained rather than radius to avoid an unnecessary square root operation when all subsequent operations involve a comparison of radii, which can be done in square form.} of an RGBD image $(R(\xi,\eta),G(\xi,\eta),B(\xi,\eta),\rho(\xi,\eta))$ with respect to a reference point $(\gamma_0,\rho_0)$ is defined as \begin{equation}
    \phi(\xi,\eta) = r^2(\xi,\eta)=|\rho(\xi,\eta)\gamma(\xi,\eta)-\rho_{0}\gamma_{0}|^2.
\end{equation} 
\end{definition}

\noindent Then, given a point $\gamma_j(\xi_j,\eta_j)$, $r(\xi_j,\eta_j)$ is computed from the first image and used to restrict the locus of $\overline{r}_j(\overline{\xi}_j,\overline{\eta}_j)$ to all points $(\overline{\xi}_j,\overline{\eta}_j)$ satisfying \begin{equation}
    \overline{r}_j^2(\overline{\xi}_j,\overline{\eta}_j)=r_j^2(\xi_j,\eta_j),
\end{equation} which is effectively a level set of the squared radial map for the second image. Figure~\ref{fig:fig 3} illustrates this on a pair of RGBD images shown in (a). A pair of corresponding curves are selected and shown in (b) and also shown superposed on the image. Thus, a point on the green curve shown in purple in the first image has a number of correspondence options lying on the corresponding given curve in the second image.

The space of potential correspondences is generally limited to a set of rank-ordered correspondences, as typically used in the classic RANSAC approach. The combination of having a discrete set of correspondences and the GDC constraint significantly reduces the set of possible correspondences. Assume that a veridical pair is available, as shown by the white square tokens, Figure~\ref{fig:radial_maps_rgbd1_12} (right), which are the center of nested curves around them. Consider now an arbitrary pair of correspondences. The GDC requires that the pair fall on corresponding curves, rules out a majority of erroneous correspondences (shown in black tokens), retaining only veridical correspondences and those non-veridical correspondences which coincidentally fall on corresponding curves. This is in fact a filter which can be used in the RANSAC scheme as discussed in the next section.

Finding the level sets in a RGBD map from a veridical correspondence and computing the distance from an image point to the corresponding curve under a RANSAC loop is, however, inefficient in practice. An alternative, efficient approach is thus proposed.
\begin{prop}
\label{lemma:distance_from_point2curve}
Let $\phi$ and $\overline{\phi}$ be the squared radial maps of the first and second RGBD images, respectively from the reference points $(\gamma_0, \rho_0)$ and $(\overline{\gamma}_0, \overline{\rho}_0)$. Given a putative correspondence, $(\gamma, \overline{\gamma})$, the distance $\overline{d}$ of $\overline{\gamma}$ from the corresponding curve is
\begin{equation}
    % d = \frac{|\phi - \overline{\phi}|}{||\nabla \overline{\phi}||},
    \overline{d} = \frac{|\phi - \overline{\phi}|}{\begin{vmatrix} 2\left(\overline{\rho}_i ||\overline{\gamma}_i||^2- \overline{\rho}_0 \overline{\gamma}_0^T \overline{\gamma_i} \right) \nabla \overline{\rho}_i + 2\overline{\rho}_i \begin{bmatrix} \overline{\rho}_i \overline{\xi} - \overline{\rho}_0 \overline{\xi}_0 \\ \overline{\rho}_i \overline{\eta} - \overline{\rho}_0 \overline{\eta}_0 \end{bmatrix} \end{vmatrix}},
\end{equation}
where $\overline{\rho}(\overline{\xi}_i, \overline{\eta}_i)$ is the depth at $\overline{\gamma}(\overline{\xi}_i, \overline{\eta}_i)$.
% \begin{equation}
%     \nabla \overline{\phi} = 2\left(\overline{\rho}_i ||\overline{\gamma}_i||^2- \overline{\rho}_0 \overline{\gamma}_0^T \overline{\gamma_i} \right) \nabla \overline{\rho}_i + 2\overline{\rho}_i \begin{bmatrix} \frac{\overline{\rho}_i \overline{\xi} - \overline{\rho}_0 \overline{\xi}_0}{f_x} \\ \frac{\overline{\rho}_i \overline{\eta} - \overline{\rho}_0 \overline{\eta}_0}{f_y} \end{bmatrix}.
% \end{equation}
\end{prop}
\noindent The proof is given in the supplementary materials.

\section{Filtered RANSAC}

The GDC filter can be used to avoid unnecessary computations in RANSAC. Observe that the computational cost of the classic approach as broken down in Table~\ref{tab:timings_break_down}. The cost of a hypothesis consists of the hypothesis formulation cost denoted by $\alpha\sim 1\mu$s, and the cost of measuring hypothesis support, which itself involves relative pose estimation and computing the number of inliers, denoted by $\beta\sim 45\mu$s. The total cost per hypothesis is clearly dominated by the latter which it is the product of the total hypothesis cost and the number of iterations $N$ required to achieve a certain success rate, {\em i.e.,} $N(\alpha + \beta)$, 
\begin{equation}
    N\leq \frac{\log(1-p)}{\log(1-(1-e)^s)},
    \label{eqn:min_n}
\end{equation} 
\begin{table}[!t]
    \centering
    \footnotesize
    \begin{tabular}{|c|c|c|}
    \hline
    Steps  & Classic ($\mu s$) & GDC ($\mu s$) \\
    \hline
    \rowcolor{lightgray} Hypothesis formulation cost & 0.96& 5.61\\
    \hline
    Absolute Pose Estimation per hypothesis & 27.7 & 27.7 \\
    \hline
    Find Number of inliers per hypothesis & 17.4 & 17.4   \\
    \hline
    \rowcolor{lightgray} Hypothesis support measurement cost & 45.2 & 45.2 \\
    \hline
    \rowcolor{yellow}Average cost of evaluating a hypothesis & 46.1 & 50.8 \\
    \hline
    \end{tabular}
    \caption{The classic RANSAC scheme first formulate hypotheses which allow for pose estimation and computing the number of inliers. The costs of the two stages and a breakdown for the second stage is given. Note that \emph{(i)} the cost of RANSAC is dominated by the second stage so that eliminating the second stage through a filter presents significant savings, and \emph{(ii)} the cost of hypothesis formation with the GDC filter is only slightly increased.}
    \label{tab:timings_break_down}
\end{table}

\noindent where $p$ is the required probability of success, $e$ is the proportion of outliers, and $s$ is the number of samples required to form a hypothesis ($s$ = 3 in our case). For example, with $e$ = 70$\%$ and $p = 99$$\%$, the required number of iterations is 169. This number changes rapidly with outlier ratio so that with $e = 80\%,$ $N = 574$ and with $e$ = 60\%, $N$ = 70. 
% For the TUM-RGBD dataset, the experiments show that $N = 89$ achieves $p = 99\%$ success rate, indicating an average outlier ratio of $e=63\%$, but as Figure~\ref{fig:dist_e} shows the outlier ratio varies quite a bit.

%> Default is 6pt
\setlength\tabcolsep{2pt} 
\begin{table*}[t]
\centering
{\footnotesize
    \begin{tabular}{|c|c|c|c|c|c|c|c|c|c|c|}
    \hline
    % \multirow{ 2}{*}{
    % } & \multicolumn{10}{|c|}{\textbf{Timings per RANSAC iteration ($\mu$s)}} \\
    % \cline{2-11}
    & \multicolumn{2}{!{\vrule width 2pt}c|}{$e$ = 60-70$\%$} & \multicolumn{2}{!{\vrule width 2pt}c|}{$e$ = 70-80$\%$} &\multicolumn{2}{!{\vrule width 2pt}c|}{$e$ = 80-90$\%$} &\multicolumn{2}{!{\vrule width 2pt}c|}{$e$ = 90-95$\%$} &\multicolumn{2}{!{\vrule width 2pt}c|}{$e$ = 95-99$\%$} \\
    \cline{2-11}
    & \multicolumn{1}{!{\vrule width 2pt}c|}{\textbf{Classic}} & \makecell{\textbf{GDC-} \\ \textbf{Filtered}} & \multicolumn{1}{!{\vrule width 2pt}c|}{\textbf{Classic}} & \makecell{\textbf{GDC-} \\ \textbf{Filtered}} & \multicolumn{1}{!{\vrule width 2pt}c|}{\textbf{Classic}} & \makecell{\textbf{GDC-} \\ \textbf{Filtered}}  & \multicolumn{1}{!{\vrule width 2pt}c|}{\textbf{Classic}} &\makecell{\textbf{GDC-} \\ \textbf{Filtered}} & \multicolumn{1}{!{\vrule width 2pt}c|}{\textbf{Classic}} &\makecell{\textbf{GDC-} \\ \textbf{Filtered}}  \\
    \hline
    \makecell{$\#$ of RANSAC iterations \\ (99$\%$ success rate)} & \multicolumn{1}{!{\vrule width 2pt}c|}{169} & 169$\to$44 & \multicolumn{1}{!{\vrule width 2pt}c|}{420} & 420$\to$85 & \multicolumn{1}{!{\vrule width 2pt}c|}{3752} & 3752$\to$533 & \multicolumn{1}{!{\vrule width 2pt}c|}{21375} & 21375$\to$876 & \multicolumn{1}{!{\vrule width 2pt}c|}{681274} & 681274$\to$14374  \\
    \hline
    Hypothesis formation cost (ms) & \multicolumn{1}{!{\vrule width 2pt}c|}{0.16} & 0.94 & \multicolumn{1}{!{\vrule width 2pt}c|}{0.40} & 2.35 & \multicolumn{1}{!{\vrule width 2pt}c|}{3.60} & 21.04 & \multicolumn{1}{!{\vrule width 2pt}c|}{20.52} & 119.91 & \multicolumn{1}{!{\vrule width 2pt}c|}{654.02} & 3821.95\\
    \hline
    Hypothesis support measurement cost (ms) & \multicolumn{1}{!{\vrule width 2pt}c|}{7.64} & 1.988 & \multicolumn{1}{!{\vrule width 2pt}c|}{18.98} & 3.84 & \multicolumn{1}{!{\vrule width 2pt}c|}{169.59} & 24.09 & \multicolumn{1}{!{\vrule width 2pt}c|}{966.15} & 35.60 & \multicolumn{1}{!{\vrule width 2pt}c|}{30793.58} & 649.72\\
    \hline
    \hline
    \rowcolor{yellow}\textbf{Total Cost (ms) TUM-RGBD} & \multicolumn{1}{!{\vrule width 2pt}c|}{7.80} & 2.94 & \multicolumn{1}{!{\vrule width 2pt}c|}{19.39} & 6.19 & \multicolumn{1}{!{\vrule width 2pt}c|}{173.19} & 45.14 & \multicolumn{1}{!{\vrule width 2pt}c|}{986.67} & 155.513 & \multicolumn{1}{!{\vrule width 2pt}c|}{31447.61} & 3822.60 \\
    % \hline
    % \rowcolor{yellow}\textbf{Total Cost (ms) ICL-NUIM} & \multicolumn{1}{!{\vrule width 2pt}c|}{40.53} & 8.03 & \multicolumn{1}{!{\vrule width 2pt}c|}{79.07} & 6.89 & \multicolumn{1}{!{\vrule width 2pt}c|}{} &  & \multicolumn{1}{!{\vrule width 2pt}c|}{} &  & \multicolumn{1}{!{\vrule width 2pt}c|}{} & \\
    % \hline
    % \rowcolor{yellow}\textbf{Total Cost (ms) RGBD Scene v2} & \multicolumn{1}{!{\vrule width 2pt}c|}{39.3} & 5.77 & \multicolumn{1}{!{\vrule width 2pt}c|}{136.2} & 7.63 & \multicolumn{1}{!{\vrule width 2pt}c|}{335.2} & 12.8 & \multicolumn{1}{!{\vrule width 2pt}c|}{1150.68} & 39.68 & \multicolumn{1}{!{\vrule width 2pt}c|}{320909} & 122.53 \\
    \hline
    \end{tabular}
    \caption{Cost of unfiltered (traditional RANSAC) and filtered RANSAC (GDC constraints applied) for 99$\%$ success rate over the entire TUM-RGBD dataset, with a grand total of 132,946 image pairs, with successful pose estimation defined as having less than 0.5 degree in rotation and 0.05 meters in translation. GDC-Filtered columns are the number of RANSAC iterations and the number of hypothesis passing the GDC test. Specifically, each image is paired with subsequent image at intervals ranging from 1 to 30 time steps. The resultant image pairs are then put into discrete outlier ratio bins. Experiments for other datasets are given in the supplementary materials.} 
    \label{tab:GDC_Profiling}
}
\end{table*}

\begin{figure}[t]
    \centering
    (a)\includegraphics[width=.2\textwidth]{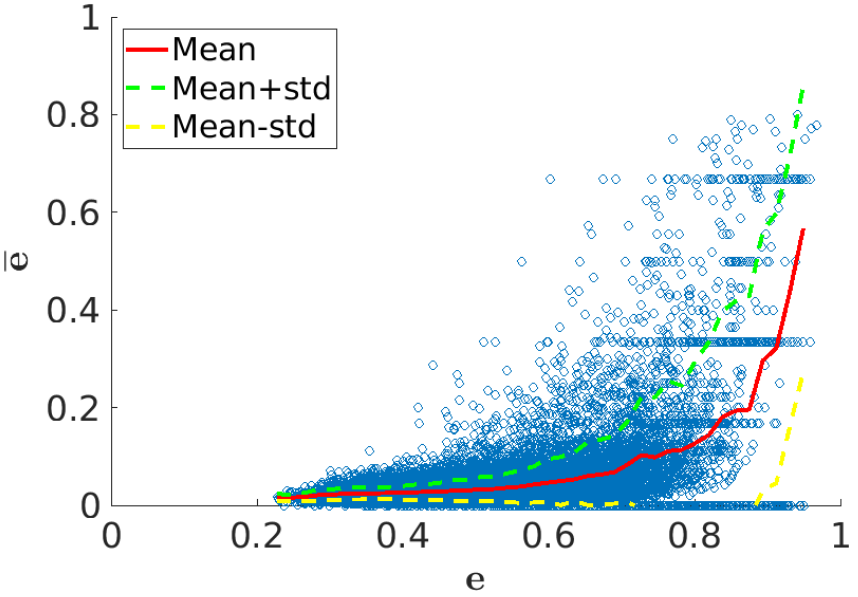}
    % (b)\includegraphics[width=.2\textwidth]{Figures/nbar_vs_n_0.99_many.png} (c)\includegraphics[width=.2\textwidth]{Figures/mu_vs_n_0.99_many.png}
    (b) \includegraphics[width=.2\textwidth]{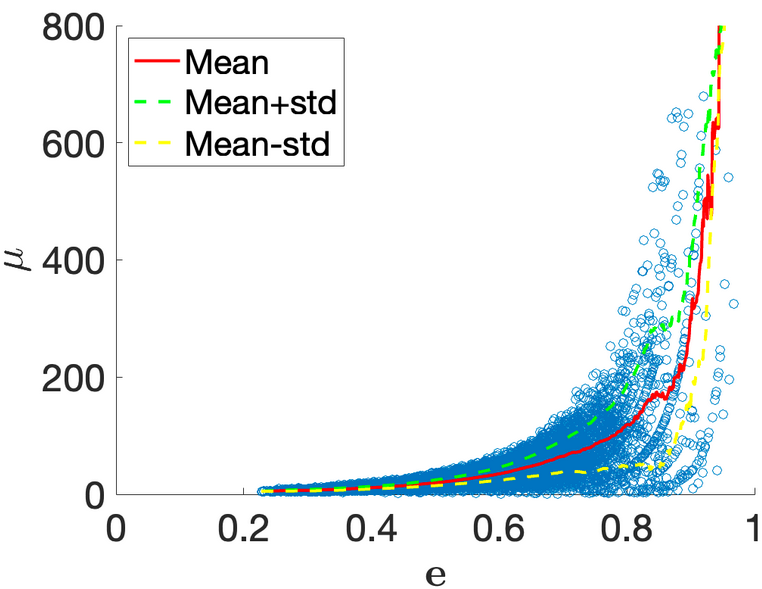}
    \caption{(a) The scatter plot of $e$ and $\overline{e}$, namely, the outlier ratios before and after the GDC filter is applied and (b) the ratio of the number of required iterations before and after applying the GDC filter to TUM-RGBD~\cite{TUMRGBD_Dataset} dataset for success probability of 0.99. Note that the scale is too small to appropriate that at (0.2, 0.3, 0.4, 0.5, 0.6) the value of $\mu$ is (5, 7, 12, 21, 37), respectively. }
    \label{fig:e}
\end{figure}

\noindent {\bf Filtering RANSAC Hypotheses:} Observe that since the main bulk of the computational expense is in measuring hypothesis support, the GDC constraint can be used as a {\em filter} to discard incorrect hypotheses, thus leading to significant savings with only a modest increase in hypothesis formulation cost, from $\alpha=0.96 \; \mu s$ to $\alpha=5.61 \; \mu s$. Figure~\ref{fig:e}(a) illustrates that the GDC filter reduces the outlier ratio significantly from $e$ to $\overline{e}$, which in turn requires significantly fewer iterations from $N$ to $\overline{N}$, where the ratio $\mu$
\begin{equation}
    \mu=\frac{N}{\overline{N}}=\frac{[\log(1-(1-\overline{e})^s)]}{[\log(1-(1-e)^s)]},
\end{equation}
measures the savings in the number of iterations.
It is interesting that the ratio $\mu$ is independent of the probability of success $p$ and is exponentially increasing with outlier ratio~$e$, Figure~\ref{fig:e}(b). Table~\ref{tab:GDC_Profiling} summarizes the time savings as a result of this filter, where the hypotheses are selected from the top $M$ = 250 of the rank-ordered list.

\begin{figure}[b]
    \centering
    (a)\includegraphics[scale=0.2]{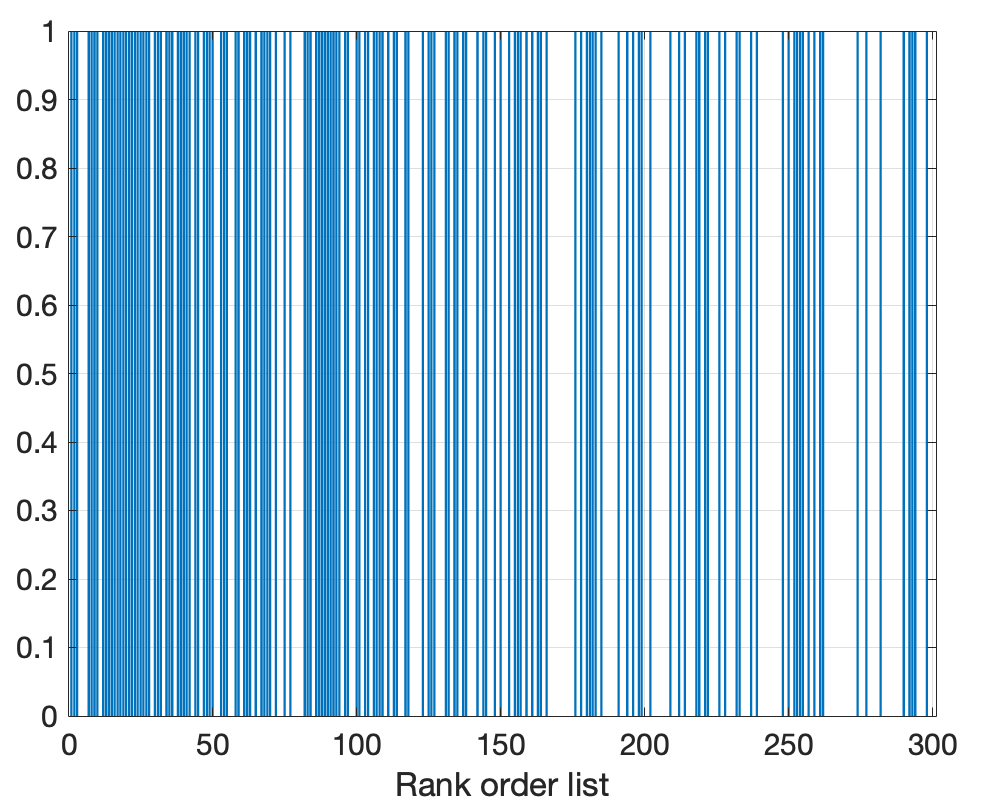}
    (b)\includegraphics[scale=0.2]{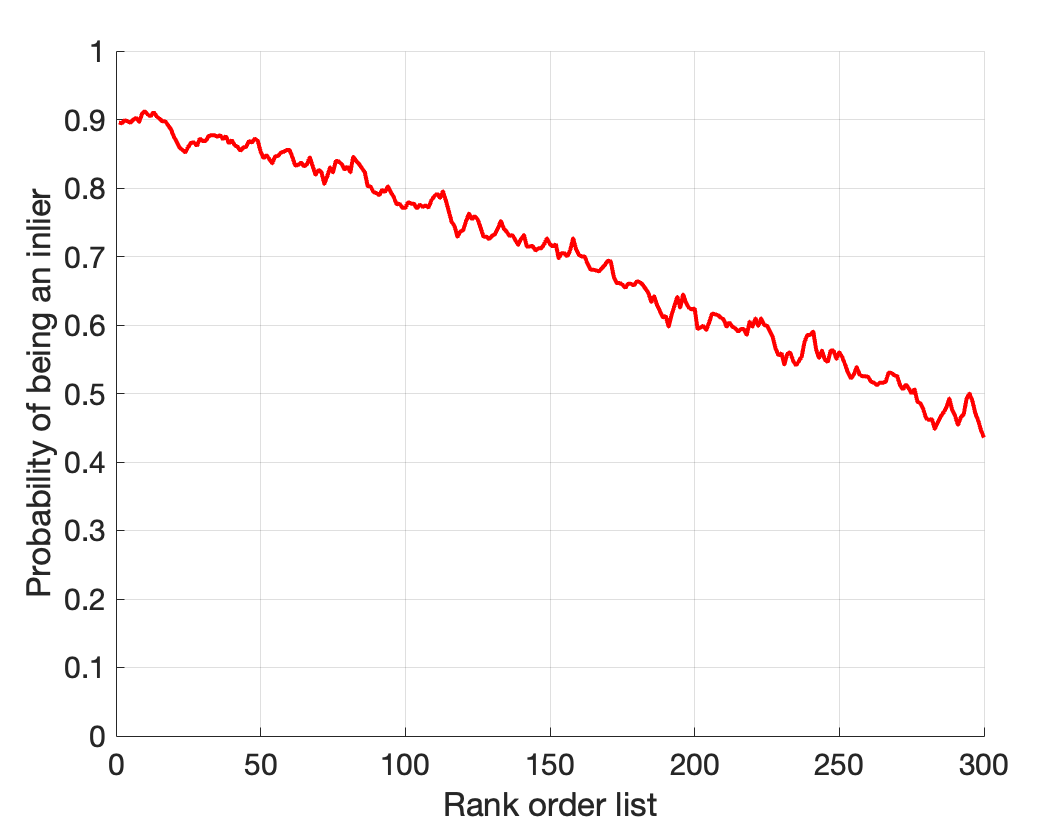}
    \caption{(a) Each correspondence for a given image is either veridical or incorrect. (b) The likelihood that the selection at rank $m$ is veridical, an average over all the binary plots in the dataset.
    % over all the images in the TUM-RGBD~\cite{TUMRGBD_Dataset} dataset.
    }
    \label{fig:prob_of_correctness}
\end{figure}

\begin{figure*}[b]
    \centering
    \includegraphics[scale=0.19]{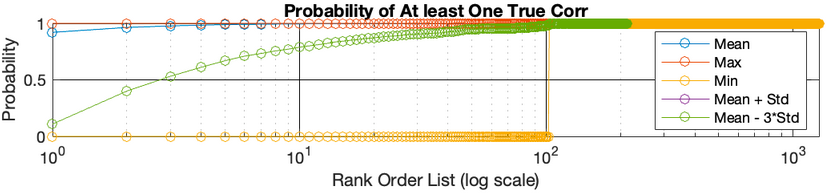}
    \includegraphics[scale=0.19]{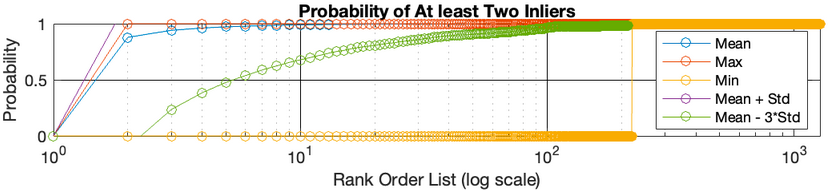}
    \includegraphics[scale=0.19]{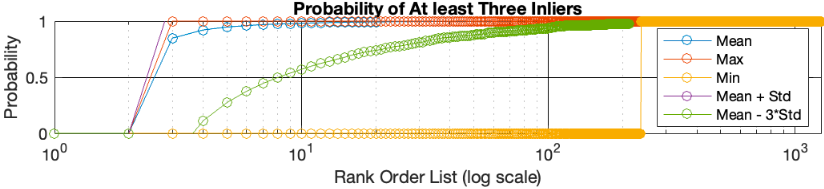}
    \caption{Left to right: The probability of finding at least one, at least two, and at least three inliers across the rank-ordered list of matches. }
    \label{fig:prob}
\end{figure*}
\section{Nested RANSAC}
\label{sec:Nested_RANSAC_New}

% The traditional RANSAC approach relates the outlier ratio~$e$ of a group of $M$ correspondences from which $s$ selections form a hypothesis and the minimum acceptable confidence $p$ to the minimum number of trials $N$ needed to reach this confidence level. Specifically, the probability of $N$ trials of $s$ picks all being incorrect is $(1-(1-e)^s)^N$, which needs to be lower than $(1-p)$,
% \begin{equation}
%     (1-(1-e)^s)^N \leq 1-p,\;\text{or}\;N \leq \frac{\log (1-p)}{\log (1-(1-e )^s)}.
% \end{equation}
% \begin{equation}
%     \left( 1-\left(1-e \right)^s \right)^N \leq 1-p,
% \end{equation}
% \noindent or,
% \begin{equation}
%     N \leq \frac{\log \left(1-p\right)}{\log \left( 1-\left(1-e \right)^s \right)}.
% \end{equation}

\begin{figure}[t]
    \centering
    %(a)\includegraphics[scale=0.2]{Figures/nu_versus_e_vary_e1.png}
    (a)\includegraphics[scale=0.19]{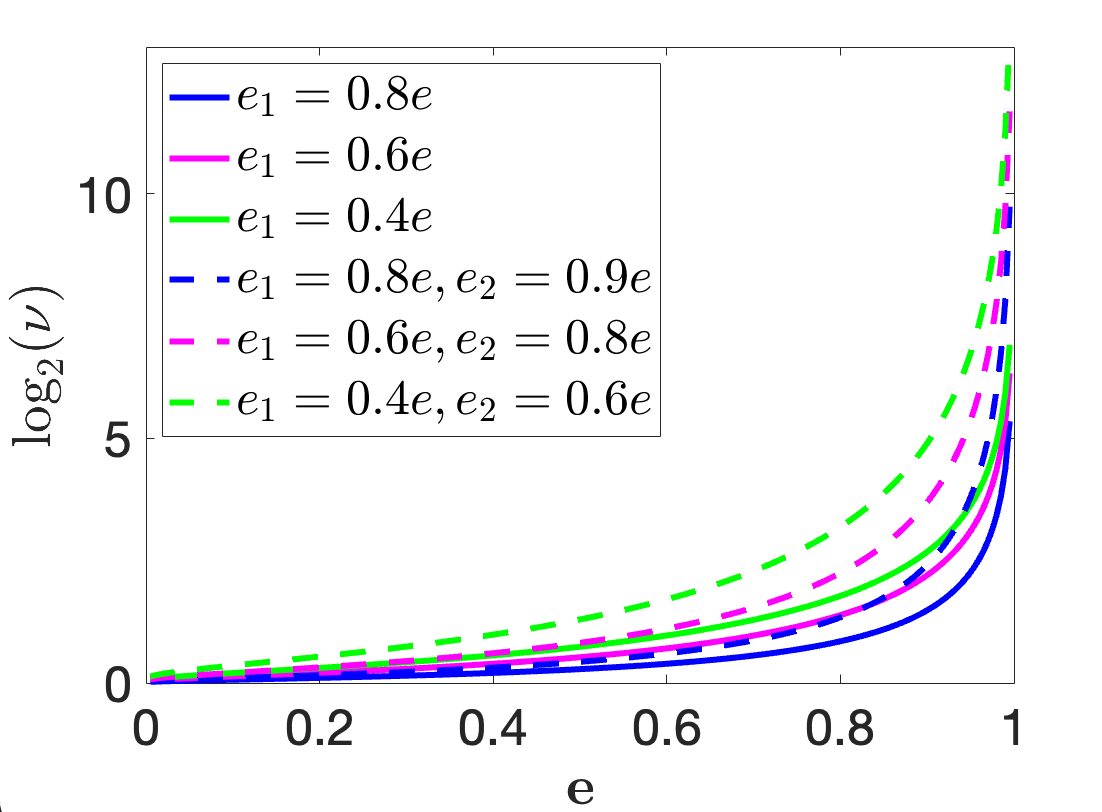}
    (b)\includegraphics[scale=0.19]{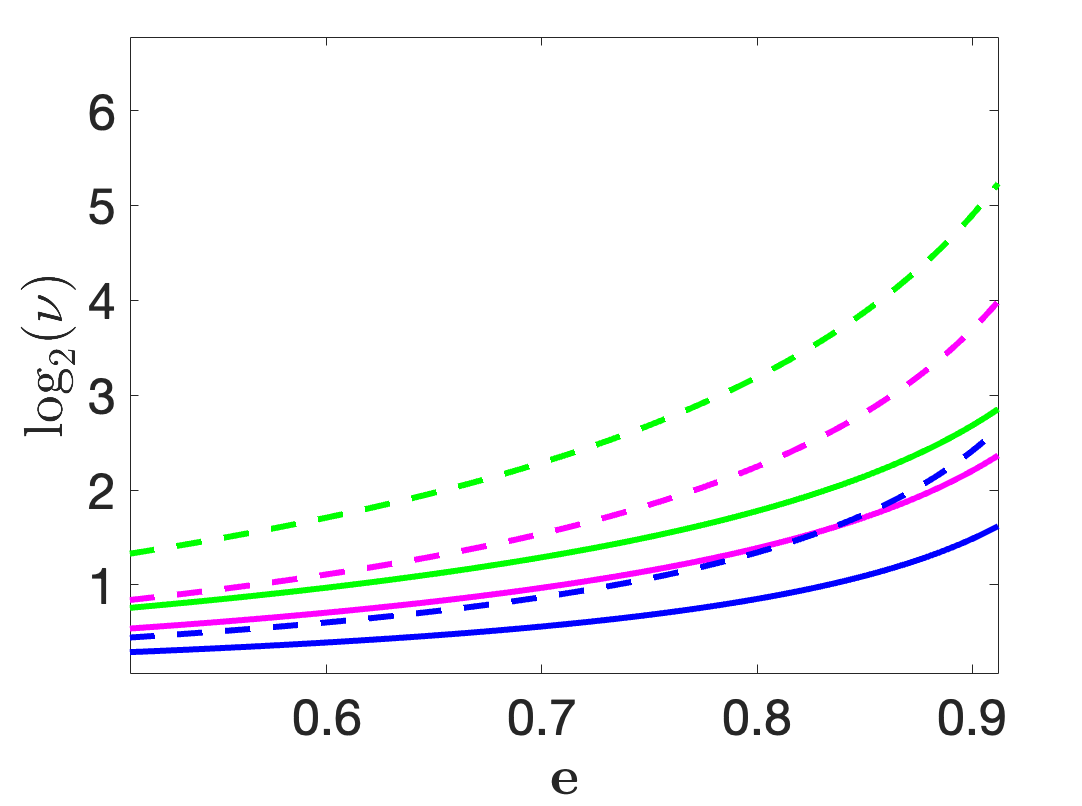}
    %(c)\includegraphics[scale=0.2]{Figures/nu_versus_e_vary_e1_and_e2.png}
    %(d)\includegraphics[scale=0.2]{Figures/nu_versus_e_vary_e1_and_e2_zoomin.png}
    \caption{The extent of savings $\nu$ as a function of outlier ratio $e$ for $s$ = 3 when (a) $e_1=0.8e$, $e_1=0.6e$, and $e_1=0.4e$ for nested (-) and $e_2=0.9e$, $e_2=0.8e$, and $e_2=0.6e$ for doubly nested (-$\,$-). (b) A zoom-in window of (a) }%at around $e=0.6$; (c) various $e_1$ and $e_2$; (d) a zoom-in window of (c) at around $e=0.6$. Note that the savings $\nu$ in (a) and (c) are in log scale for clear visualizations.}
    \label{fig:extent_of_savings_e1}
\end{figure}

In the classic RANSAC approach, all the $s$ correspondences are selected uniformly from the $M$ top-ranking of correspondences. However, the likelihood of a correspondence being correct is not uniform! It drops as one goes down the rank-ordered list. Figure~\ref{fig:prob_of_correctness}(a) plots for each rank $0 \leq m \leq M$ on the $x$-axis whether the selection at the rank is correct (1) or incorrect (0). The higher density at lower values of $m$ indicate that the higher the rank the greater the probability of the selection being correct. This is verified in Figure~\ref{fig:prob_of_correctness}(b) which averages the binary plot over all pairs of images in the TUM-RGBD~\cite{TUMRGBD_Dataset}. Clearly, the high-ranking choices are more likely to be correct and this expectation drops as the rank increases.\\
\indent This non-uniformity behooves us to bias the selection of the $s$ correspondences in favor of the top-ranking choices, in contrast to the traditional RANSAC where the selection is uniform. 
Unfortunately, the option of reducing $M$ outright has the negative effect of either removing all veridical $s$-tuplets, or reducing the probability of selecting them from a smaller pool when faced with a high ratio of outliers. Instead, this paper proposes that biasing the selection of $s$ selections towards the top-ranking hypotheses, perhaps by a probability distribution, would increase the chance of finding an a veridical set of $s$ correspondences.\\\noindent % \begin{figure}[!htbp]
%   \centering
%   \centering
%   %\includegraphics[scale=0.35]{Figures/inlier_ratio_27Aug.png} 
%   \includegraphics[width=0.3\textwidth]{Figures/inlier_ratio_all_aug30.png}
%   \label{fig:s1}
%   \caption{The inlier ratio ($\#$ of inliers/$N$) for the selected rank order feature correspondence as a function of $N$ for over RGBD image pairs}
%   \label{fig:inlier1_new}
% \end{figure}
% \begin{figure}[!htbp]
%     \centering
%     \includegraphics[scale=0.25]{Figures/inlier_ratio_vs_num_of_matches_27Aug.png}
%     \caption{The mean, std deviation of the first occurrence of an inlier in the first $N$ selected matches, where $N$ varies from 1 to total number of available matches. The minimum case is 6 \emph{i.e.} in the worst case, the top 6 has no inlier.}
%     \label{fig:enter-label_new}
% \end{figure}
%> Default is 6pt
\setlength\tabcolsep{1.5pt} 
\begin{table*}[t]
\centering
{\footnotesize
    \begin{tabular}{|c|c|c|c|c|c|c|c|c|c|c|c|c|c|c|c|}
    \hline
    & \multicolumn{3}{!{\vrule width 2pt}c|}{$e$ = 60-70$\%$} & \multicolumn{3}{!{\vrule width 2pt}c|}{$e$ = 70-80$\%$} & \multicolumn{3}{!{\vrule width 2pt}c|}{$e$ = 80-90$\%$} & \multicolumn{3}{!{\vrule width 2pt}c|}{$e$ = 90-95$\%$} & \multicolumn{3}{!{\vrule width 2pt}c|}{$e$ = 95-99$\%$} \\
    \cline{2-16}
    & \multicolumn{1}{!{\vrule width 2pt}c|}{{\rotatebox{90}{\textbf{Classic}}}} & {\rotatebox{90}{\textbf{Nested}}} & {\rotatebox{90}{\textbf{\makecell{Doubly\\ Nested}}\;}} & \multicolumn{1}{!{\vrule width 2pt}c|}{{\rotatebox{90}{\textbf{Classic}}}} & {\rotatebox{90}{\textbf{Nested}}} & {\rotatebox{90}{\textbf{\makecell{Doubly\\ Nested}}}} & \multicolumn{1}{!{\vrule width 2pt}c|}{{\rotatebox{90}{\textbf{Classic}}}} & {\rotatebox{90}{\textbf{Nested}}} & {\rotatebox{90}{\textbf{\makecell{Doubly\\ Nested}}}} & \multicolumn{1}{!{\vrule width 2pt}c|}{{\rotatebox{90}{\textbf{Classic}}}} & {\rotatebox{90}{\textbf{Nested}}} & {\rotatebox{90}{\textbf{\makecell{Doubly\\ Nested}}}} & \multicolumn{1}{!{\vrule width 2pt}c|}{{\rotatebox{90}{\textbf{Classic}}}} & {\rotatebox{90}{\textbf{Nested}}} & {\rotatebox{90}{\textbf{\makecell{Doubly\\ Nested}}}} \\
    \hline
    \makecell{$\#$ of matches from top rank-\\ordered list: $M_1$/$M_2$/$M_3$} & \multicolumn{1}{!{\vrule width 2pt}c|}{250} &  \makecell{100/\\250} & \makecell{100/\\150/\\250}& \multicolumn{1}{!{\vrule width 2pt}c|}{250} & \makecell{100/\\250}& \makecell{100/\\150/\\250} & \multicolumn{1}{!{\vrule width 2pt}c|}{250} & \makecell{100/\\250}& \makecell{100/\\150/\\250}& \multicolumn{1}{!{\vrule width 2pt}c|}{250} & \makecell{100/\\250} & \makecell{100/\\150/\\250} & \multicolumn{1}{!{\vrule width 2pt}c|}{250} &  \makecell{100/\\250} & \makecell{100/\\150/\\250} \\
    \hline
    \makecell{$\#$ of RANSAC iterations \\ (99$\%$ success rate)} & \multicolumn{1}{!{\vrule width 2pt}c|}{169} & 146 & 137& \multicolumn{1}{!{\vrule width 2pt}c|}{420} & 371 & 227 & \multicolumn{1}{!{\vrule width 2pt}c|}{3752} & 2218 & 2126 & \multicolumn{1}{!{\vrule width 2pt}c|}{21375} & 17509 & 6872 & \multicolumn{1}{!{\vrule width 2pt}c|}{681274} & 220637 &68824  \\
    \hline
    \hline
    \rowcolor{yellow}\textbf{Total Cost (ms) TUM-RGBD} & \multicolumn{1}{!{\vrule width 2pt}c|}{7.80} & {6.73} & {6.32} & \multicolumn{1}{!{\vrule width 2pt}c|}{19.38} & {17.1} & {10.46} & \multicolumn{1}{!{\vrule width 2pt}c|}{173.19} &{102.25} & {98} & \multicolumn{1}{!{\vrule width 2pt}c|}{986.67} & {808.21} & { 317.21} & \multicolumn{1}{!{\vrule width 2pt}c|}{31447.61} & {10184.60} & {3176.92} \\
    \hline
    \end{tabular}
    \caption{Cost of traditional, nested, and doubly nested RANSAC for 99$\%$ success rate over the entire TUM-RGBD dataset. Experiments for other datasets are given in the supplementary materials.} 
    \label{tab:nested_RANSAC_time_profiling}
}
\end{table*}
\indent Specifically, observe that when making a single correspondence selection, as compared to three or five sets of correspondences, the number of top ranking correspondences $M$ can be drastically reduced without affecting the outcome. Figure~\ref{fig:prob} shows the likelihood of finding $s$-tuplets in the top $m$ set of correspondences for $1 \leq m \leq M$ for different values of $s$, showing that selecting a single correspondence can be done in the top $M_1$, where $M_1$ is significantly lower than $M$, \emph{e.g.}, $M_1 \geq 150$ in Figure~\ref{fig:prob}. Let $e_1$ be the outlier rate of the top $M_1$ with the expectation that $e_1$ is significantly less than $e$. Thus, 
the probability of picking the first selection being correct is $(1-e_1)$, while the probability of the next $(s-1)$ selection from the top $M$ is $(1-e)^{s-1}$, with $(1-e_1)< (1-e)$. Thus, the probability of the overall $s$ selections being correct is $(1-(1-e_1) (1-e)^{s-1} )$ which is greater than the classic value of $1-(1-e)^s$. Hence, the probability of $\overline{N}$  RANSAC iterations satisfying the $p$ confidence level is $(1-( 1-e_1) (1-e )^{s-1} )^{\overline{N}} < 1-p$, or
\begin{equation}
    \overline{N} \leq \frac{\log (1-p)}{\log (1-( 1-e_1) (1-e )^{s-1} )}.
\end{equation}
\noindent Thus, $\overline{N}$ is significantly lower than $N$ with the improvement captured by the ratio
\begin{equation}
    \nu = \frac{N}{\overline{N}} = \frac{\log (1-( 1-e_1) (1-e)^{s-1} ) }{\log (1-(1-e)^s)}.
\end{equation}
Figure~\ref{fig:extent_of_savings_e1}(a) plots $\nu$ as a function of outlier ratio $e$ and shows exponentially increasing savings for each of the cases $e_1=0.8e$, $e_1=0.6e$, and $e_1=0.4e$. 

Consider now taking this ``nested RANSAC approach" a step further, \emph{i.e.}, let the first choice be from the top $M_1$ with outlier ratio $e_1$ and the second choice be from $M_2$ with outlier ratio $e_2$, where the remaining $s-2$ selections are again from the top $M$. Then the improvement in the number of iterations is
\begin{equation}
    \nu = \frac{N}{\overline{N}} = \frac{\log (1-( 1-e_1)( 1-e_2)(1-e)^{s-2} ) }{\log ( 1-(1-e)^s)}.
\end{equation}
Figure~\ref{fig:extent_of_savings_e1}(b) dipicts even greater savings with this doubly nested approach. With $s$ = 3 the process stops here, but with $s$ = 5, nesting can be done two additional steps. Table~\ref{tab:nested_RANSAC_time_profiling} captures the savings for the TUM-RGBD~\cite{TUMRGBD_Dataset} dataset.

\section{ Ground-Truth Correspondences}
\label{sec:GT}

\begin{figure}[H]
    \centering
    (a)\includegraphics[scale=0.18]{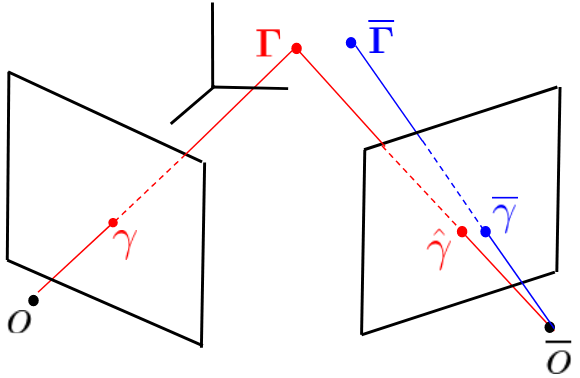}
    (b)\includegraphics[scale=0.18]{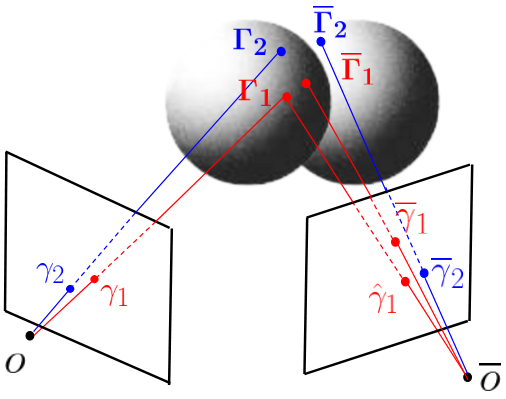}
    \caption{(a) The construction of ground truth correspondence requires both a comparison of the projection error of feature $\gamma$ as $\widehat{\gamma}$ and the putative correspondence $\overline{\gamma}$ as well as a comparison of depth $\widehat{\rho}$ with $\overline{\rho}$. (b) The depth value is unstable near occluding contours due to large depth gradient as in $\Gamma_1$ and due to crossing over the occluding contour, as in $\Gamma_2$.}
    \label{fig:consistency_GT}
\end{figure}

Datasets constructed for the evaluation of RGBD pose estimation generally contain the ground-truth (GT) relative pose between pairs of cameras, but they do not explicitly indicate GT for the correspondences between their image points nor the resulting 3D points. 
The construction of such a ground-truth, however, seems straightforward: 
since the depth value for each feature $\gamma$ in the first image is available, it can be projected onto the second image as $\widehat{\gamma}$, so that the corresponding point $\overline{\gamma}$ can be identified, Figure~\ref{fig:consistency_GT} (a).

\begin{figure}[t]
    \centering
    \includegraphics[width=.15\textwidth]{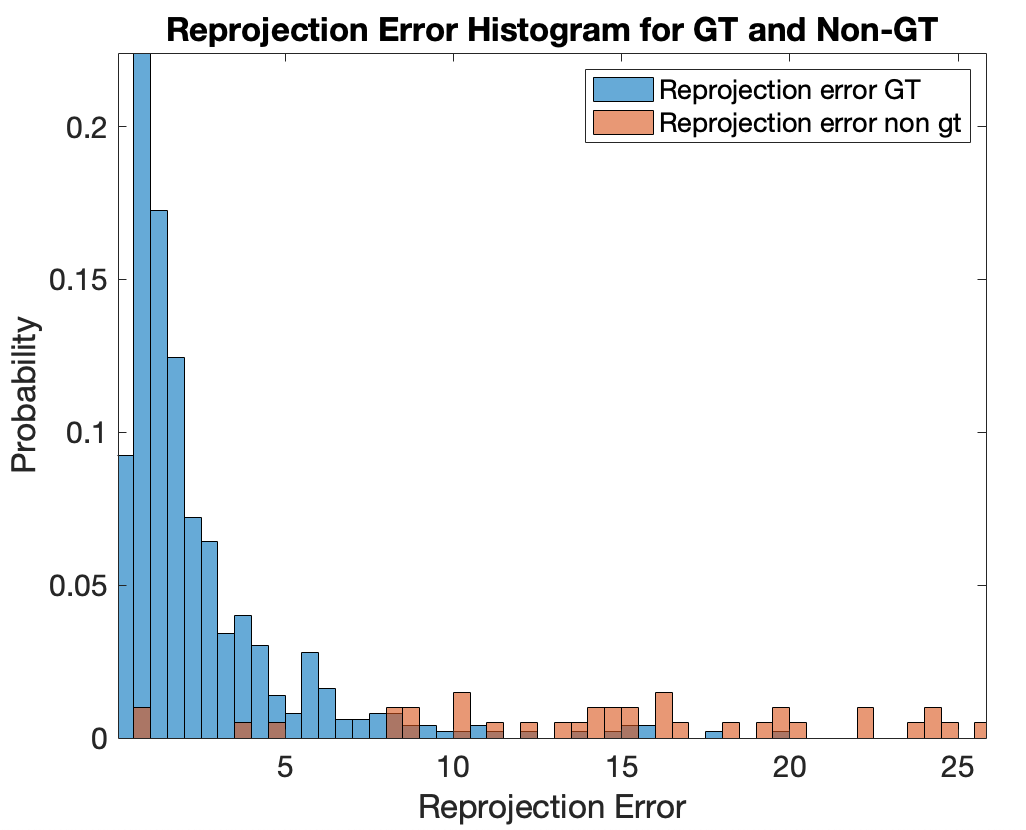}
    \includegraphics[width=.15\textwidth]{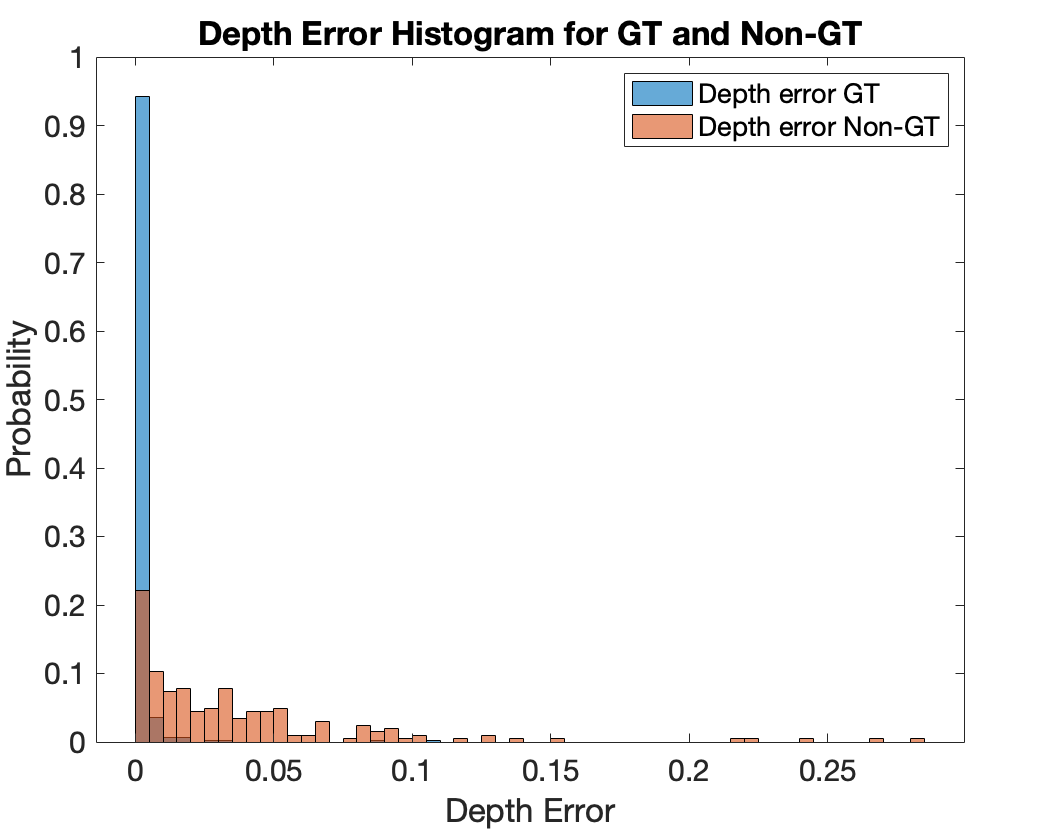}
    \includegraphics[width=.15\textwidth]{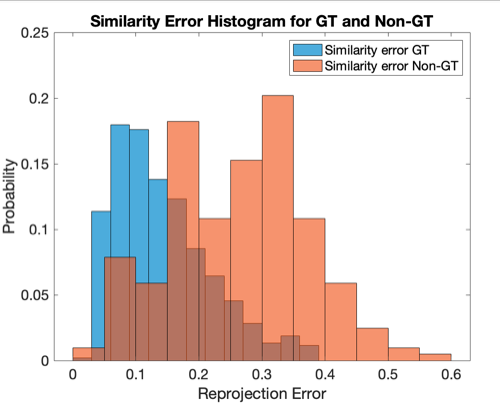}
    \caption{The distribution of reprojection error (a), depth error (b), and similarity error for valid (blue) and invalid (red) correspondences shows that thresholds of $\tau_{\gamma}$ = 8 (pixels), $\tau_{\rho}$ = 0.01 (m), and $\tau_s$ = 0.4 largely differentiate between the two groups. Note that the distribution of non-GT correspondences in (a) continues well into distances of 500 not shown here.}
    \label{fig:hist_gt_non_gt}
\end{figure}

In practice, however, due to feature localization and relative pose errors, a feature $\gamma$ is placed in the vicinity of its corresponding feature $\overline{\gamma}$, forcing a threshold on the distance $|\overline{\gamma}-\widehat{\gamma}|$ between a reprojected point $\widehat{\gamma}$ and the closest corresponding point $\overline{\gamma}$, 
to differentiate between veridical and non-veridical correspondence. The distribution of distances for GT and non-GT correspondences in Figure~\ref{fig:hist_gt_non_gt} (a) shows a trade-off in the selection of this threshold: the smaller the threshold, the larger the confidence in the correspondences and simultaneously the larger the chance of missing some veridical correspondences. The larger the threshold, the smaller the chance of missing veridical correspondences, but simultaneously admitting more false correspondences which coincidentally fall in the neighborhood of $\widehat{\gamma}$.

The dilemma can be resolved by choosing a threshold that discards the vast majority of invalid correspondences while discarding as few true correspondences as possible, \emph{e.g.}, $\tau_{\gamma}=8$ pixels, Figure~\ref{fig:hist_gt_non_gt} (a), and instead relying on depth consistency to further distinguish valid and invalid correspondences. Specifically, 
depth values $\widehat{\rho}$ and $\overline{\rho}$ must be close, {\em i.e.,} $\Gamma$ and $\overline{\Gamma}$ should be close, thus requiring both spatial proximity and depth similarity, 
\begin{equation} \label{eq:gt}
|\overline{\gamma}-\widehat{\gamma}|<\tau_\gamma \quad \text{and} \quad
|\overline{\rho}-\widehat{\rho}|<\tau_\rho,
\end{equation}
where $\tau_{\gamma}$ and $\tau_{\rho}$ are thresholds of distance in the image plane and depth differences, respectively. 
The distributions of depth for valid and invalid correspondences , Figure~\ref{fig:hist_gt_non_gt}(b), suggests a threshold of $\tau_{\rho}$ = 0.01 (m) to discard the vast majority of non-veridical correspondences while not discarding many veridical ones. The direct comparison of depth values, however, does not take into account that depth errors are proportional to depth so that a more appropriate depth similarity constraint is,
\begin{equation} \label{eq:gt2}
    |\overline{\gamma}-\widehat{\gamma}|<\tau_\gamma \quad \text{and} \quad
    2\frac{|\overline{\rho}-\widehat{\rho}|}{|\overline{\rho}+\widehat{\rho}|}<\tau_\rho.
\end{equation}

The above approach establishes correspondences well in general. However, variations of non-planar surfaces, especially occurring near occluding contours, {\em e.g.,} features $\gamma_1$ and $\gamma_2$ shown in Figure~\ref{fig:consistency_GT} (b), where
the depth gradient at $\gamma_1$ is larger than that predicted by linear depth variation. Thus, veridical correspondence $(\gamma_1,\overline{\gamma}_1)$ may not satisfy the depth proximity constraint of Equation~\ref{eq:gt2}. If a putative correspondence satisfies Equation~\ref{eq:gt2}, it is considered a veridical correspondence, but otherwise further examination is required: Denote the depth variation of $\widehat{\gamma}$ within the neighborhood of $\gamma$ bounded by $\tau_\gamma$, by $(\widehat{\rho}_{min},\widehat{\rho}_{max})$, and similarly, the depth variations of $\overline{\gamma}$ is denoted as $(\overline{\rho}_{min},\overline{\rho}_{max})$. Now, allowing for variation for both $\gamma$ and $\overline{\gamma}$, the correspondence pair $(\gamma^*,\overline{\gamma}^*)$ with the closest depths is found, {\em i.e.,} 
\begin{equation}
\left \{
\begin{aligned}
    \overline{\rho}^* &= \overline{\rho}_{max}, & \widehat{\rho}^* &= \widehat{\rho}_{min} & \text{if ~} \overline{\rho}_{max} < \widehat{\rho}_{min} \\
    \overline{\rho}^* &= \overline{\rho}_{min}, & \widehat{\rho}^* &= \widehat{\rho}_{max} & \text{if ~} \overline{\rho}_{min} < \widehat{\rho}_{max} \\
    \multicolumn{2}{l}{$\overline{\rho}$ - $\widehat{\rho}$ = 0} & & & \text{otherwise} 
\end{aligned}
\right.
\label{eq:gt3}
\end{equation}

In such a case, Equation~\ref{eq:gt2} can be tested for $(\overline{\rho}^*,\widehat{\rho}^*)$ and if satisfied, the correspondence can be considered veridical. Note that a potential correspondence, say $(\gamma_2,\overline{\gamma}_2)$ in Figure~\ref{fig:consistency_GT}(b) which lie on distinct surfaces have spatial proximity, but they cannot be accepted as a valid correspondence. 

Finally, the extent of feature similarity of putative correspondences can also be used. Figure~\ref{fig:hist_gt_non_gt} (c) shows that while these distributions for GT and non-GT correspondences are broadly overlapping, the slight shift between the two enables a certain degree of differentiation that discards some invalid matches, \emph{e.g.}, with a similarity threshold of $\tau_{s}$ = 0.4.

\begin{figure}[!htbp]
    \centering
    \includegraphics[width=.15\textwidth]{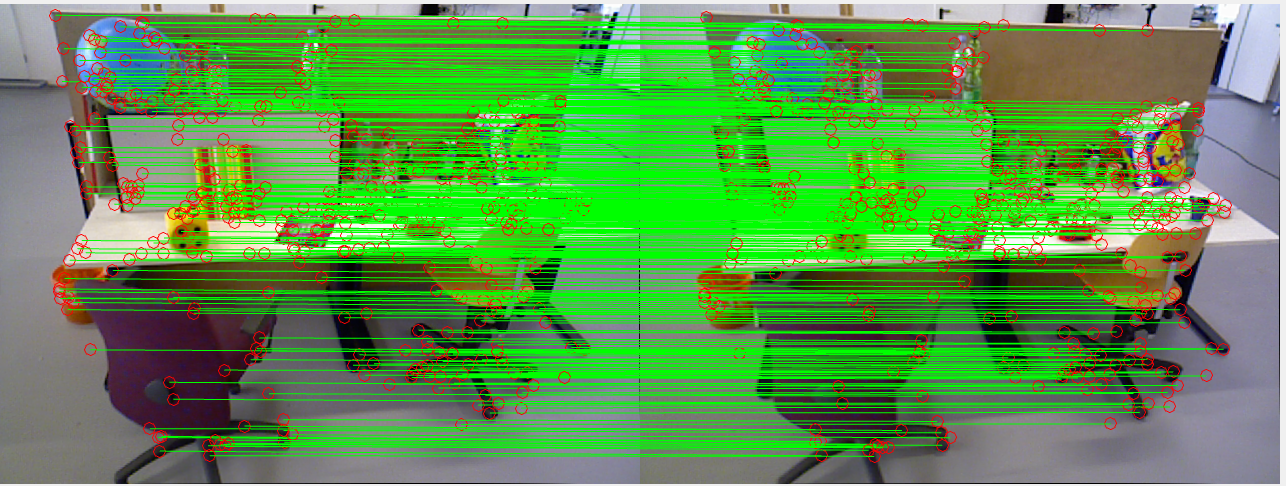}
    \includegraphics[width=.15\textwidth]{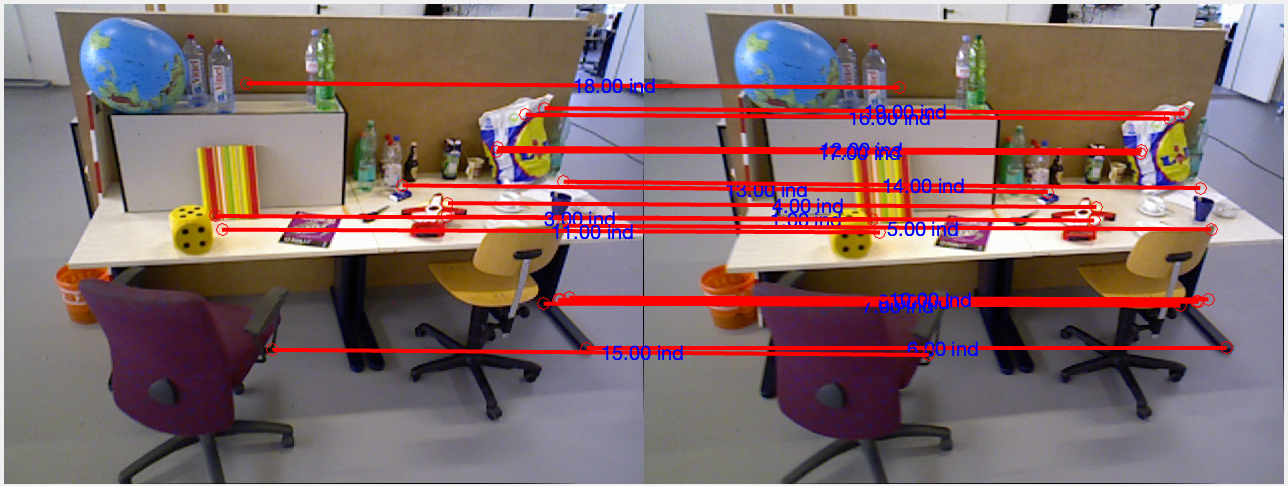}
    \includegraphics[width=.15\textwidth]{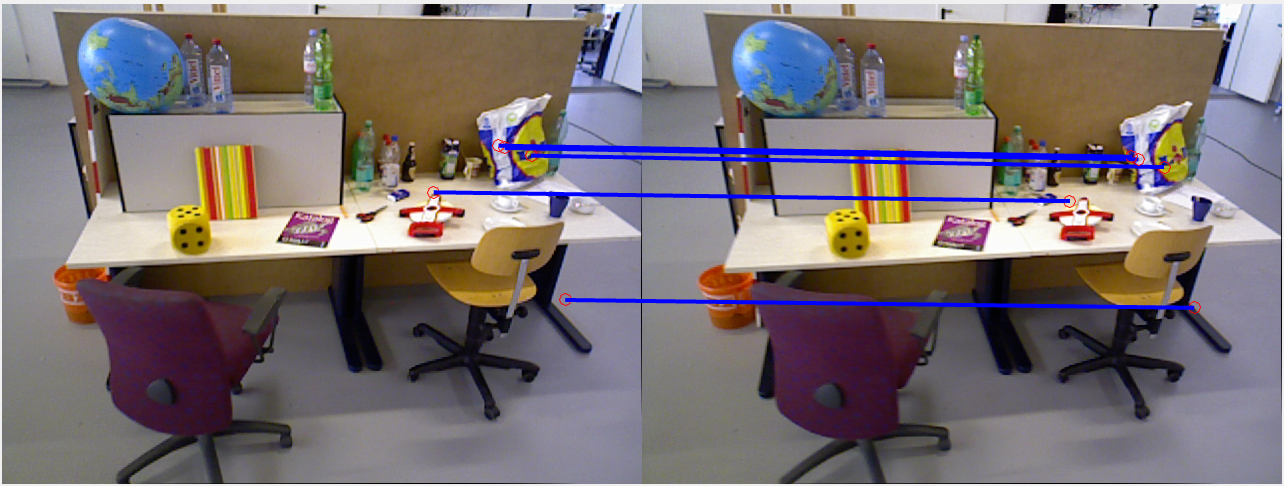}
    \\
    \includegraphics[width=.15\textwidth]{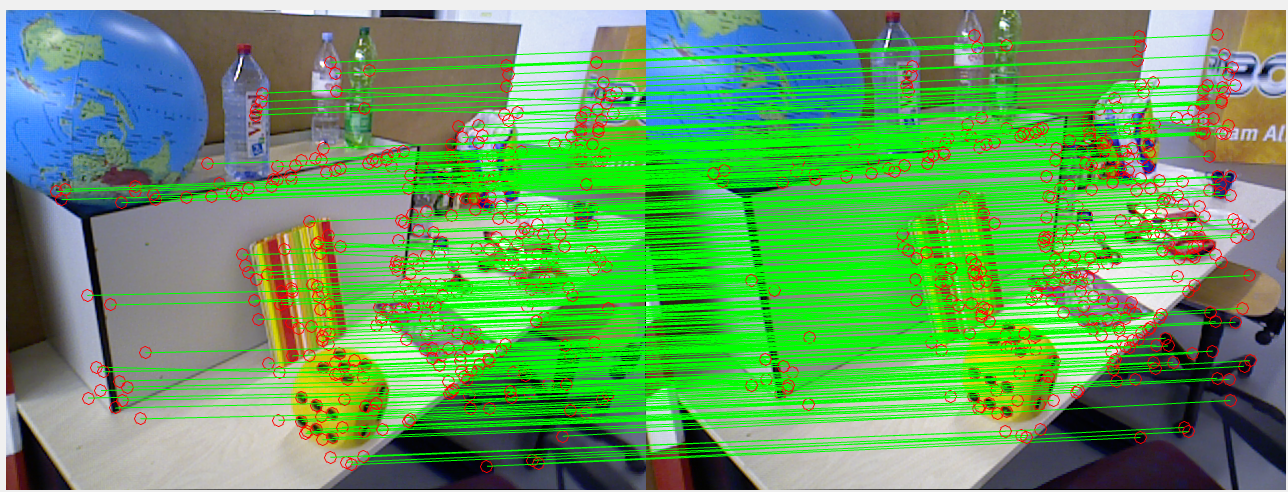}
    \includegraphics[width=.15\textwidth]{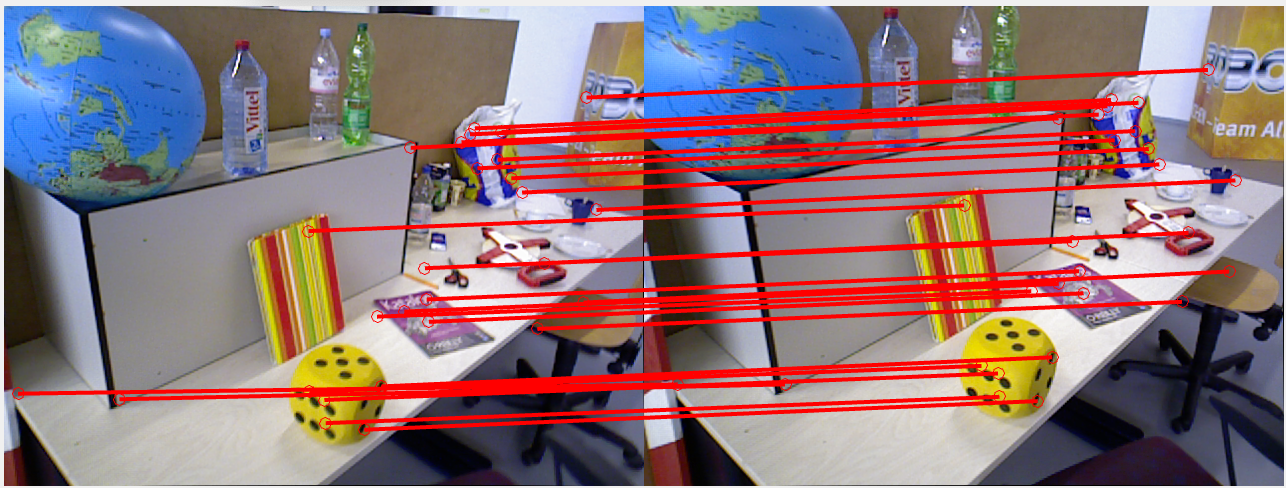}
    \includegraphics[width=.15\textwidth]{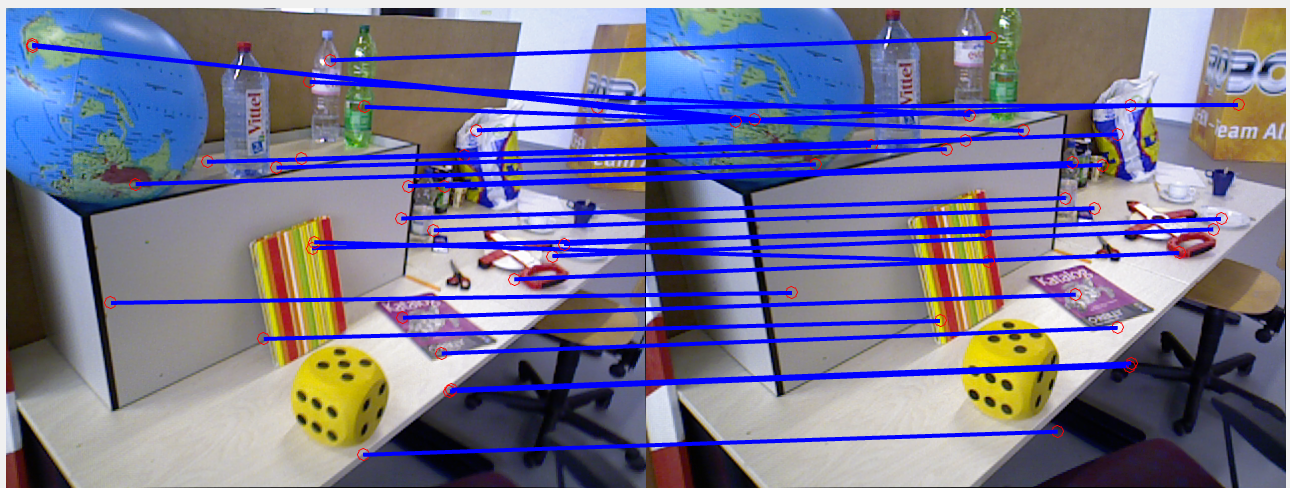}
    \\
    \caption{Ground-truth correspondences on the small dataset of five images which are manually labeled compared to algorithmic GT showing TP (green), FN (red), and FP (blue). More results can be found in the supplementary materials.}
    \label{fig:gt_tp_pic}
\end{figure}

The algorithm then relies on three cues to establish GT correspondences for standard datasets such as TUM-RGBD~\cite{TUMRGBD_Dataset} which does not have correspondence GT. The proposed algorithmic GT needs to be validated against manual GT. 
Five pairs of images were randomly selected, the putative feature correspondences were manually examined, and their labels were corrected so that each feature in each image was either identified as having a corresponding feature or having none. Table~\ref{tab:table_comparing_algo_gt} evaluates the algorithm's determination of GT against the manually determined GT for each of the five images.
Figure~\ref{fig:gt_tp_pic} visually illustrates the quality of the algorithmic GT with TP, FN, and FP correspondence shown in green, red, and blue, respectively. Theses results indicate that the algorithmic GT proposed here is a suitable surrogate for the manual GT.

%> Default is 6pt
\setlength\tabcolsep{1.3pt} 
\begin{table}[t]
{\footnotesize
\begin{tabular}{|c|c|c|}
    \hline
    & \textbf{T} & \textbf{F} \\
    \hline
    \textbf{T} & 1068 & 400\\
    \hline
    \textbf{F} & 116 & 368 \\
    \hline
    \hline
    & \textbf{T} & \textbf{F} \\
    \hline
    \textbf{T} &1165  & 5\\
    \hline
    \textbf{F} & 19  & 763\\
    \hline
    \multicolumn{3}{c}{(a)}
\end{tabular}
\begin{tabular}{|c|c|c|}
    \hline
    & \textbf{T} & \textbf{F} \\
    \hline
    \textbf{T} & 1456 & 774 \\
    \hline
    \textbf{F} & 144 &455\\
    \hline
    \hline
    & \textbf{T} & \textbf{F} \\
    \hline
    \textbf{T} & 1567 & 15\\
    \hline
    \textbf{F} & 27 & 1214\\
    \hline
    \multicolumn{3}{c}{(b)}
\end{tabular}
% \hspace{0.4em}
\begin{tabular}{|c|c|c|}
    \hline
    & \textbf{T} & \textbf{F} \\
    \hline
    \textbf{T} & 1132 &558 \\
    \hline
    \textbf{F} & 164 &828 \\
    \hline
    \hline
    & \textbf{T} & \textbf{F} \\
    \hline
    \textbf{T} & 1254 & 23\\
    \hline
    \textbf{F} & 42 &1363 \\
    \hline
    \multicolumn{3}{c}{(c)}
\end{tabular}
% \hspace{0.1em}
\begin{tabular}{|c|c|c|}
    \hline
    & \textbf{T} & \textbf{F} \\
    \hline
    \textbf{T} & 1120 &696 \\
    \hline
    \textbf{F} & 176&561\\
    \hline
    \hline
    & \textbf{T} & \textbf{F} \\
    \hline
    \textbf{T} & 1278 & 16\\
    \hline
    \textbf{F} & 18 &1241 \\
    \hline
    \multicolumn{3}{c}{(d)}
\end{tabular}
\begin{tabular}{|c|c|c|}
    \hline
    & \textbf{T} & \textbf{F} \\
    \hline
    \textbf{T} & 1368 &730 \\
    \hline
    \textbf{F} & 142&651\\
    \hline
    \hline
    & \textbf{T} & \textbf{F} \\
    \hline
    \textbf{T} & 1481 & 12\\
    \hline
    \textbf{F} & 29 &1369 \\
    \hline
    \multicolumn{3}{c}{(e)}
\end{tabular}
\caption{Two methods of establishing GT correspondences are evaluated against manual ground-truth for five image pairs randomly selected from the TUM-RGBD~\cite{TUMRGBD_Dataset} dataset. The top row shows confusion matrices for similarity-based correspondences where the number of features in image one, the number of features in image two, and the number of correspondences obtained by thresholding similarity at $\tau_s$ = 0.8 are
(a) (986,966,1468), (b) (1511,1318,1115) (c) (1179,1503,845)) (d) (1342,1211,908) (e) (1472,1419,1049). Observe the large number of false positives (FP) and false negatives (FN) which prevent this approach from being used as a suitable algorithmic GT for evaluating correspondences. The bottom row evaluates the triple-cue algorithmic GT proposed here depicting a very small number of FP and FN. } 
\label{tab:table_comparing_algo_gt}
}
\end{table}
%> Default is 6pt
\setlength\tabcolsep{1pt} 
\begin{table*}[t]
\centering
{\footnotesize
    \begin{tabular}{|c|c|c|c|c|c|c|c|c|c|c|c|c|c|c|c|c|c|c|c|c|}
    \hline
    & \multicolumn{4}{!{\vrule width 2pt}c|}{$e$ = 60-70$\%$} & \multicolumn{4}{!{\vrule width 2pt}c|}{$e$ = 70-80$\%$} &\multicolumn{4}{!{\vrule width 2pt}c|}{$e$ = 80-90$\%$} &\multicolumn{4}{!{\vrule width 2pt}c|}{$e$ = 90-95$\%$} &\multicolumn{4}{!{\vrule width 2pt}c|}{$e$ = 95-99$\%$} \\
    \cline{2-21}
    & \multicolumn{1}{!{\vrule width 2pt}c|}{{\rotatebox{90}{\textbf{Classic}}}} & {\rotatebox{90}{\textbf{GDC}}} & {\rotatebox{90}{\textbf{\makecell{GDC\\Nested}}}} & {\rotatebox{90}{\textbf{\makecell{GDC\\ Doubly\\ Nested}}\;}} & \multicolumn{1}{!{\vrule width 2pt}c|}{{\rotatebox{90}{\textbf{Classic}}}} & {\rotatebox{90}{\textbf{GDC}}} & {\rotatebox{90}{\textbf{\makecell{GDC\\Nested}}}} & {\rotatebox{90}{\textbf{\makecell{GDC\\ Doubly\\ Nested}}\;}} & \multicolumn{1}{!{\vrule width 2pt}c|}{{\rotatebox{90}{\textbf{Classic}}}} & {\rotatebox{90}{\textbf{GDC}}} & {\rotatebox{90}{\textbf{\makecell{GDC\\Nested}}}} & {\rotatebox{90}{\textbf{\makecell{GDC\\ Doubly\\ Nested}}\;}} & \multicolumn{1}{!{\vrule width 2pt}c|}{{\rotatebox{90}{\textbf{Classic}}}} & {\rotatebox{90}{\textbf{GDC}}} & {\rotatebox{90}{\textbf{\makecell{GDC\\Nested}}}} & {\rotatebox{90}{\textbf{\makecell{GDC\\ Doubly\\ Nested}}\;}} & \multicolumn{1}{!{\vrule width 2pt}c|}{{\rotatebox{90}{\textbf{Classic}}}} & {\rotatebox{90}{\textbf{GDC}}} & {\rotatebox{90}{\textbf{\makecell{GDC\\Nested}}}} & {\rotatebox{90}{\textbf{\makecell{GDC\\ Doubly\\ Nested}}\;}} \\
    \hline
    \makecell{$\#$ of matches from \\ top rank-ordered \\ list: $M_1$/$M_2$/$M_3$} & \multicolumn{1}{!{\vrule width 2pt}c|}{250} & 250 & \makecell{100/\\250} & \makecell{100/\\150/\\250} & \multicolumn{1}{!{\vrule width 2pt}c|}{250} & 250 & \makecell{100/\\250} & \makecell{100/\\150/\\250} & \multicolumn{1}{!{\vrule width 2pt}c|}{250} & 250 & \makecell{100/\\250} & \makecell{100/\\150/\\250}  & \multicolumn{1}{!{\vrule width 2pt}c|}{250} & 250 & \makecell{100/\\250} & \makecell{100/\\150/\\250} & \multicolumn{1}{!{\vrule width 2pt}c|}{250} & 250 & \makecell{100/\\250} & \makecell{100/\\150/\\250} \\
    \hline
    \makecell{$\#$ of RANSAC \\ iterations \\ (99$\%$ success rate)} & \multicolumn{1}{!{\vrule width 2pt}c|}{169} & \makecell{169\\$\downarrow$\\44} & \makecell{146\\$\downarrow$\\35} & \makecell{137\\$\downarrow$\\32} & \multicolumn{1}{!{\vrule width 2pt}c|}{420} & \makecell{420\\$\downarrow$\\85} & \makecell{371\\$\downarrow$\\76} & \makecell{227\\$\downarrow$\\53} & \multicolumn{1}{!{\vrule width 2pt}c|}{3752} & \makecell{3752\\$\downarrow$\\533} & \makecell{2218\\$\downarrow$\\272} & \makecell{2126\\$\downarrow$\\240} & \multicolumn{1}{!{\vrule width 2pt}c|}{21375} & \makecell{21375\\$\downarrow$\\876}& \makecell{17509\\$\downarrow$\\916} & \makecell{6872\\$\downarrow$\\340} & \multicolumn{1}{!{\vrule width 2pt}c|}{681274} & \makecell{681274\\$\downarrow$\\14374}& \makecell{220637\\$\downarrow$\\9708} & \makecell{68824\\$\downarrow$\\1446} \\
    \hline
    \hline
    \rowcolor{yellow}{\textbf{Total Cost (ms)}} & \multicolumn{1}{!{\vrule width 2pt}c|}{7.8} & 2.9 & 2.4 & 2.2 & \multicolumn{1}{!{\vrule width 2pt}c|}{19.4} & 6.2 & 5.5 & 3.7 & \multicolumn{1}{!{\vrule width 2pt}c|}{173.2} & 45.1 & 24.7 & 22.8 & \multicolumn{1}{!{\vrule width 2pt}c|}{986.7} & 155.5 & 139.6 & 53.9 & \multicolumn{1}{!{\vrule width 2pt}c|}{31447.6} & 3822.6 & 1676.6 & 451.5 \\
    \hline
    \end{tabular}
    \caption{A comparison of timings for classic RANSAC, GDC-Filtered RANSAC, nested RANSAC, and doubly nested RANSAC for the TUM-RGBD dataset. Note that the change in the number of RANSAC iterations indicates the number of hypothesis passing the GDC test. Experimental settings are identical to Tables~\ref{tab:GDC_Profiling} and~\ref{tab:nested_RANSAC_time_profiling}. Experiments for other datasets are given in the supplementary materials.} 
    \label{tab:GDC_nested_RANSAC_time_profiling}
}
\end{table*}

\section{Experiments}

Tables~\ref{tab:GDC_Profiling} and \ref{tab:nested_RANSAC_time_profiling} in previous sections show significantly improved speedup over the classic RANSAC. This section demonstrates \emph{(i)} the efficiency from the combined GDC filtered RANSAC and the nested RANSAC, Table~\ref{tab:GDC_nested_RANSAC_time_profiling}; \emph{(ii)} improvements in accuracy supported by the comparisons with the existing methods.

\noindent \textbf{Datasets:} 
Evaluations are benchmarked using TUM-RGBD~\cite{TUMRGBD_Dataset}, ICL-NUIM~\cite{handa2014benchmark}, and RGBD Scenes v2~\cite{RGBD_Scenes_v2_Dataset} datasets. Details of the selected sequences for each dataset are given in the supplementary materials.

\noindent \textbf{Metrics:} The relative pose error (RPE)~\cite{zhang2018tutorial} measures both rotation and translation drifts of one frame $n$ with respect to another frame $n-\Delta$, where $\Delta$ is the number of frames apart. $\Delta = 1$ in our experiments, if otherwise specified. 
% (typically, $\Delta=1$).
% For a small $\Delta$, \emph{i.e.}, frames that are spatially close, RPE reflects the local estimation consistency. In contrast, for a large $\Delta$, it reflects the degree of the drift in a long camera movement. 

\setlength\tabcolsep{1pt}
\begin{table*}[!htbp]
\centering
\footnotesize
\begin{tabular}
{lcc|cccccccc|cccc}
 & {\rotatebox{45}{fr1/desk}} & {\rotatebox{45}{fr3/office}} & {\rotatebox{45}{lr kt0}} & {\rotatebox{45}{lr kt1}} & {\rotatebox{45}{lr kt2}} & {\rotatebox{45}{lr kt3}} & {\rotatebox{45}{of kt0}} & {\rotatebox{45}{of kt1}} & {\rotatebox{45}{of kt2}} & {\rotatebox{45}{of kt3}} & {\rotatebox{45}{s05}} & {\rotatebox{45}{s06}} & {\rotatebox{45}{s07}} & {\rotatebox{45}{s08}} \\
\toprule
% \multirow{2}{*}{Methods} & \multicolumn{2}{c}{TUM RGBD} & \multicolumn{2}{c}{ICL-NUIM} & \multicolumn{2}{c}{OpenLORIS} \\
% & ATE & Rot. \& Trans. Errors & ATE & Rot. \& Trans. Errors & ATE & Rot. \& Trans. Errors \\
{Methods} & \multicolumn{2}{c|}{TUM RGBD} & \multicolumn{8}{c|}{ICL-NUIM} & \multicolumn{4}{c}{RGBD Scenes v2} \\
\midrule
ORB SLAM2~\cite{mur2017orb}   &2.00 & \underline{0.83}& \underline{4.29}& 28.07 & 9.68 & 14.35 & 6.00 & 16.53 & 6.40 & 25.42 &\underline{4.37} &\underline{3.89} &\underline{2.40} &\underline{3.76} \\
% BiDirectional DVO~\cite{cai2020bi} & - & - &0.86 & 1.94 & - & - & 5.61 & 2.83 & - & - & -& - & - & -\\
CVO~\cite{ghaffari2019continuous} &2.09 &3.74 & 7.71 & 2.68&4.63 &32.58 &11.14 &12.37 &5.64 &15.63 &20.9 &30.8 &33.52 &37.75  \\
ACO~\cite{lin2019adaptive}  & 10.13& $\times$ &7.92 &1.77 &4.10 &33.59 &10.8 &11.05 &5.87 &15.75 &22.12 &35.62  &34.35 &33.73 \\
RGBD DVO~\cite{cai2021direct} & 1.3 & $\times$ &$-$ &\underline{0.78} &3.28 &3.30 &1.27 &\underline{0.77} &2.65 &2.07 &11.36 &15.53 &12.40  &11.79 \\
KinectFusion~\cite{izadi2011kinectfusion}   &34.43 &21.32 & 32.17&10.05 &5.30 &32.46 &17.5 &29.34 &28.44 &42.45 &178.67 &177.69 &173.94 &165.87 \\
Edge DVO~\cite{christensen2019edge} &17.32 &1.04 & $\times$ & 1.51 & 3.68 & $\times$ &1.95 & $\times$ &\underline{2.46} & \underline{1.14} & - &- &- & -\\
Canny VO~\cite{zhou2018canny} &5.1 & 1.9 & - & 0.9 & \underline{1.1} & \underline{0.7} & - & - & - & - & - & - & - & - \\
RGBD DSO$^\dag$~\cite{yuan2021rgb}   & \textbf{0.12} & \textbf{0.56} & - & - & - & - & - & - & - & -  &5.76 &39.18 &2.88 &5.56 \\
%GG-VO~\cite{yao2023rgb} & & & & & &  &  & &  &  & &  & & \\
% PLP-SLAM~\cite{shu2023structure} & & & & & &  &  & &  &  & &  & & \\
%CVO-SLAM~\cite{lin2023robust} & & & & & &  &  & &  &  & &  & & \\
% OVD-SLAM~\cite{he2023ovd} & & & & & &  &  & &  &  & &  & & \\
% RGBD DSO$^\ddag$ ~\cite{yuan2021rgb}   &\textbf{0.12} &\textbf{0.48} &\textbf{0.28} &\textbf{0.39} &1.12 &7.09 &1.58 &\textbf{1.06} &1.75 &3.49 &3.87 &2.29 &2.88 \\
Our Method  & \underline{1.19} & 0.86 & {\bf 0.37}&\textbf{0.39} &\textbf{0.38} &{\bf 0.35}&\textbf{0.58} & {\bf 0.52}&\textbf{2.3} & \textbf{0.44}& \textbf{0.96}& \textbf{1.04}& \textbf{1.02}& \textbf{1.07}\\
\midrule
\multicolumn{14}{l}{\textbf{Boldfaced:} the best. \quad \underline{Underlined:} the second best. \quad -: Result not available from the original paper. } \\
\multicolumn{14}{l}{$\times$: Failure to complete the entire sequence. \quad $^\dag$: Disable depth refinement and occlusion removal modules. } \\
\bottomrule
\end{tabular}
\caption{$\text{RPE}_\text{trans}$ (cm) comparisons of our method against contemporary RGBD VO/SLAM pipelines.}
\label{tab:RPE_trans_comparison}
\end{table*}

\setlength\tabcolsep{1pt}
\begin{table*}[!htbp]
\centering
\footnotesize 

\begin{tabular}
{lcc|cccccccc|cccc}
 & {\rotatebox{45}{fr1/desk}} & {\rotatebox{45}{fr3/office}} &   {\rotatebox{45}{lr kt0}}
 &{\rotatebox{45}{lr kt1}} & {\rotatebox{45}{lr kt2}} & {\rotatebox{45}{lr kt3}} & {\rotatebox{45}{of kt0}} & {\rotatebox{45}{of kt1}} & {\rotatebox{45}{of kt2}} & {\rotatebox{45}{of kt3}} & {\rotatebox{45}{s05}} & {\rotatebox{45}{s06}} & {\rotatebox{45}{s07}} & {\rotatebox{45}{s08}} \\
\toprule
{Methods} & \multicolumn{2}{c|}{TUM RGBD} & \multicolumn{8}{c|}{ICL-NUIM} & \multicolumn{4}{c}{RGBD Scenes v2} \\
\midrule
ORB SLAM2~\cite{mur2017orb}  &0.94&1.25 &5.61& $\times$ &2,37 &3.22 &0.93 & 2.46& 2.90&6.58 &1.54 &\underline{1.26} &{\bf 0.96} &\underline{1.08}  \\
% BiDirectional DVO~\cite{cai2020bi} & - &-& - & - & - & - & - & - & - & -& -& -& -& -\\
CVO~\cite{ghaffari2019continuous}   &0.76 &1.53 &\underline{2.11}&1.36 &2.13 &6.43 &2.89 &3.49 &2.39 &5.86 &8.28 &12.30 &13.03 &15.05 \\
ACO~\cite{lin2019adaptive}  & 0.72& 1.54&2.88 &1.23 &1.91 &4.79 &2.53 &3.35 &2.52 &7.16 &8.63 &14.20 &13.49  &13.64 \\
RGBD DVO~\cite{cai2021direct} & 1.75&8.87 & - &\underline{0.17} & 0.91& 0.56&0.24 &0.26 &\underline{1.03} &0.34 &4.21 &5.96 &4.83 &4.44 \\
KinectFusion~\cite{izadi2011kinectfusion}   &3.09 &8.00 & 9.12&1.20&1.37&9.98&1.16&1.23 &2.93&1.16 &80.77 &86.98&81.97 &77.71 \\

Edge DVO~\cite{christensen2019edge} &15.17 & 0.56 & $\times$& 0.18& \underline{0.12} & $\times$ &\textbf{0.16}& $\times$ & {\bf 0.36}&0.17 &- &- &- & -\\
Canny VO~\cite{zhou2018canny} & 2.393 & 0.906 &- & 0.208 &0.269 &{\bf 0.152} & -& -&- &- &- &-&-&- \\
RGBD DSO$^\dag$ ~\cite{yuan2021rgb}   &{\underline{0.32}} &{\bf 0.23} &-  &-  &- &- &-&- &-&- &\underline{1.29} &8.95  &\underline{1.02} &1.60 \\
%GG-VO~\cite{yao2023rgb} & & & & & &  &  & &  &  & &  & & \\
% PLP-SLAM~\cite{shu2023structure} & & & & & &  &  & &  &  & &  & & \\
%CVO-SLAM~\cite{lin2023robust} & & & & & &  &  & &  &  & &  & & \\
% OVD-SLAM~\cite{he2023ovd} & & & & & &  &  & &  &  & &  & & \\
%NICE-SLAM~\cite{zhu2022nice}  & & & & & &  &  & &  &  & &  & & \\
% RGBD DSO$^\ddag$~\cite{yuan2021rgb}   &\textbf{0.23} &\textbf{0.32} &\textbf{0.04} &\textbf{0.06} &\textbf{0.06} &2.32 &\textbf{0.19} &\textbf{0.13} &\textbf{0.09} &\textbf{1.26} &\textbf{1.01} &\textbf{0.9} &0.95 \\
Our Method  & { \textbf{0.57}} & { \underline{0.34}}&{\bf 0.14}&{\bf 0.09} &{\bf 0.10}
& { \underline{0.16}}& \underline{0.32}& \textbf{0.15} &1.84 &\textbf{0.12}& {\bf 1.11}& {\bf 1.14} &{1.11}& \textbf{1.08}\\
\midrule
\multicolumn{14}{l}{\textbf{Boldfaced:} the best. \quad \underline{Underlined:} the second best. \quad -: Result not available from the original paper. } \\
\multicolumn{14}{l}{$\times$: Failure to complete the entire sequence. \quad $^\dag$: Disable depth refinement and occlusion removal modules. } \\
\bottomrule
\end{tabular}

\caption{$\text{RPE}_\text{rot}$ (degree) comparisons of our method against contemporary RGBD VO/SLAM pipelines.}
\label{tab:RPE_rot_comparison}
\end{table*}

\noindent \textbf{Comparison with Other Methods}: 
Tables~\ref{tab:RPE_trans_comparison} and~\ref{tab:RPE_rot_comparison} demonstrate comparisons of our method against several contemporary RGBD VO/SLAM pipelines. The back-end optimization of ORB-SLAM2 is disabled so that only its VO mode is used for comparison. The two tables show that our GDC-filtered RANSAC in conjunction with the nested RANSAC for RGBD relative pose estimation delivers comparable or superior results in the almost all the sequences in the tables. Notably, even though the depth refinement and occlusion removal modules are disabled for RGBD DSO, pose refinement is still supported. Nevertheless, our method provides orders of magnitude accuracy improvements in the RGBD Scene v2 dataset. As RGBD Scene v2 dataset exhibits higher outlier ratio in scenes compared to other datasets, the proposed GDC and nested RANSAC effectively contribute to robust pose estimations. More experimental results are provided in the supplementary materials. 
% KinectFusion~\cite{izadi2011kinectfusion} and ORB-SLAM~\cite{mur2017orb} which serve as baselines for many methods; DVO\cite{cai2020bi} which is recognized for its excellence in direct dense RGBD VO; CVO~\cite{ghaffari2019continuous} and ACO~\cite{lin2019adaptive} stands out for two state-of-the-art continuous RGBD VO frameworks. RGBD DSO~\cite{yuan2021rgb} is also employed for comparison. 

% \newline
% \noindent \textbf{Conclusion: } 

\section{Conclusion}
This paper proposes a RGBD relative pose estimation approach using \emph{(i)} a filter RANSAC from geometric depth consistency (GDC) constraint to avoid computing hypotheses from outliers, and \emph{(ii)} a nested RANSAC which picks correspondences from different ranking levels to increase the likelihood of computing hypotheses from inliers. A combination of both techniques facilitates significant speedup over classic RANSAC scheme, enabling using large RANSAC iterations without the cost of losing efficiency. Thus, the proposed approach outperforms other methods, especially in very high outlier ratio scenarios.

\maketitlesupplementary

\setcounter{section}{0}
\setcounter{figure}{0}
\setcounter{equation}{0}
\setcounter{table}{0}
\section{Proof of Proposition 1}
\begin{figure}[!htbp]
    \centering
    \includegraphics[width=.45\textwidth]{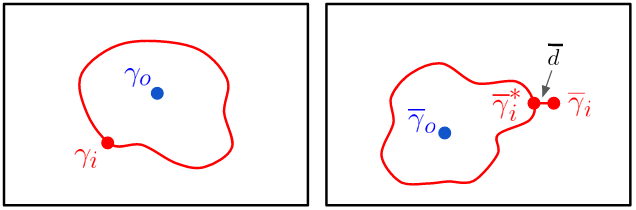}
    \caption{The geometric depth consistency constrains a correspondence $(\gamma_i,\rho_i)$ and $(\overline{\gamma}_i,\overline{\rho}_i)$ to lie on the corresponding level-sets of $\phi$ and $\overline{\phi}$ constructed with respect to a reference point correspondence $(\gamma_0,\rho_0)$ and $(\overline{\gamma}_0,\overline{\rho}_0)$. Due to noise in feature location and depth measurement, the observed correspondence $\overline{\gamma}_i$ is a perturbation of the true corresponding point $\overline{\gamma}_i^{*}$ which must lie on the level-set $\overline{\phi}$. The distance $\overline{d}$ is the extent of this perturbation. }
    \label{fig:fig_proof_of_prop1}
\end{figure}

\setcounter{prop}{0}
\begin{prop}
Let $\phi$ and $\overline{\phi}$ be the squared radial maps of the first and second RGBD images, respectively, with respect to a veridical corresponding reference points $(\gamma_0, \rho_0)$ and $(\overline{\gamma}_0, \overline{\rho}_0)$. Given a putative correspondence between $(\gamma_i, \rho_i)$ and $(\overline{\gamma}_i, \overline{\rho}_i)$, the distance $\overline{d}$ from $\overline{\gamma}_i$ to the curve it must lie on in image 2, Figure~\ref{fig:fig_proof_of_prop1}, is
\begin{equation}
    % d = \frac{|\phi - \overline{\phi}|}{||\nabla \overline{\phi}||},
    \overline{d} = \frac{|\phi - \overline{\phi}|}{2\begin{vmatrix} \left(\overline{\rho}_i \begin{vmatrix}\overline{\gamma}_i\end{vmatrix}^2- \overline{\rho}_0 \overline{\gamma}_0^T \overline{\gamma_i} \right) \nabla \overline{\rho}_i + \overline{\rho}_i \begin{bmatrix} 1\;\;0\;\;0 \\ 0\;\;1\;\;0 \end{bmatrix} (\overline{\rho}_i \overline{\gamma}_i - \overline{\rho}_0 \overline{\gamma}_0) \end{vmatrix}}.
\end{equation}
\end{prop}
\begin{proof}
The true correspondence point, $\overline{\gamma}_i^{*}$, must lie on the corresponding curve to the curve $\gamma_i$ lies on, Figure~\ref{fig:fig_proof_of_prop1}, \emph{i.e.},
\begin{equation}
    \overline{\phi}(\overline{\gamma}_i^{*}) = \phi(\gamma_i).
\label{eq:equal_phi_on_two_images}
\end{equation}
Thus, $\overline{\gamma}_i^{*}$ can be identified as the point on the level-set $\overline{\phi}$ that has the least perturbation from the observed point $\overline{\gamma}_i$, \emph{i.e.},
\begin{equation}
    \overline{d}^2 = 
    \min_{\overline{\gamma},\overline{\phi}(\overline{\gamma}) = \phi(\gamma_i)} d^2 \left(\overline{\gamma}, \overline{\gamma}_i \right).
\label{eq:argmin_distance}
\end{equation}
Denoting $\overline{\gamma}_i=(\overline{\xi}_i,\overline{\eta}_i)$, $\overline{\gamma}_i^{*}=(\overline{\xi}_i^{*},\overline{\eta}_i^{*})$, and $\overline{\gamma}=(\overline{\xi}, \overline{\eta})$, this can be written as
\begin{equation}
    \overline{d}^2 = \min_{(\overline{\xi},\overline{\eta}),\overline{\phi}(\overline{\xi}, \overline{\eta})=\phi(\xi_i,\eta_i)} \left[(\overline{\xi} - \overline{\xi}_i)^2 + (\overline{\eta} - \overline{\eta}_i)^2 \right].
\label{eq:argmin_rewritten}
\end{equation}
Now, since the perturbation of $\overline{\gamma}_i$ is small, a first-order approximation holds, \emph{i.e.},
\begin{equation}
    \overline{\phi} \left(\overline{\xi}, \overline{\eta} \right) \cong \overline{\phi}\left(\overline{\xi}_i, \overline{\eta}_i \right) + \nabla \overline{\phi} \left(\overline{\xi}_i, \overline{\eta}_i \right) \begin{bmatrix} \overline{\xi} - \overline{\xi}_i \\ \overline{\eta} - \overline{\eta}_i \end{bmatrix}.
\label{eq:Taylor_approx}
\end{equation}
Using $\overline{\phi}(\overline{\xi}, \overline{\eta})=\phi(\xi_i, \eta_i)$, this gives one equation in the unknown $(\overline{\xi}, \overline{\eta})$, so that $\overline{\eta}$ can be written in terms of $\overline{\xi}$ by solving
\begin{equation}
    \phi(\xi_i, \eta_i) = \overline{\phi}(\overline{\xi}_i, \overline{\eta}_i) + \overline{\phi}_{\overline{\xi}} (\overline{\xi}_i, \overline{\eta}_i)(\overline{\xi} - \overline{\xi}_i) + \overline{\phi}_{\overline{\eta}} (\overline{\xi}_i, \overline{\eta}_i)(\overline{\eta} - \overline{\eta}_i).
\end{equation}
This gives
\begin{equation}
    (\overline{\eta} - \overline{\eta}_i) = \frac{\phi \left(\xi_i, \eta_i\right) - \overline{\phi}\left(\overline{\xi}_i, \overline{\eta}_i \right) - \overline{\phi}_{\overline{\xi}}(\overline{\xi}_i, \overline{\eta}_i) (\overline{\xi} - \overline{\xi}_i)}{\overline{\phi}_{\overline{\eta}}(\overline{\xi}_i, \overline{\eta}_i)}.
\label{eq:xi_as_a_function_of_eta}
\end{equation}
Thus, the minimization over two variables in Equation~\ref{eq:argmin_rewritten} can be written over a single variable $\overline{\xi}$,
\begin{equation}
\begin{aligned}
    \overline{d}^2 = & \argmin_{\overline{\xi}} \Biggl[ (\overline{\xi}-\overline{\xi}_i)^2 \\ 
    & + \left( \frac{\phi(\xi_i,\eta_i)-\overline{\phi}(\overline{\xi}_i, \overline{\eta}_i) - \overline{\phi}_{\overline{\xi}}(\overline{\xi}, \overline{\eta})(\overline{\xi} - \overline{\xi}_i)}{\overline{\phi}_{\overline{\eta}}(\overline{\xi}, \overline{\eta})} \right)^2 \Biggr] \\
    = & \argmin_{\overline{\xi}} \Biggl[ \left(1 + \frac{\overline{\phi}_{\overline{\xi}}^2 (\overline{\xi}, \overline{\eta})}{\overline{\phi}_{\overline{\eta}}^2 (\overline{\xi}, \overline{\eta})} \right)^2 (\overline{\xi} - \overline{\xi}_i)^2 \\
    & - 2\left(\phi(\xi_i, \eta_i) - \overline{\phi}(\overline{\xi}_i, \overline{\eta}_i) \right) \frac{\overline{\phi}_{\overline{\xi}}(\overline{\xi}_i, \overline{\eta}_i)}{\overline{\phi}_{\overline{\eta}}^2 (\overline{\xi}_i, \overline{\eta}_i)} (\overline{\xi} - \overline{\xi}_i) \\
    & + \left( \frac{\phi(\xi_i, \eta_i) - \overline{\phi}(\overline{\xi}_i, \overline{\eta}_i)}{\overline{\phi}_{\overline{\eta}}(\overline{\xi}_i, \overline{\eta}_i)} \right)^2 \Biggr]
\end{aligned}
\end{equation}
Differentiating this equation with respect to $\overline{\xi}$ and setting to zero gives
\begin{equation}
\begin{aligned}
    & 2\left(1 + \frac{\overline{\phi}_{\overline{\xi}}^2 (\overline{\xi}, \overline{\eta})}{\overline{\phi}_{\overline{\eta}}^2 (\overline{\xi}, \overline{\eta})} \right)(\overline{\xi}^{*} - \overline{\xi}_i) \\
    - & 2\left(\phi(\xi_i, \eta_i) - \overline{\phi}(\overline{\xi}_i, \overline{\eta}_i) \right) \frac{\overline{\phi}_{\overline{\xi}}(\overline{\xi}_i, \overline{\eta}_i)}{\overline{\phi}_{\overline{\eta}}^2 (\overline{\xi}_i, \overline{\eta}_i)} = 0,
\end{aligned}
\end{equation}
so that
\begin{equation}
\begin{aligned}
    (\overline{\xi}^{*} - \overline{\xi}_i) & = \frac{\left(\phi(\xi_i, \eta_i) - \overline{\phi}(\overline{\xi}_i,\overline{\eta}_i) \right)\overline{\phi}_{\overline{\xi}}(\overline{\xi}_i, \overline{\eta}_i)}{\overline{\phi}_{\overline{\xi}}^2(\overline{\xi}_i, \overline{\eta}_i) + \overline{\phi}_{\overline{\eta}}^2(\overline{\xi}_i, \overline{\eta}_i)} \\
    & = \frac{\phi(\xi_i, \eta_i) - \overline{\phi}(\overline{\xi}_i,\overline{\eta}_i)}{|\nabla \overline{\phi}|^2 (\overline{\xi}_i, \overline{\eta}_i)} \overline{\phi}_{\overline{\xi}}(\overline{\xi}_i, \overline{\eta}_i).
\end{aligned}
\end{equation}
Similarly,
\begin{equation}
    (\overline{\eta}^{*} - \overline{\eta}_i) = \frac{\phi(\xi_i, \eta_i) - \overline{\phi}(\overline{\xi}_i,\overline{\eta}_i)}{|\nabla \overline{\phi}|^2 (\overline{\xi}_i, \overline{\eta}_i)} \overline{\phi}_{\overline{\eta}}(\overline{\xi}_i, \overline{\eta}_i).
\end{equation}
Thus, the optimal distance $\overline{d}$ is,
\begin{equation}
% \begin{aligned}
    \overline{d}^2 = (\overline{\xi}^{*} - \overline{\xi}_i)^2 + (\overline{\eta}^{*} - \overline{\eta}_i)^2
    = \frac{\left(\phi(\xi_i, \eta_i) - \overline{\phi}(\overline{\xi}_i,\overline{\eta}_i) \right)^2}{|\nabla \overline{\phi}|^2 (\overline{\xi}_i, \overline{\eta}_i)},
% \end{aligned}
\end{equation}
so that
\begin{equation}
    \overline{d} = \frac{|\phi(\xi_i, \eta_i) - \overline{\phi}(\overline{\xi}_i,\overline{\eta}_i)|}{|\nabla \overline{\phi}|(\overline{\xi}_i, \overline{\eta}_i)}.
\label{eq:optimal_dist}
\end{equation}
Now, this expression can be reduced to gradient of $\rho$ which is directly available, since by definition,
\begin{equation}
\begin{aligned}
    \overline{\phi}(\overline{\xi}, \overline{\eta}) & = |\left( \overline{\rho} \overline{\xi}, \overline{\rho} \overline{\eta}, \overline{\rho} \right) - \left( \overline{\rho}_0 \overline{\xi}_0, \overline{\rho}_0 \overline{\eta}_0, \overline{\rho}_0 \right)|^2 \\
    & = (\overline{\rho} \overline{\xi} - \overline{\rho}_0 \overline{\xi}_0)^2 + (\overline{\rho} \overline{\eta} - \overline{\rho}_0 \overline{\eta}_0)^2 + (\overline{\rho} - \overline{\rho}_0)^2.
\end{aligned}
\end{equation}
The gradient $\nabla \overline{\phi}$ can be written in terms of $\nabla \rho$,
\begin{equation}
\begin{aligned}
    \nabla \overline{\phi}(\overline{\xi}, \overline{\eta}) = & 2(\overline{\rho} \overline{\xi} - \overline{\rho}_0 \overline{\xi}_0) \nabla \overline{\rho} \, \overline{\xi} + 2(\overline{\rho} \overline{\xi} - \overline{\rho}_0 \overline{\xi}_0) \overline{\rho} \, e_1 \\
    & + 2(\overline{\rho} \, \overline{\eta} - \overline{\rho}_0 \overline{\eta}_0) \nabla \overline{\rho} \, \overline{\eta} + 2(\overline{\rho} \, \overline{\eta} - \overline{\rho}_0 \overline{\eta}_0) \overline{\rho} \, e_2 \\
    & + 2(\overline{\rho} - \overline{\rho}_0) \nabla \overline{\rho} \\
    = & 2 \Biggl[ (\overline{\rho} \overline{\xi} - \overline{\rho}_0 \overline{\xi}_0) \overline{\xi} + (\overline{\rho} \, \overline{\eta} - \overline{\rho}_0 \overline{\eta}_0) \overline{\eta} + (\overline{\rho} - \overline{\rho}_0) \Biggr] \nabla \overline{\rho} \\
    & + 2 \overline{\rho} \begin{bmatrix} \overline{\rho} \overline{\xi} - \overline{\rho}_0 \overline{\xi}_0 \\ \overline{\rho} \,\overline{\eta} - \overline{\rho}_0 \overline{\eta}_0 \end{bmatrix} \\
    = & 2(\overline{\rho} |\overline{\gamma}|^2 - \overline{\rho}_0 \overline{\gamma}_0^T \overline{\gamma}) \nabla \overline{\rho} + 2 \overline{\rho} \begin{bmatrix} 1\;\;0\;\;0 \\ 0\;\;1\;\;0 \end{bmatrix} (\overline{\rho} \, \overline{\gamma} - \overline{\rho}_0 \overline{\gamma}_0).
\end{aligned}
\end{equation}
Using this expression in Equation~\ref{eq:optimal_dist} at $(\overline{\xi}_i, \overline{\eta}_i)$ proves the proposition.
\end{proof}

\section{Details of the Datasets Used in the Paper}

Three popular datasets, namely, TUM-RGBD~\cite{TUMRGBD_Dataset}, ICL-NUIM~\cite{handa2014benchmark}, and RGBD Scenes v2~\cite{RGBD_Scenes_v2_Dataset}, are used in experiments. Specifically, from the TUM-RGBD dataset, six sequences are used, \emph{i.e.} \emph{freiburg1\_desk} (fr1/desk), \emph{freiburg1\_room} (fr1/room), \emph{freiburg1\_xyz} (fr1/xyz), \emph{freiburg2\_desk} (fr2/desk), \emph{freiburg3\_long\_office\_household} (fr3/office), and \emph{freiburg3\_structure\_texture\_near\_validation} (fr3/struct). These sequences were chosen to cover a diverse set of conditions: 
The first three sequences exhibit blurry images and illumination variations; the fourth sequence exhibits a generic textureless scene; and, the last two sequences exhibit mixtures of texture/textureless and planar/non-planar scenes. Second, all eight sequences of the ICL-NUIM dataset are used, exhibiting low contrast and low texture synthetic indoor scenes with artificial depth noise. 
Finally, all 14 sequences of RGBD Scene v2 dataset are used, exhibiting low illumination, repetitive features, homogeneous indoor scenes with a large portion of the image having no depth values. Image resolutions of all three datasets are identical and comparatively small, \emph{i.e.}, 480$\times$640.

\begin{figure}[t]
    \centering
    (a)\includegraphics[scale=0.11]{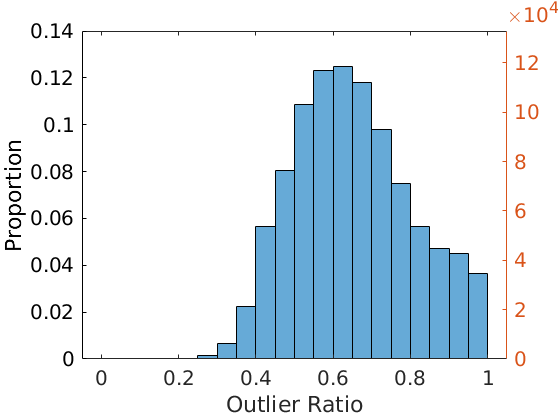}
    (b)\includegraphics[scale=0.11]{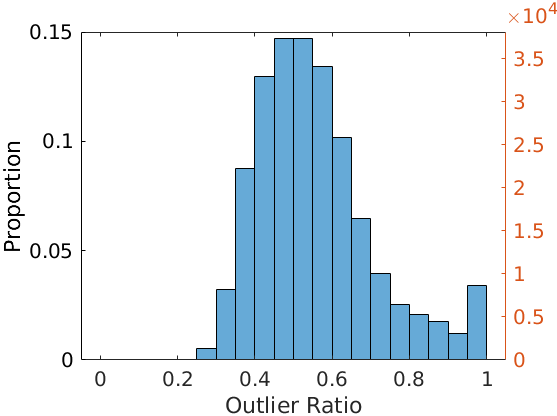}
    (c)\includegraphics[scale=0.11]{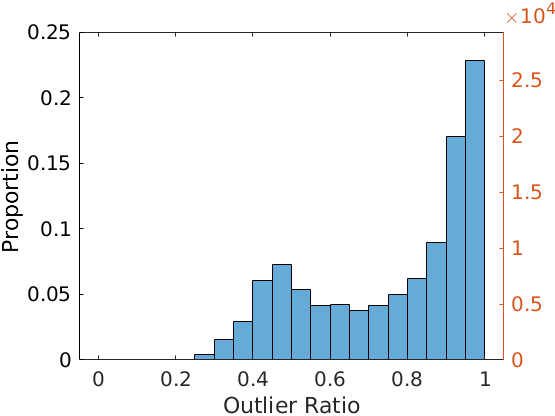}
    \caption{Distribution of outlier ratio $e$ for pairs of images from the (a) TUM-RGBD~\cite{TUMRGBD_Dataset}, (b) ICL-NUIM~\cite{handa2014benchmark}, and (c) RGBD Scene v2~\cite{RGBD_Scenes_v2_Dataset} datasets. Number of image pairs are 132,946, 38,085, and 39,325 image pairs, respectively. The bin size used in this histogram is 0.05. }
    \label{fig:dataset_outlier_ratio_distribution}
\end{figure}

Pairs of images from these datasets show varying extent of outlier correspondences, as shown in the outlier distributions in
Figure~\ref{fig:dataset_outlier_ratio_distribution}, for 132,946, 38,085, and 39,325 image pairs from the TUM-RGBD, ICL-NUIM, and RGBD Scene v2 datasets, respectively. Specifically, each image is paired with subsequent images at intervals ranging from 1 to 30 time steps. The figure shows that while each of the dataset has high outlier ratio, RGBD Scene v2 particularly exhibits very high outlier ratio, providing situations where the proposed GDC filter can be applied effectively. 

\section{Reducing Outlier Ratio by the GDC Filter}
\begin{figure}[!htbp]
    \centering
    \includegraphics[width=.21\textwidth]{Figures/e_before_and_after_many.png}\quad
    \includegraphics[width=.21\textwidth]{Figures/mu_vs_e_many.png}\\
    \includegraphics[width=.21\textwidth]{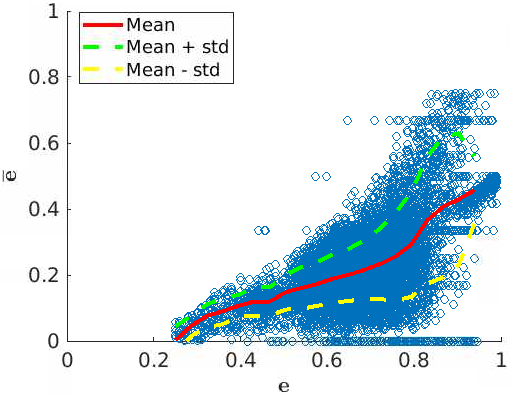}\quad
    \includegraphics[width=.21\textwidth]{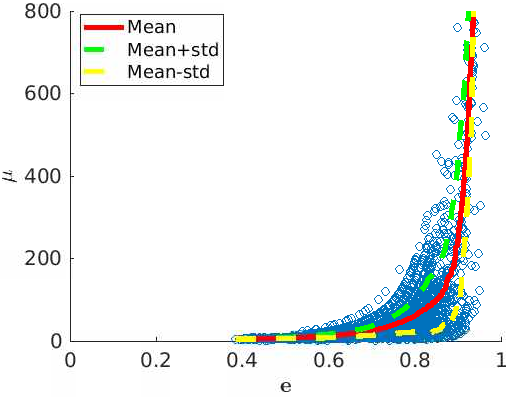}\\
    \includegraphics[width=.21\textwidth]{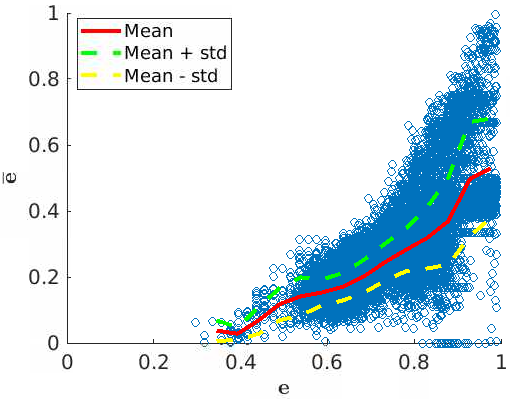}\quad
    \includegraphics[width=.21\textwidth]{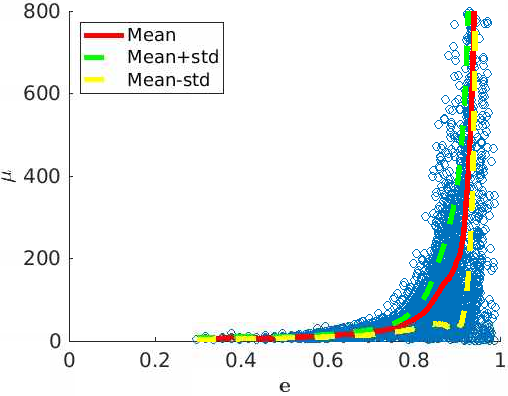}\\
    (a) \hspace{10em} (b)
    \caption{\textbf{Top to bottom:} TUM-RGBD, ICL-NUIM, and RGBD Scene v2 datasets. \textbf{Left to right:} (a) The scatter plot of $e$ and $\overline{e}$, namely, the outlier ratios before and after the GDC filter is applied. (b) The ratio of the number of required iterations before and after applying the GDC filter to the three datasets, for the minimum success probability of 0.99 using 2000 RANSAC iterations.}
    \label{fig:scatter_plot_outlier_ratios}
\end{figure}

\begin{figure*}[t]
    \centering
    \includegraphics[scale=0.19]{Figures/Prob_One_True.png}
    \includegraphics[scale=0.19]{Figures/Prob_Two_Inliers.png}
    \includegraphics[scale=0.19]{Figures/Prob_Three_Inliers.png} \\
    \includegraphics[scale=0.285]{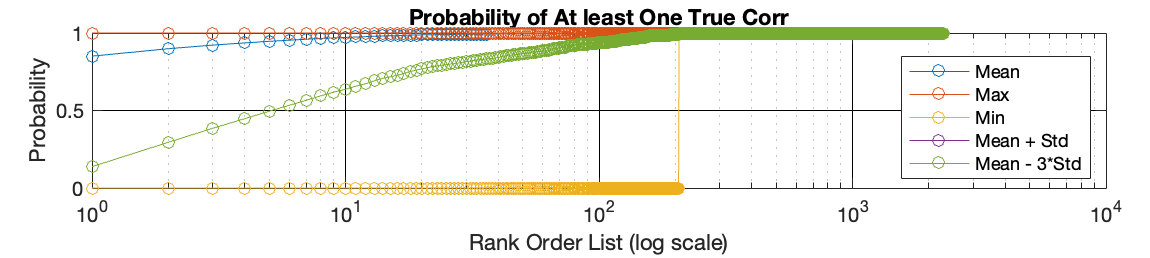}
    \includegraphics[scale=0.285]{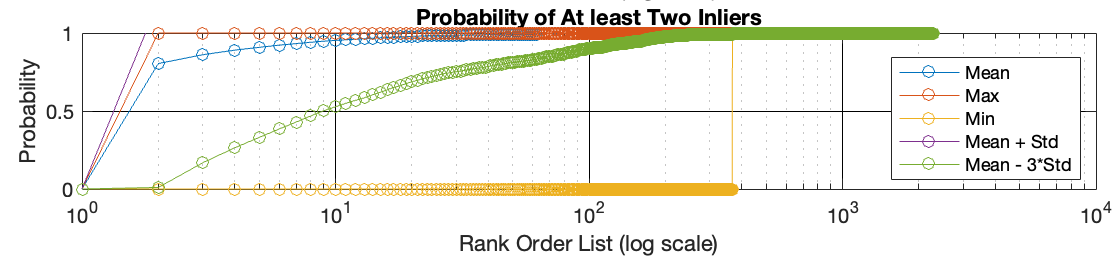}
    \includegraphics[scale=0.285]{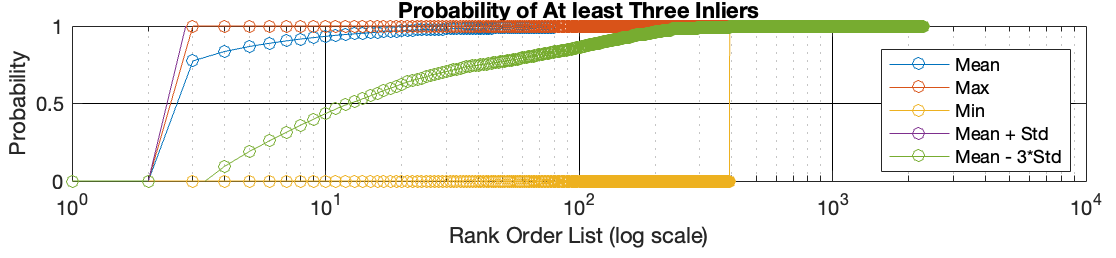} \\
    \includegraphics[scale=0.17]{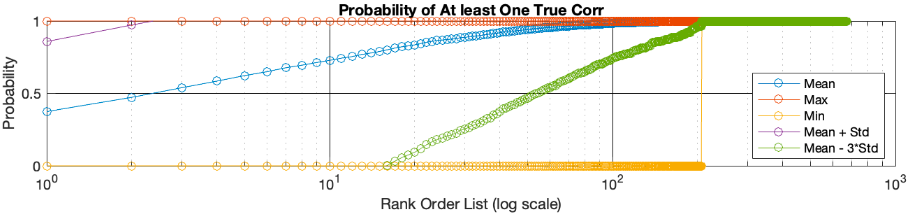}
    \includegraphics[scale=0.17]{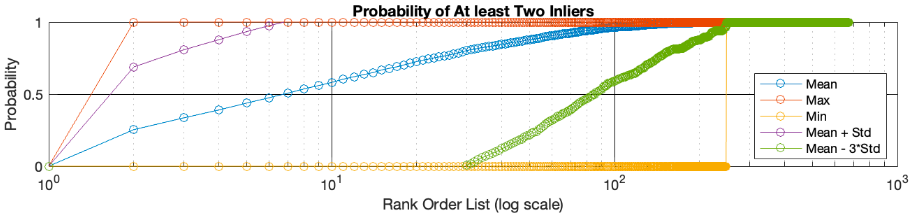}
    \includegraphics[scale=0.17]{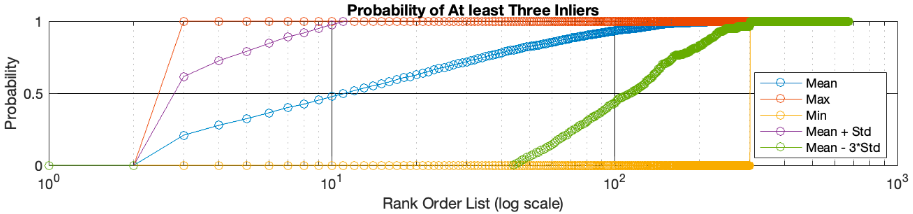} \\
    \caption{\textbf{Top to bottom:} The TUM-RGBD, ICL-NUIM, and RGBD Scene v2 datasets. \textbf{Left to right:} The probability of finding at least one, at least two, and at least three inliers across the rank-ordered list of matches.}
    \label{fig:prob_all_three_datasets}
\end{figure*}

The main paper showed that the GDC filter reduced the outlier ratio on the TUM-RGBD dataset significantly. Here, results for the ICL-NUIM and RGBD Scene v2 datasets are also shown, duplicating the TUM-RGBD dataset result for ease of comparison, Figure~\ref{fig:scatter_plot_outlier_ratios}. This shows that the GDC filter is effective across a diverse set of datasets. 

\section{Likelihood of Finding $s$-Tuplets in the Rank-Ordered List}

The likelihood of finding $s$-tuplets in the top $m$ set of correspondences for $1 \leq m \leq M$ for different values of $s$ which were shown for the TUM-RGBD dataset in the main paper, are now shown for the ICL-NUIM and the RGBD Scene v2 datasets in Figure~\ref{fig:prob_all_three_datasets}. 
Theses figures confirm that the experimental setup of $M_1 \geq 100$, $M_2 \geq 150$, and $M_3 \geq 250$ is consistent for both ICL-NUIM and RGBD Scene v2 datasets.

\section{Ground-Truth Construction}

\begin{figure}[t]
    \centering
    \includegraphics[width=.15\textwidth]{Figures/reprojection_err_gt_nongt.png}
    \includegraphics[width=.15\textwidth]{Figures/depth_err_gt_non_gt.png}
    \includegraphics[width=.15\textwidth]{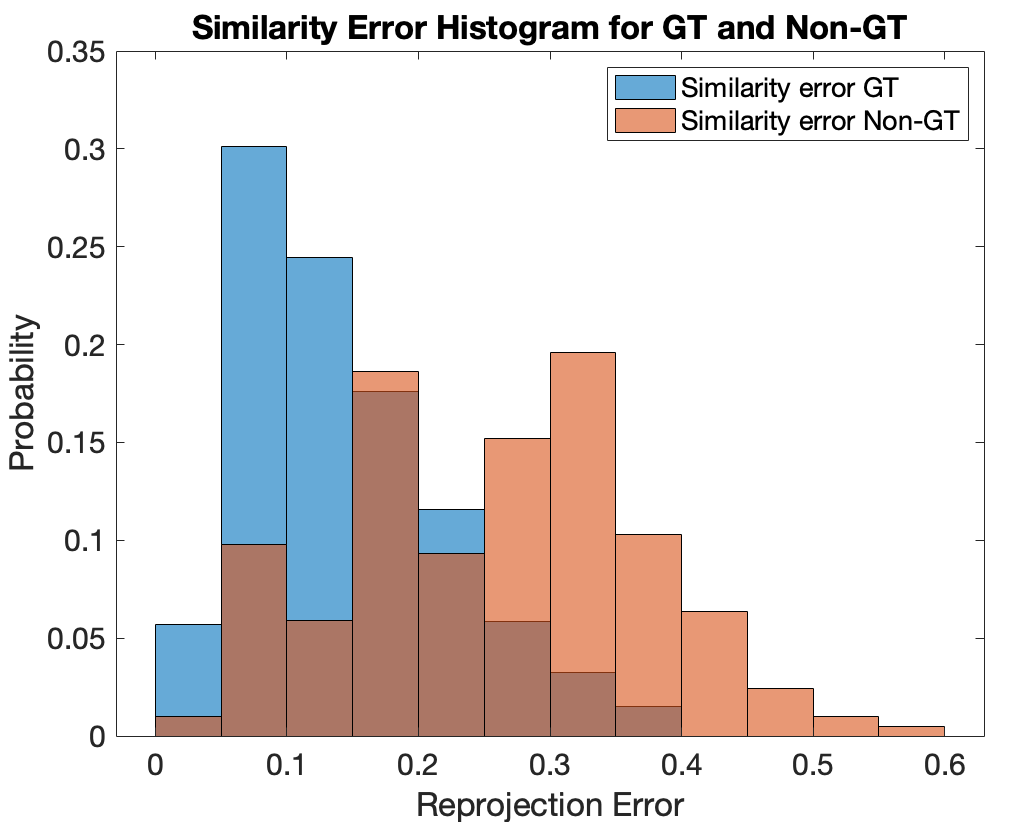}\\
    \includegraphics[width=.15\textwidth]{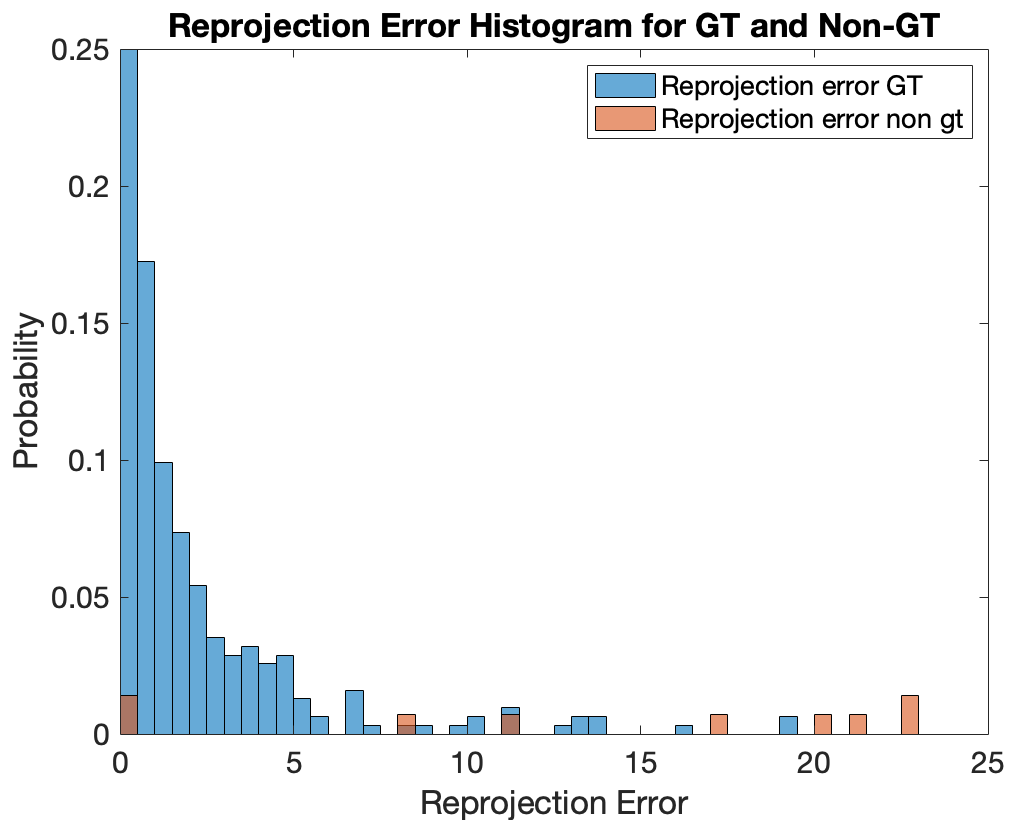}
    \includegraphics[width=.15\textwidth]{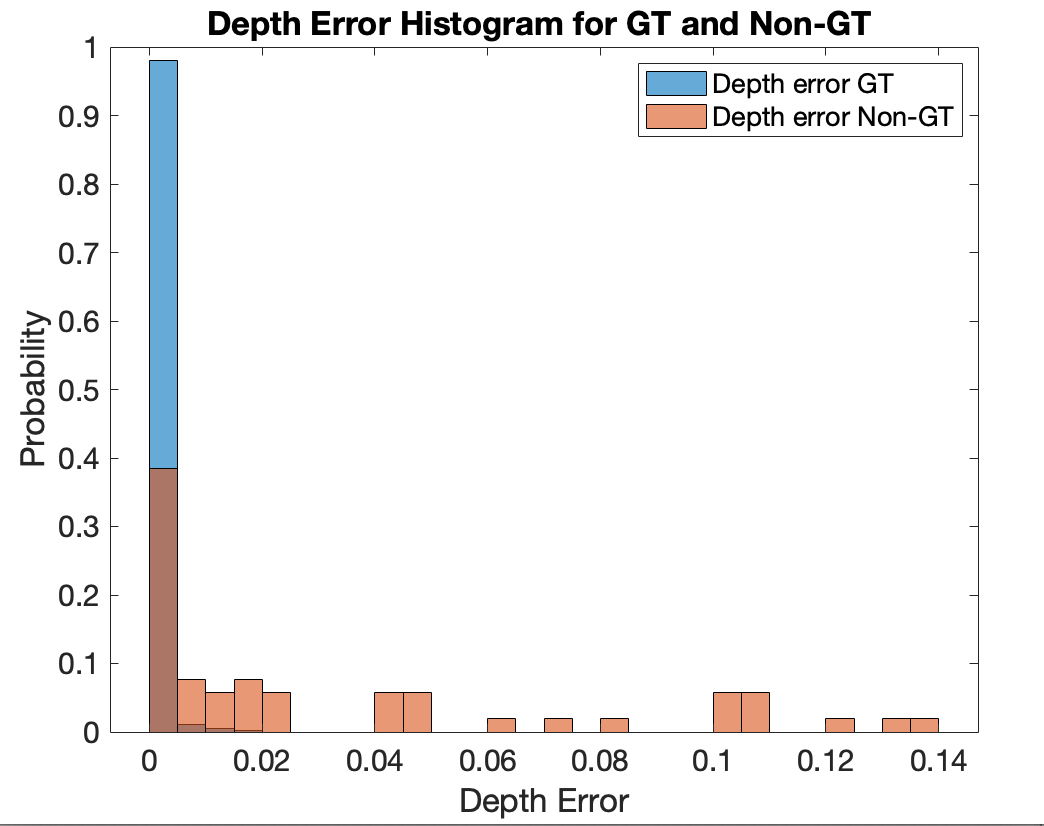}
    \includegraphics[width=.15\textwidth]{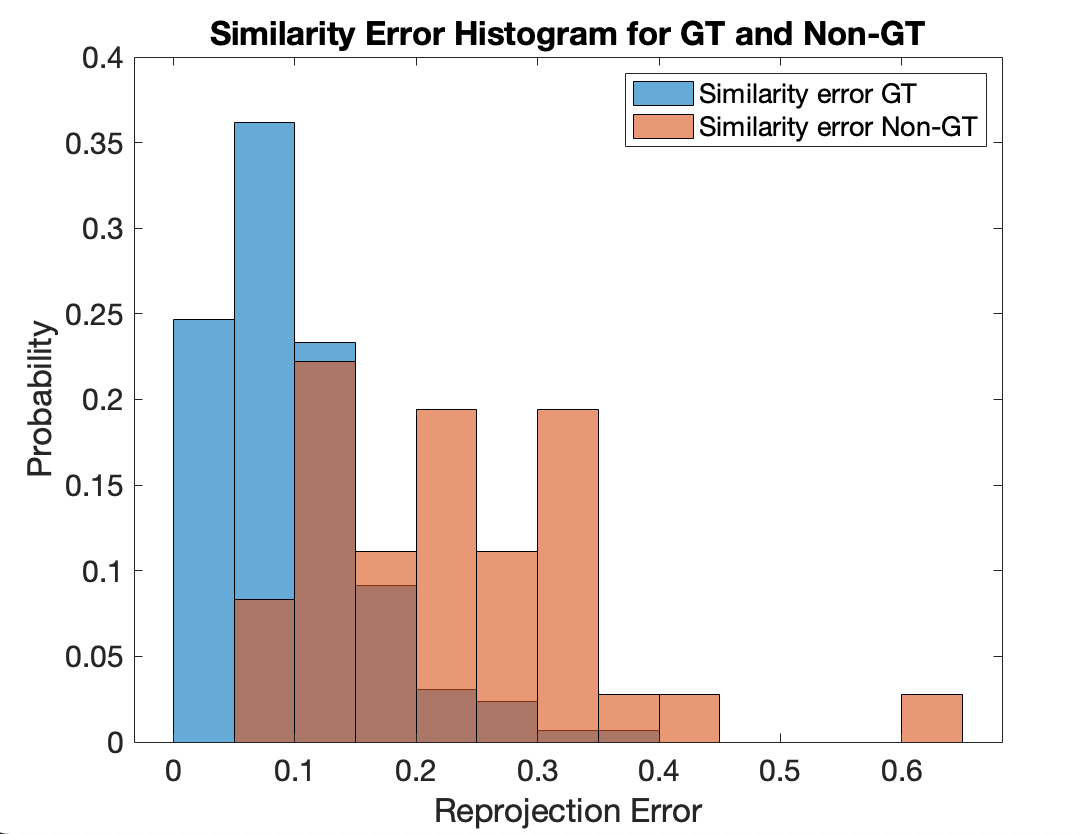}\\
    \includegraphics[width=.15\textwidth]{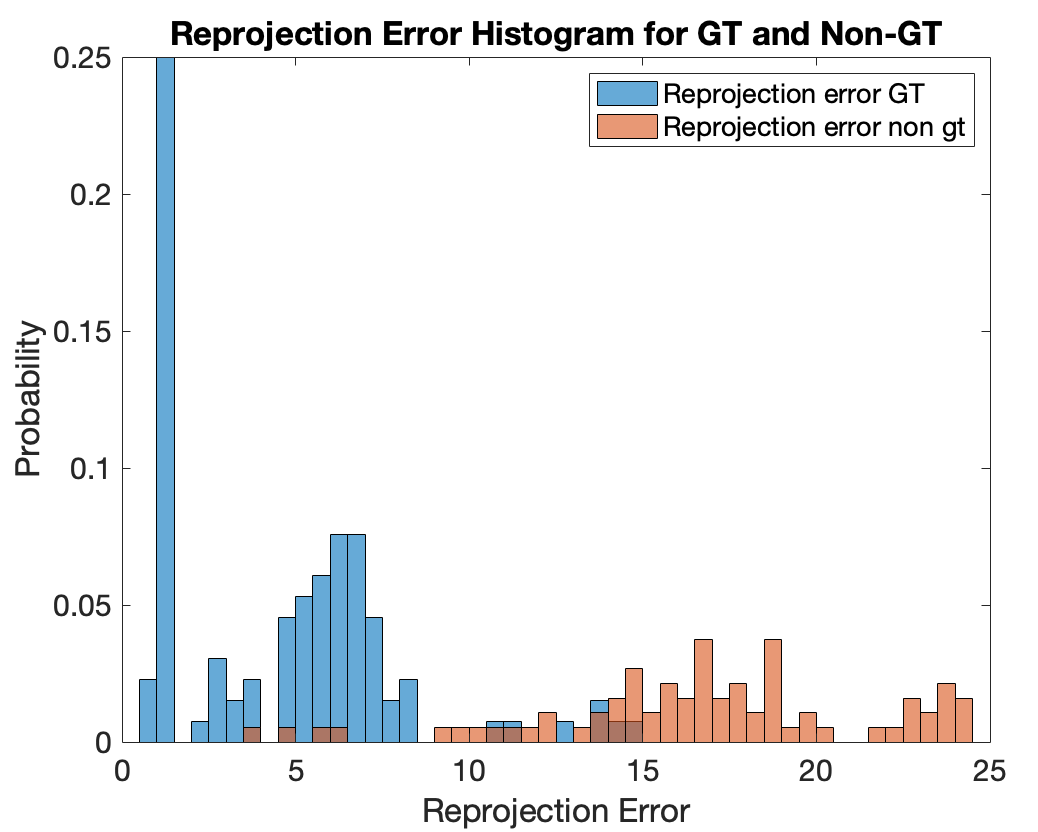}
    \includegraphics[width=.15\textwidth]{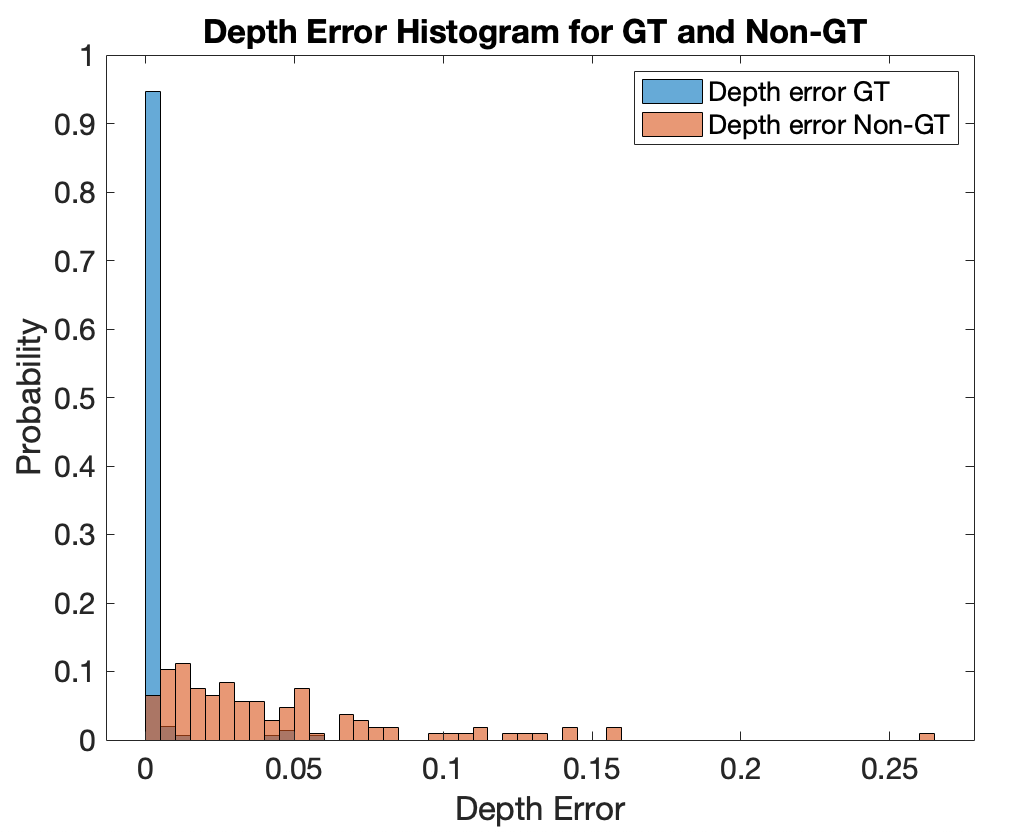}
    \includegraphics[width=.15\textwidth]{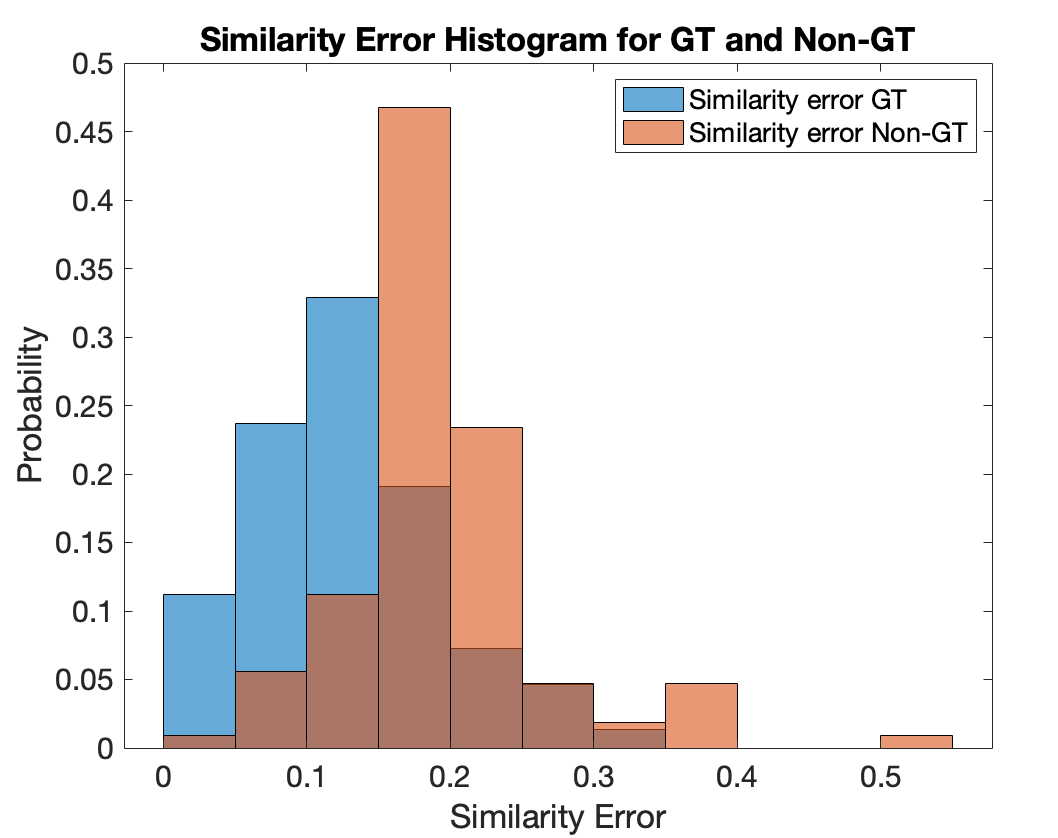}\\
    (a) \hspace{6em} (b) \hspace{6em} (c)
    \caption{\textbf{Top to bottom:} TUM-RGBD, ICL-NUIM, and RGBD Scene v2 datasets. \textbf{Left to right:} The distribution of reprojection error (a), depth error (b), and similarity error (c) for valid (blue) and invalid (red) correspondences shows that thresholds of $\tau_{\gamma}$ = 8 (pixels), $\tau_{\rho}$ = 0.01 (m), and $\tau_s$ = 0.4 are nearly optimal to differentiate between the two groups. Note that the distribution of non-GT correspondences in (a) continues well into distances of 500 not shown here.}
    \label{fig:supp_hist_gt_non_gt}
\end{figure}

The distributions of reprojection errors, depth errors, and the similarity error for GT and non-GT correspondences which were shown in the main paper are now shown for the ICL-NUIM and the RGBD Scene v2 datasets, Figure~\ref{fig:supp_hist_gt_non_gt}. 
Observe that the set of thresholds optimally selected for the TUM-RGBD dataset are also nearly optimal for the ICL-NUIM and the RGBD Scene v2 datasets. The thresholds are $\tau_{\gamma}$ = 8 (pixels), $\tau_{\rho}$ = 0.01 (m), and $\tau_s$ = 0.4. This is supported by Table~\ref{tab:supp_table_comparing_algo_gt}(a), (b), and (c), where three images from the ICL-NUIM dataset are selected to evaluate the algorithm's determination of GT against the manually determined GT. Both the FP and FN are significantly reduced using the algorithmic, compared to the similarity-based correspondence. Table~\ref{tab:supp_table_comparing_algo_gt}(d) and (e) show the two selected images from the RGBD-Scene v2, which also demonstrate the effectiveness of the proposed algorithmic GT construction. 

\begin{figure}[!htbp]
    \centering
    \begin{tabular}{ccc}
        \includegraphics[width=.15\textwidth]{Figures/gt_a_tp.png} &
        \includegraphics[width=.15\textwidth]{Figures/gt_a_fn.png} &
        \includegraphics[width=.15\textwidth]{Figures/gt_a_fp.png} \\
        \includegraphics[width=.15\textwidth]{Figures/gt_b_tp.png} &
        \includegraphics[width=.15\textwidth]{Figures/gt_b_fn.png} &
        \includegraphics[width=.15\textwidth]{Figures/gt_b_fp.png} \\
        \includegraphics[width=.15\textwidth]{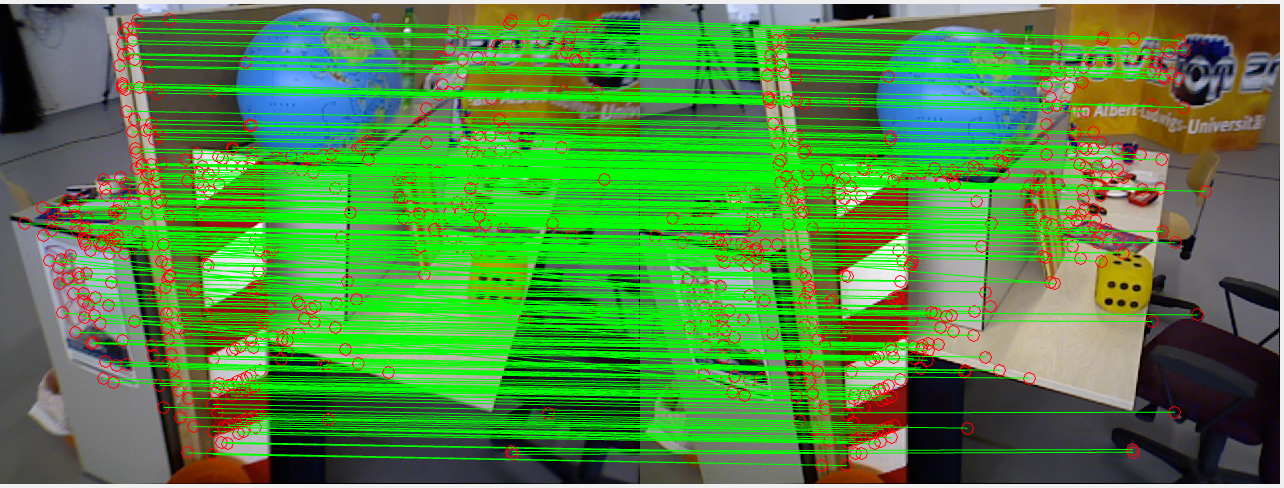} &
        \includegraphics[width=.15\textwidth]{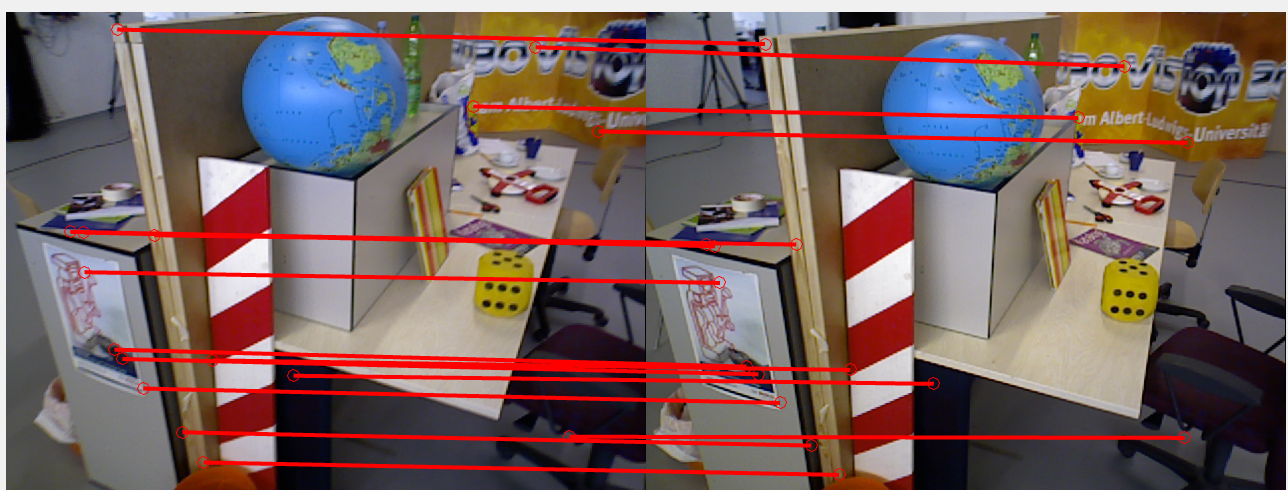} &
        \includegraphics[width=.15\textwidth]{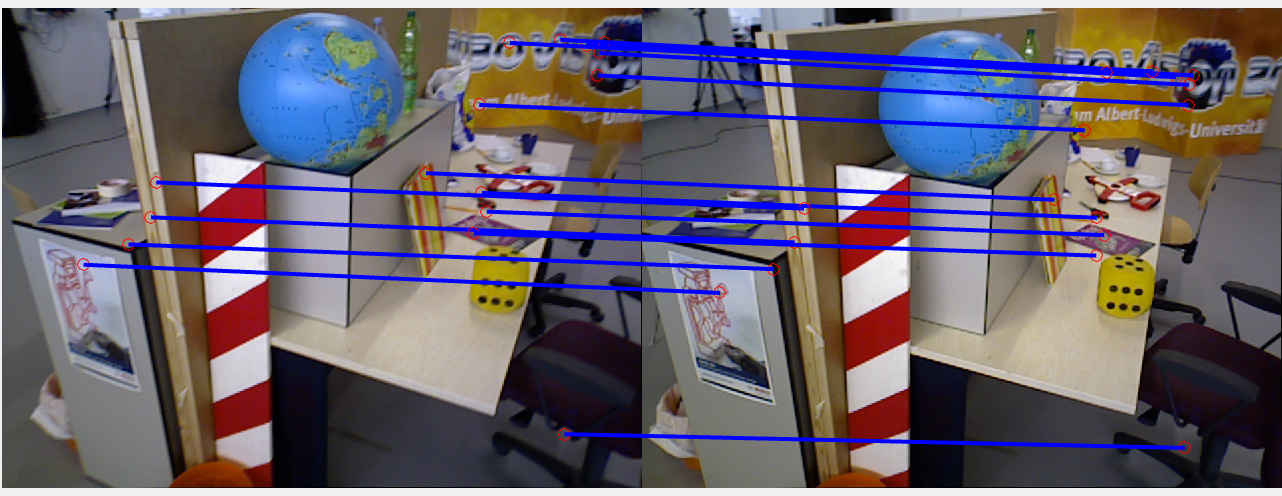} \\
        \includegraphics[width=.15\textwidth]{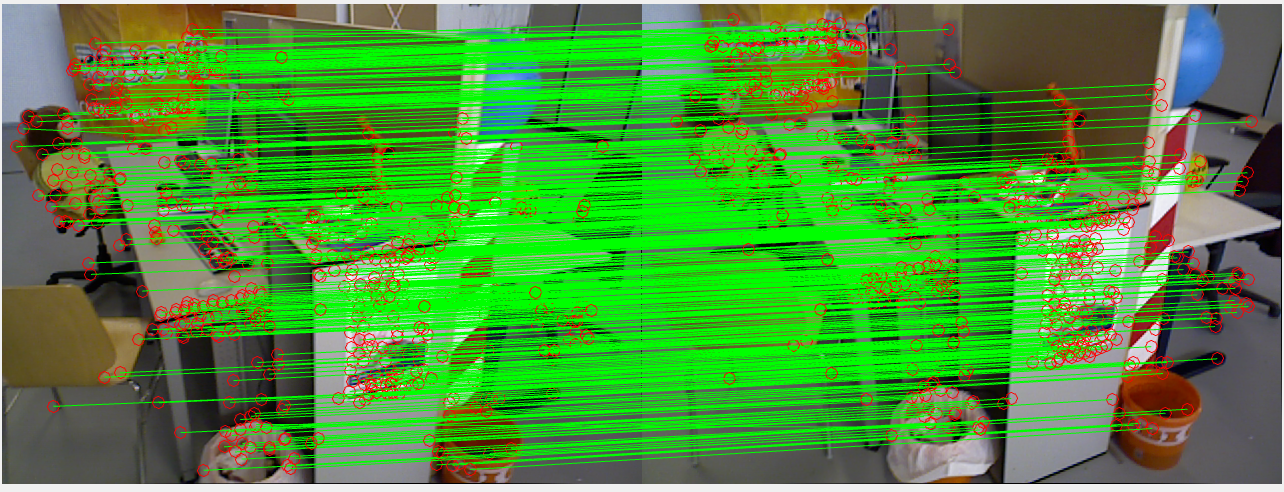} &
        \includegraphics[width=.15\textwidth]{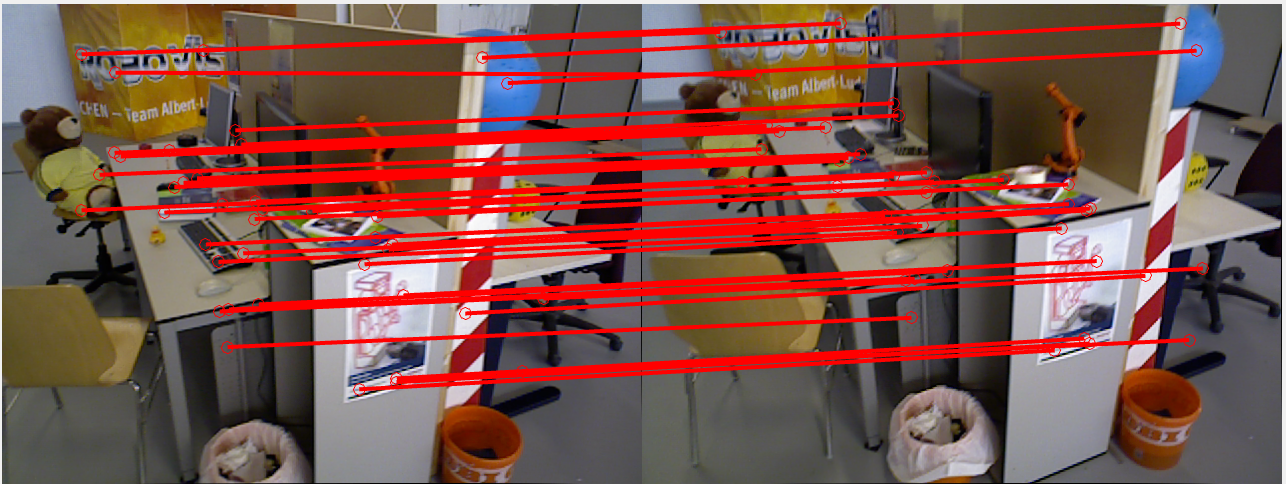} &
        \includegraphics[width=.15\textwidth]{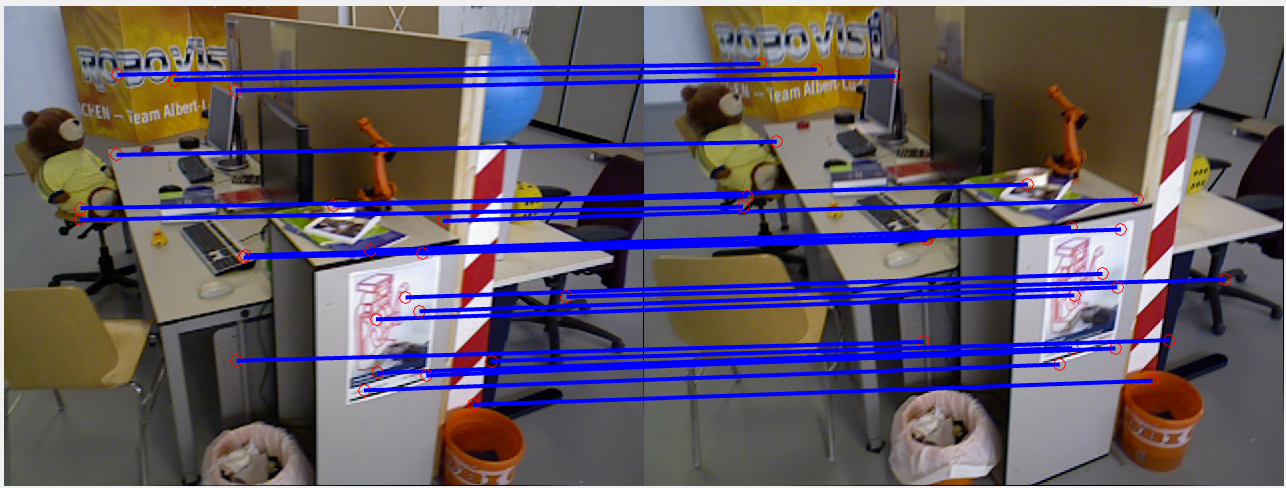} \\
        \includegraphics[width=.15\textwidth]{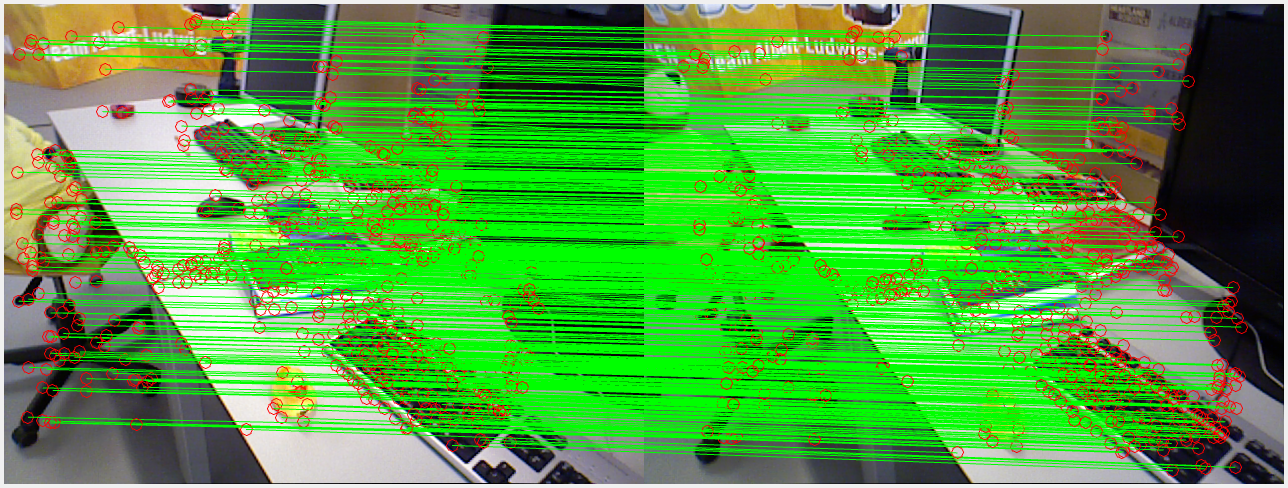} &
        \includegraphics[width=.15\textwidth]{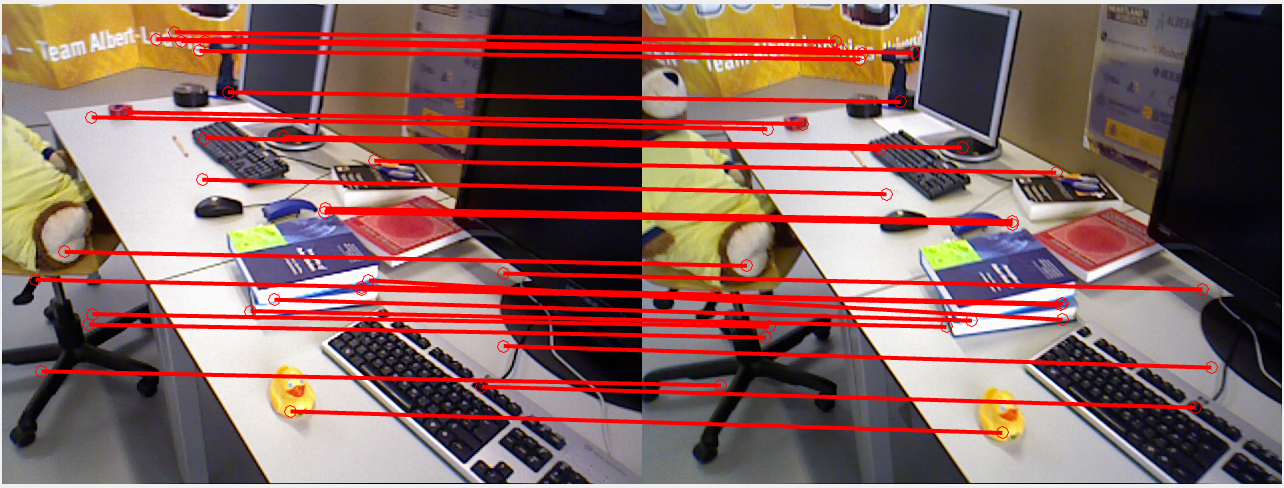} &
        \includegraphics[width=.15\textwidth]{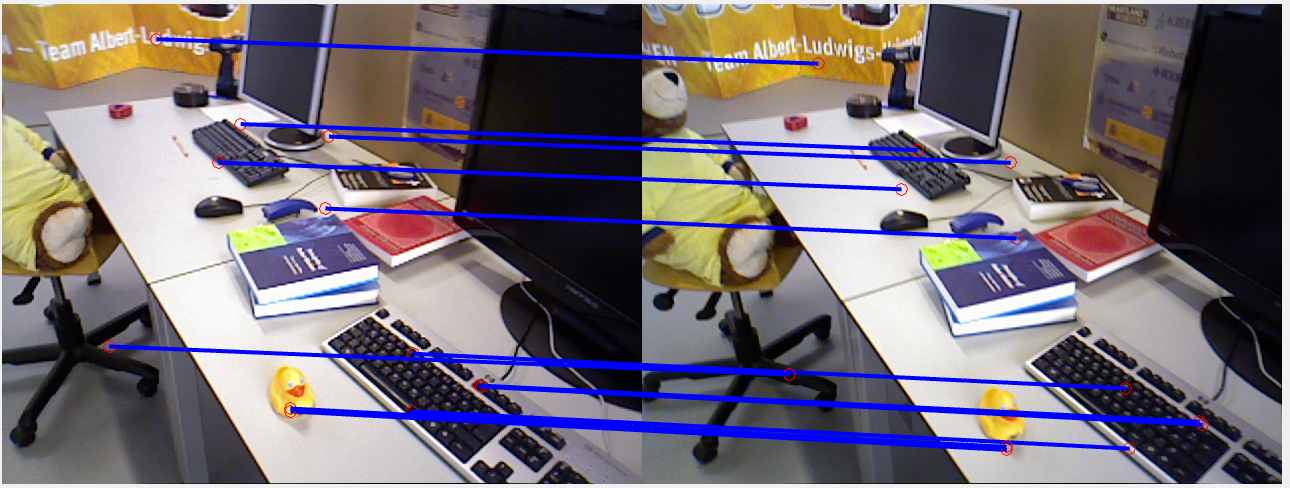} \\
        \hline 
        \vspace{-0.8em} \\
        \includegraphics[width=.15\textwidth]{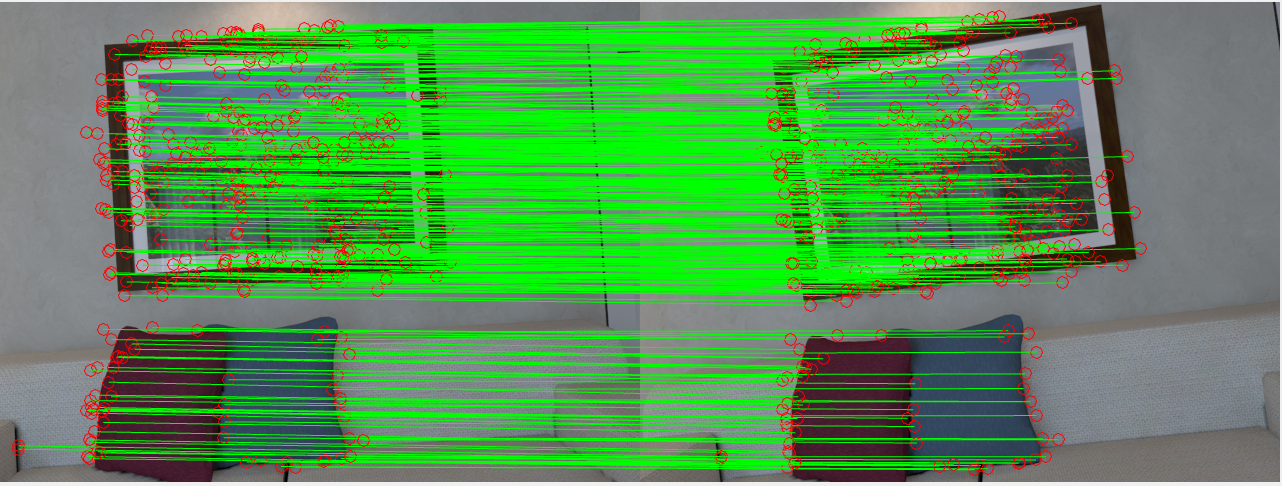} &
        \includegraphics[width=.15\textwidth]{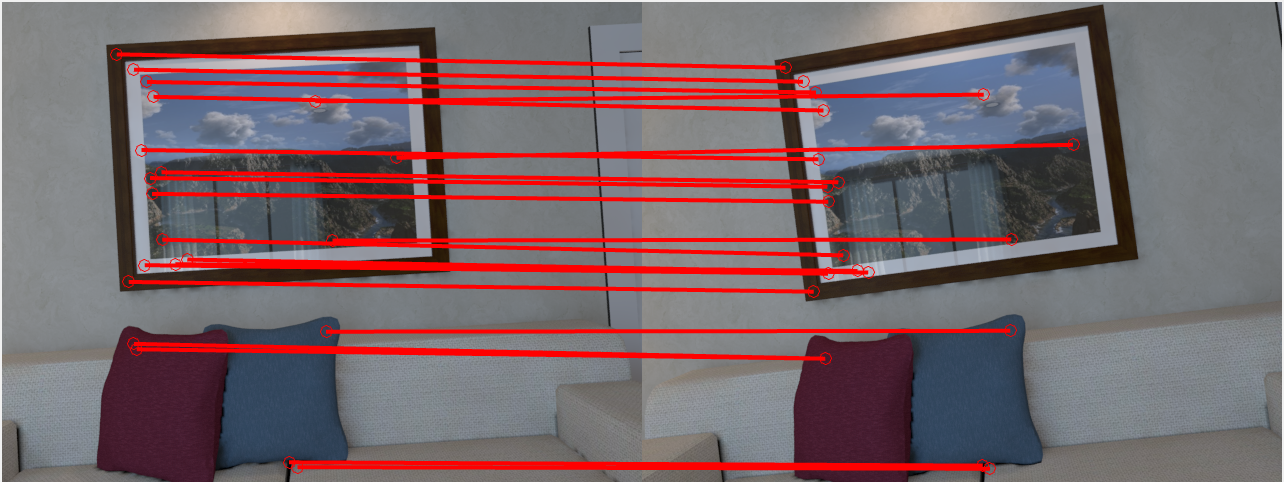} &
        \includegraphics[width=.15\textwidth]{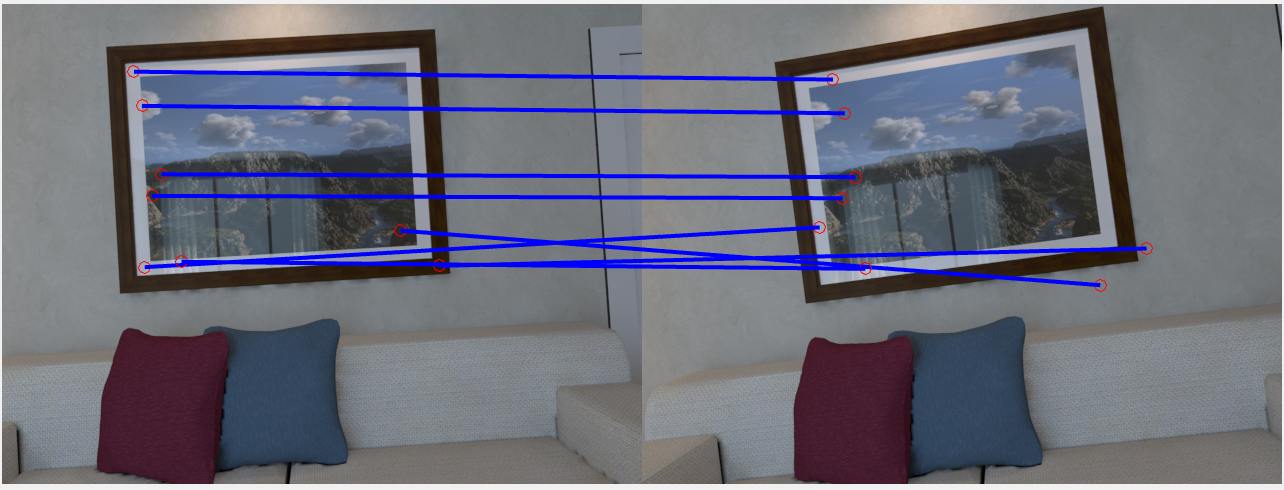} \\
        \includegraphics[width=.15\textwidth]{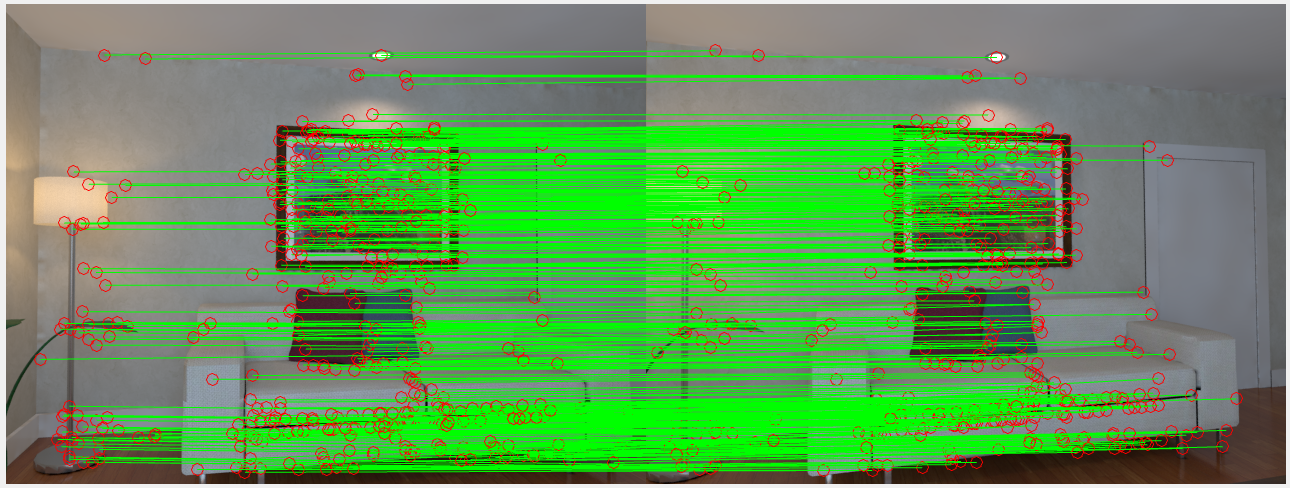} &
        \includegraphics[width=.15\textwidth]{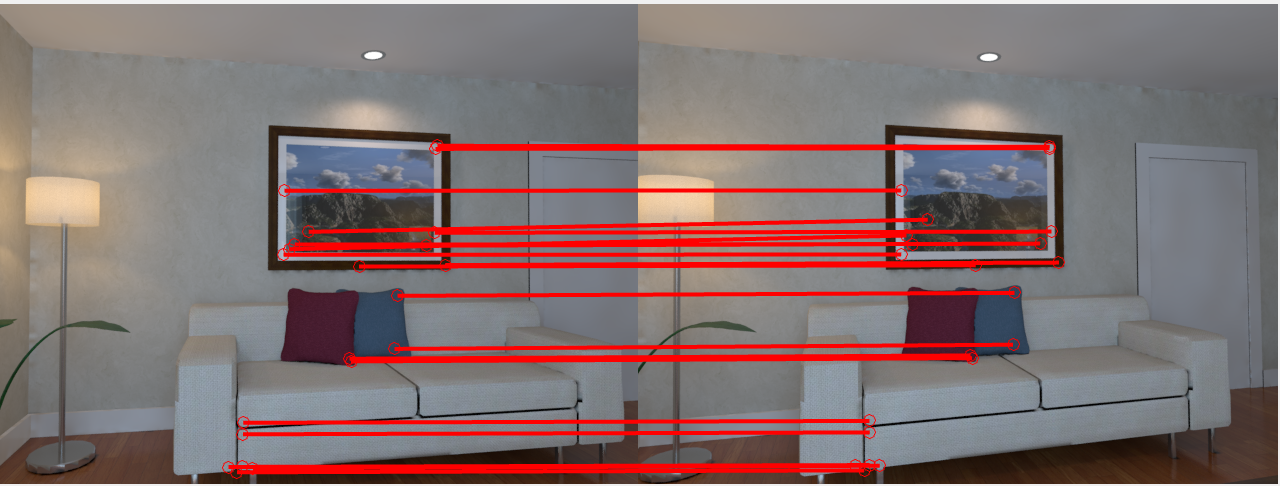} &
        \includegraphics[width=.15\textwidth]{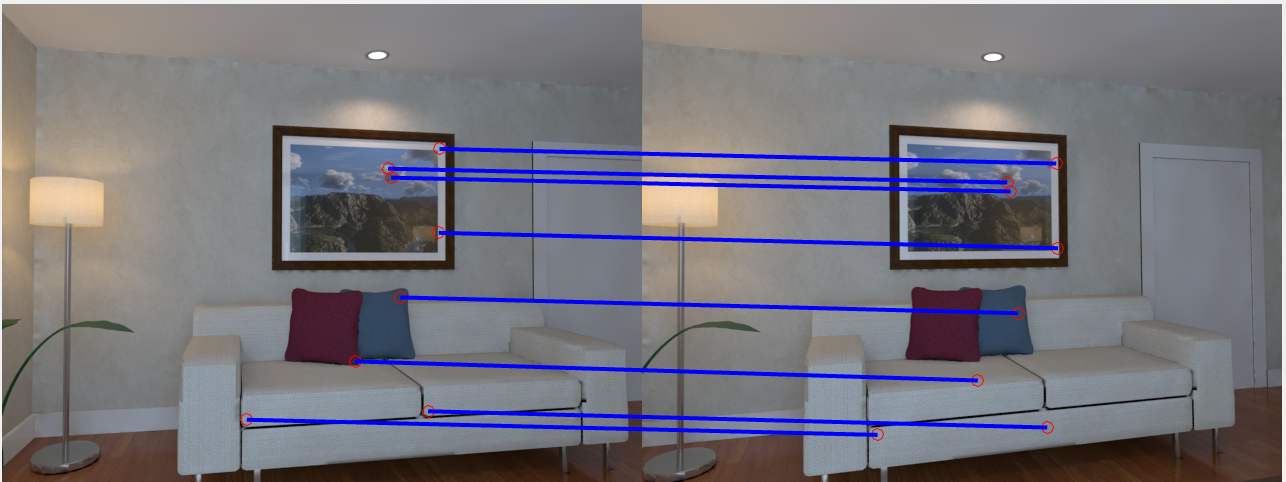} \\
        \includegraphics[width=.15\textwidth]{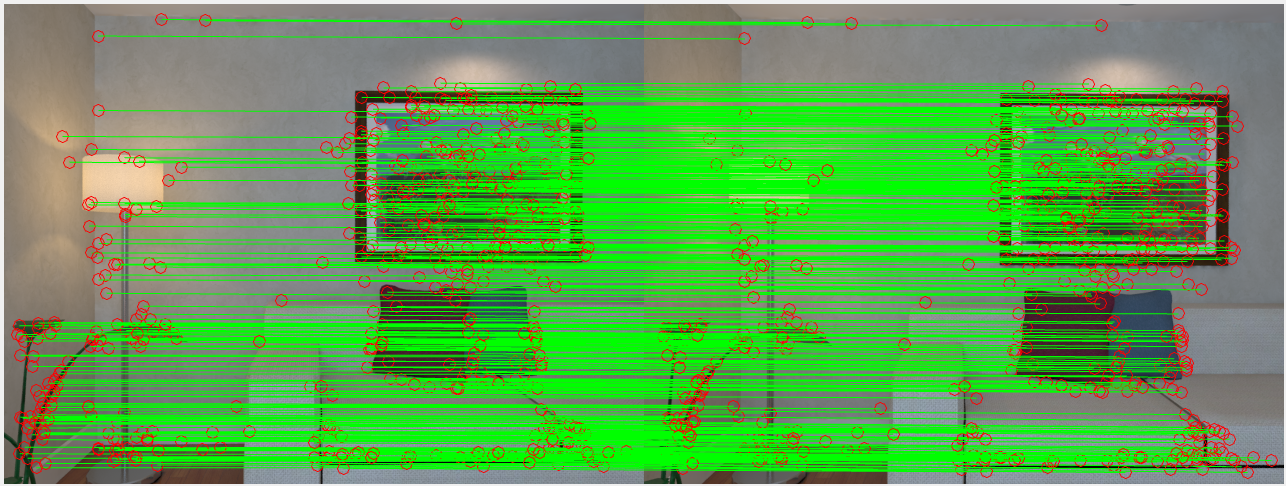} &
        \includegraphics[width=.15\textwidth]{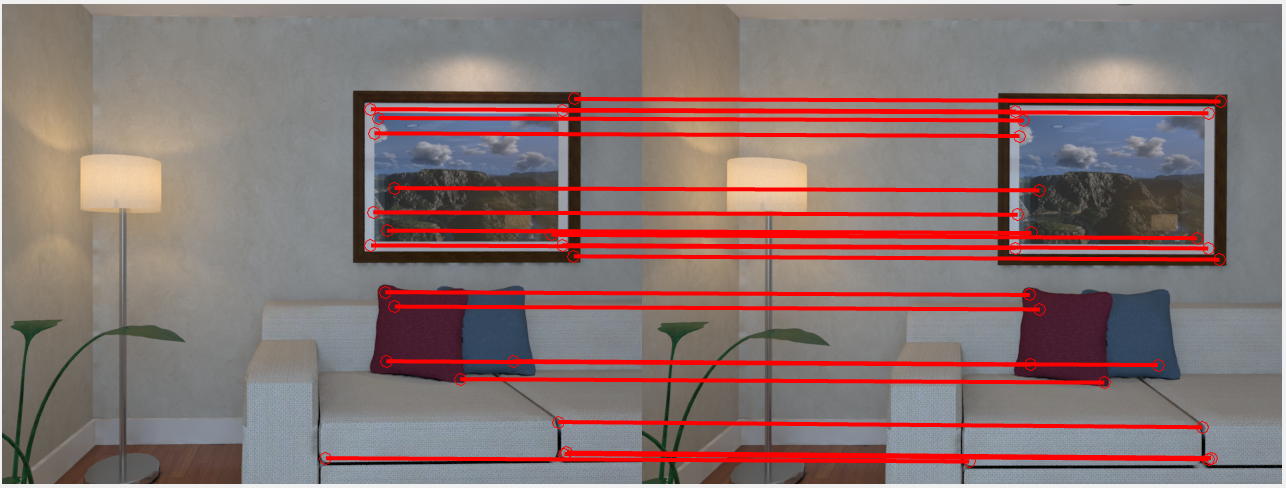} &
        \includegraphics[width=.15\textwidth]{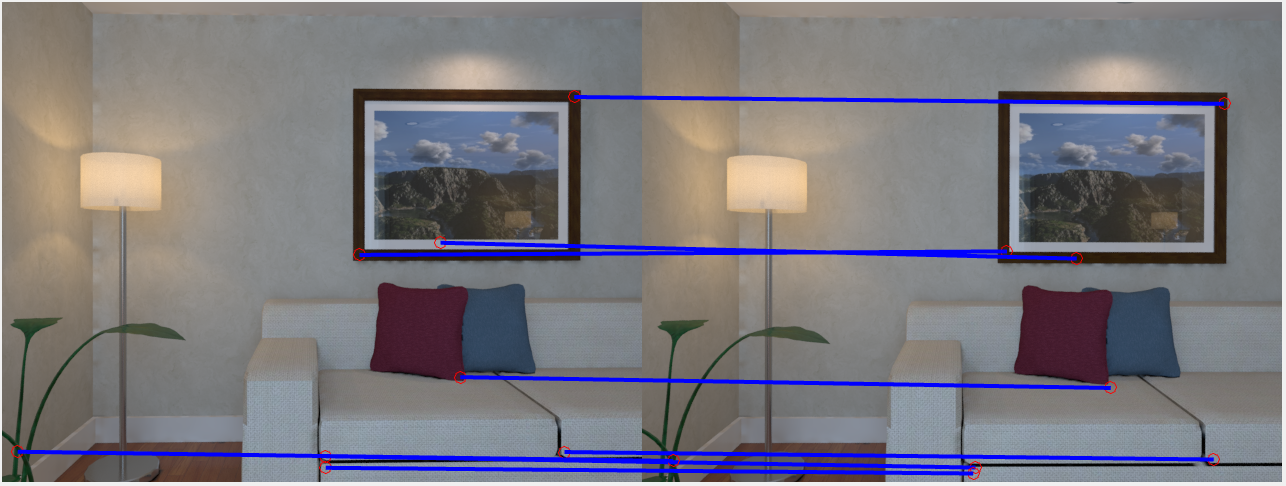} \\
        \hline 
        \vspace{-0.8em} \\
        \includegraphics[width=.15\textwidth]{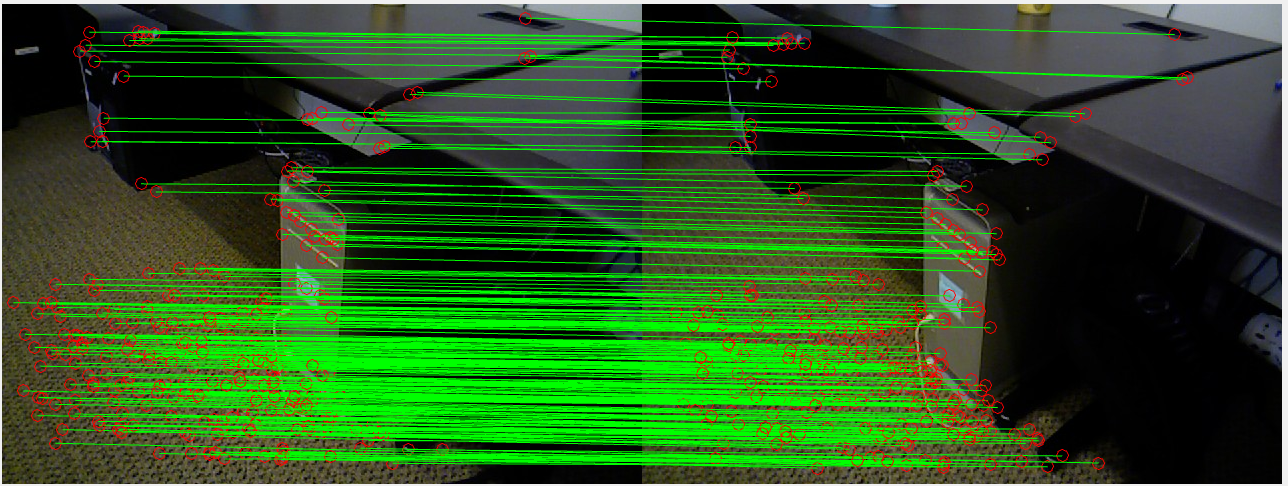} &
        \includegraphics[width=.15\textwidth]{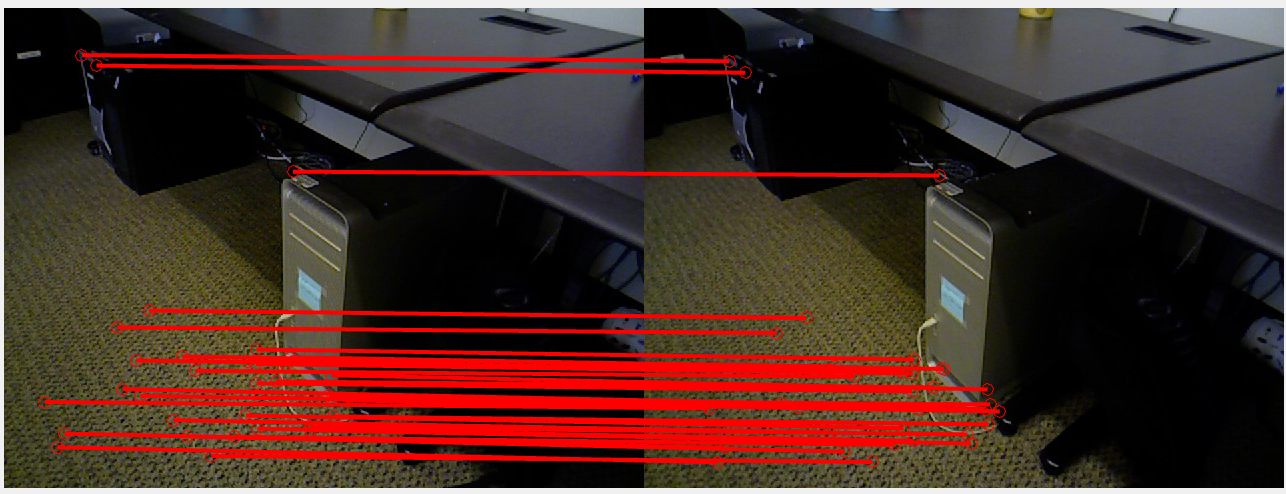} &
        \includegraphics[width=.15\textwidth]{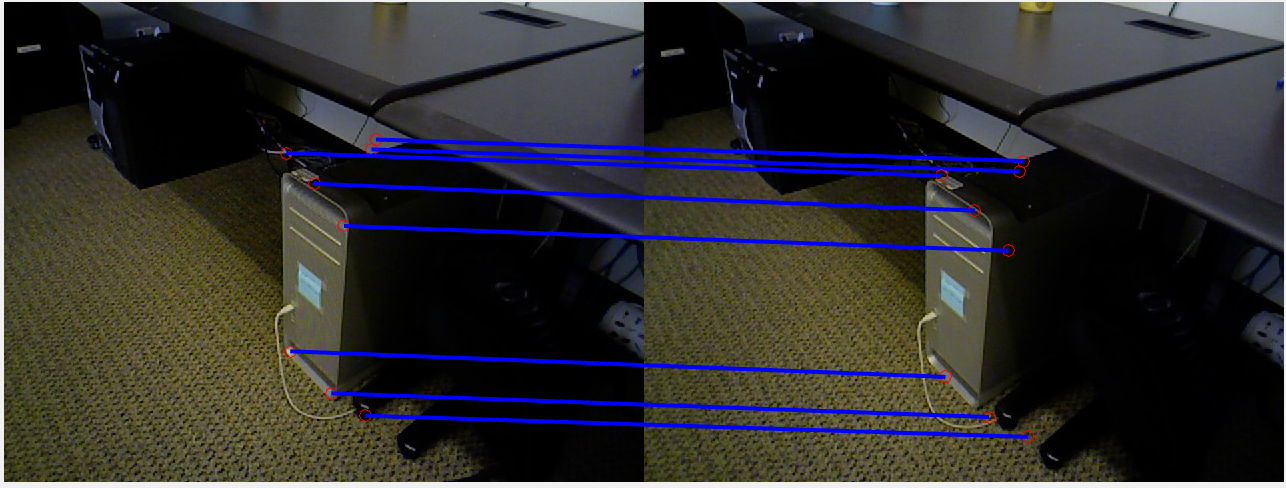} \\
        \includegraphics[width=.15\textwidth]{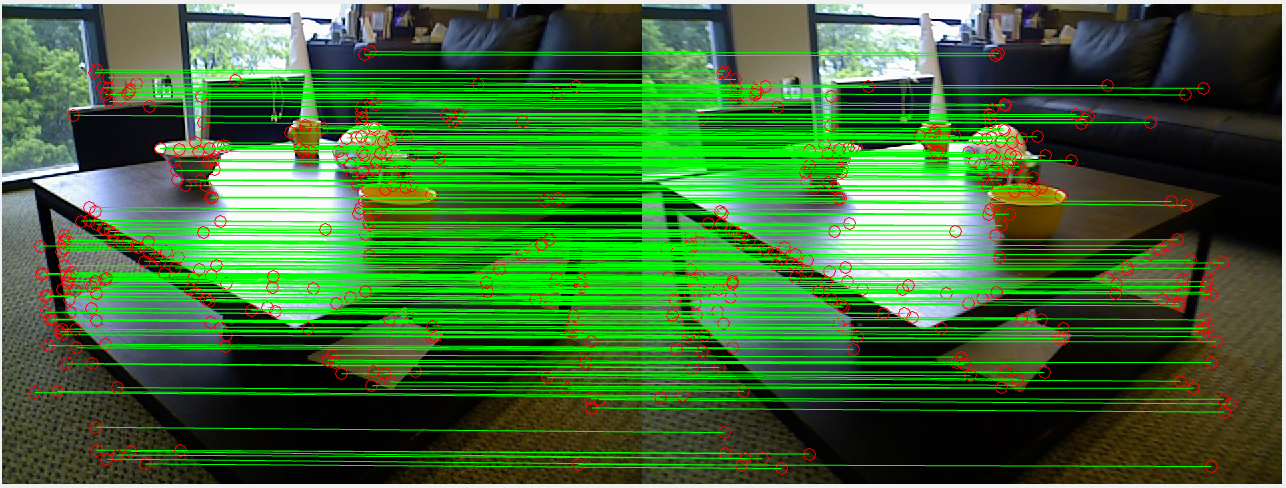} &
        \includegraphics[width=.15\textwidth]{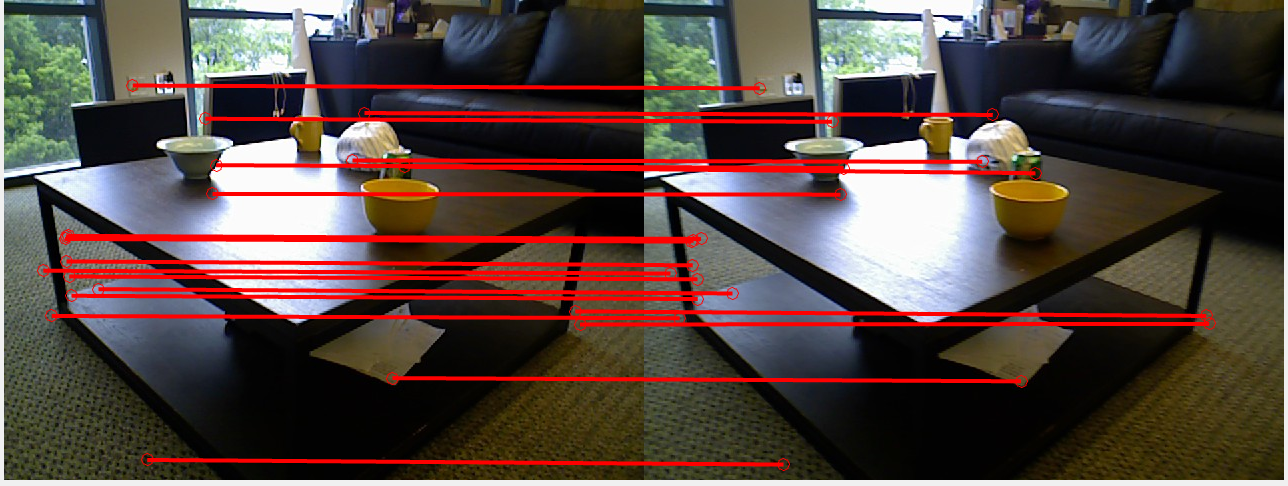} &
        \includegraphics[width=.15\textwidth]{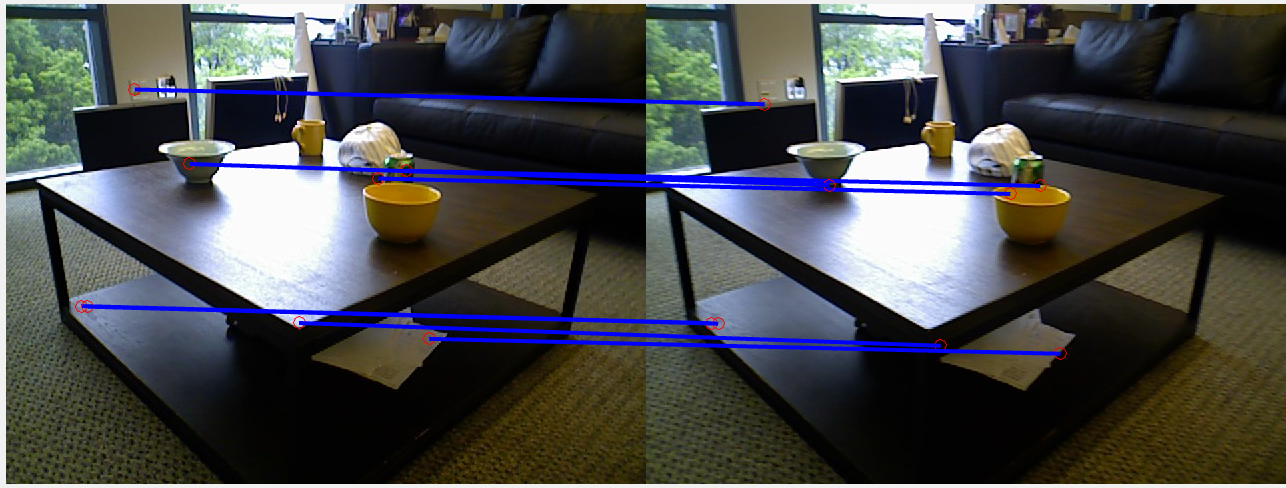}        
    \end{tabular}
    \caption{\textbf{Top 5 rows:} TUM-RGBD. \textbf{Middle 3 rows:} ICL-NUIM. \textbf{Bottom 2 rows:} RGBD Scene v2. Ground-truth correspondences on the selected images of the three datasets which are manually labeled compared to algorithmic GT showing TP (green), FN (red), and FP (blue).}
    \label{fig:gt_tp_tum}
\end{figure}

The quality of the algorithmic GT construction on the two selected images from the TUM-RGBD dataset visually illustrated in the main paper, are now shown for another three selected images from the TUM-RGBD, three selected images from the ICL-NUIM, and two selected images from the RGBD Scene v2 datasets, 
Figure~\ref{fig:gt_tp_tum}, with TP, FN, and FP correspondences shown in green, red, and blue, respectively.

\setlength\tabcolsep{2pt} 
\begin{table}[!htbp]
{\footnotesize
\centering
\begin{tabular}{|c|c|c|}
    \hline
    & \textbf{T} & \textbf{F} \\
    \hline
    \textbf{T} & 626& 284\\
    \hline
    \textbf{F} & 102 & 602\\
    \hline
    \hline
    & \textbf{T} & \textbf{F} \\
    \hline
    \textbf{T} &706  & 8\\
    \hline
    \textbf{F} & 22 & 878\\
    \hline
    \multicolumn{3}{c}{(a)}
\end{tabular}
\begin{tabular}{|c|c|c|}
    \hline
    & \textbf{T} & \textbf{F} \\
    \hline
    \textbf{T} & 776 & 302\\
    \hline
    \textbf{F} & 126 &411\\
    \hline
    \hline
    & \textbf{T} & \textbf{F} \\
    \hline
    \textbf{T} & 870 & 10\\
    \hline
    \textbf{F} & 32 & 703\\
    \hline
    \multicolumn{3}{c}{(b)}
\end{tabular}
% \hspace{0.4em}
\begin{tabular}{|c|c|c|}
    \hline
    & \textbf{T} & \textbf{F} \\
    \hline
    \textbf{T} & 960 &182 \\
    \hline
    \textbf{F} & 84 &222\\
    \hline
    \hline
    & \textbf{T} & \textbf{F} \\
    \hline
    \textbf{T} & 1030 & 16\\
    \hline
    \textbf{F} & 14 &388 \\
    \hline
    \multicolumn{3}{c}{(c)}
\end{tabular}
% \hspace{0.1em}
\begin{tabular}{|c|c|c|}
    \hline
    & \textbf{T} & \textbf{F} \\
    \hline
    \textbf{T} &536 &374\\
    \hline
    \textbf{F} & 174&716\\
    \hline
    \hline
    & \textbf{T} & \textbf{F} \\
    \hline
    \textbf{T} & 674 & 8\\
    \hline
    \textbf{F} & 36 &1082 \\
    \hline
    \multicolumn{3}{c}{(d)}
\end{tabular}
\begin{tabular}{|c|c|c|}
    \hline
    & \textbf{T} & \textbf{F} \\
    \hline
    \textbf{T} & 566 &174 \\
    \hline
    \textbf{F} & 76&351\\
    \hline
    \hline
    & \textbf{T} & \textbf{F} \\
    \hline
    \textbf{T} & 622 & 6\\
    \hline
    \textbf{F} & 20 &519\\
    \hline
    \multicolumn{3}{c}{(e)}
\end{tabular}
\caption{Two methods of establishing GT correspondences are evaluated against manual ground-truth for five image pairs randomly selected from the ICL-NUIM (a,b,c) and RGBD Scenes (d,e) datasets. The top row shows confusion matrices for similarity-based correspondences where the number of features in image one, the number of features in image two, and the number of correspondences obtained by thresholding similarity at $\tau_s$ = 0.8 are
(a) (748,866,728), (b) (790,825,902) (c) (726,722,1044)) (d) (978,822,710) (e) (583,584,642). Observe the large number of false positives (FP) and false negatives (FN) which prevent this approach from being used as a suitable algorithmic GT for evaluating correspondences. The bottom row evaluates the triple-cue algorithmic GT proposed here depicting a very small number of FP and FN. } 
\label{tab:supp_table_comparing_algo_gt}
}
\end{table}

\setlength\tabcolsep{2pt} 
\begin{table*}[t]
\centering
{\footnotesize
    \begin{tabular}{|c|c|c|c|c|c|c|c|c|c|c|}
    \hline
    & \multicolumn{2}{!{\vrule width 2pt}c|}{$e$ = 60-70$\%$} & \multicolumn{2}{!{\vrule width 2pt}c|}{$e$ = 70-80$\%$} &\multicolumn{2}{!{\vrule width 2pt}c|}{$e$ = 80-90$\%$} &\multicolumn{2}{!{\vrule width 2pt}c|}{$e$ = 90-95$\%$} &\multicolumn{2}{!{\vrule width 2pt}c|}{$e$ = 95-99$\%$} \\
    \cline{2-11}
    & \multicolumn{1}{!{\vrule width 2pt}c|}{\textbf{Classic}} & \makecell{\textbf{GDC-} \\ \textbf{Filtered}} & \multicolumn{1}{!{\vrule width 2pt}c|}{\textbf{Classic}} & \makecell{\textbf{GDC-} \\ \textbf{Filtered}} & \multicolumn{1}{!{\vrule width 2pt}c|}{\textbf{Classic}} & \makecell{\textbf{GDC-} \\ \textbf{Filtered}}  & \multicolumn{1}{!{\vrule width 2pt}c|}{\textbf{Classic}} &\makecell{\textbf{GDC-} \\ \textbf{Filtered}} & \multicolumn{1}{!{\vrule width 2pt}c|}{\textbf{Classic}} &\makecell{\textbf{GDC-} \\ \textbf{Filtered}}  \\
    \hline
    \makecell{$\#$ of RANSAC iterations (99$\%$ success rate)} & \multicolumn{1}{!{\vrule width 2pt}c|}{169} & 169$\to$44 & \multicolumn{1}{!{\vrule width 2pt}c|}{420} & 420$\to$85 & \multicolumn{1}{!{\vrule width 2pt}c|}{3752} & 3752$\to$533 & \multicolumn{1}{!{\vrule width 2pt}c|}{21375} & 21375$\to$876 & \multicolumn{1}{!{\vrule width 2pt}c|}{681274} & 681274$\to$14374  \\
    \hline
    Hypothesis formation cost (ms) & \multicolumn{1}{!{\vrule width 2pt}c|}{0.16} & 0.94 & \multicolumn{1}{!{\vrule width 2pt}c|}{0.40} & 2.35 & \multicolumn{1}{!{\vrule width 2pt}c|}{3.60} & 21.04 & \multicolumn{1}{!{\vrule width 2pt}c|}{20.52} & 119.91 & \multicolumn{1}{!{\vrule width 2pt}c|}{654.02} & 3821.95\\
    \hline
    Hypothesis support measurement cost (ms) & \multicolumn{1}{!{\vrule width 2pt}c|}{7.64} & 1.988 & \multicolumn{1}{!{\vrule width 2pt}c|}{18.98} & 3.84 & \multicolumn{1}{!{\vrule width 2pt}c|}{169.59} & 24.09 & \multicolumn{1}{!{\vrule width 2pt}c|}{966.15} & 35.60 & \multicolumn{1}{!{\vrule width 2pt}c|}{30793.58} & 649.72\\
    \hline
    \hline
    \rowcolor{yellow}\textbf{Total Cost (ms) TUM-RGBD} & \multicolumn{1}{!{\vrule width 2pt}c|}{7.80} & 2.94 & \multicolumn{1}{!{\vrule width 2pt}c|}{19.39} & 6.19 & \multicolumn{1}{!{\vrule width 2pt}c|}{173.19} & 45.14 & \multicolumn{1}{!{\vrule width 2pt}c|}{986.67} & 155.513 & \multicolumn{1}{!{\vrule width 2pt}c|}{31447.61} & 3822.60 \\
    \hline
    \hline
    \makecell{$\#$ of RANSAC iterations (99$\%$ success rate)} & \multicolumn{1}{!{\vrule width 2pt}c|}{199} & 199$\to$ 42 & \multicolumn{1}{!{\vrule width 2pt}c|}{400} & 400$\to$64 & \multicolumn{1}{!{\vrule width 2pt}c|}{2619} &2619 $\to$314 & \multicolumn{1}{!{\vrule width 2pt}c|}{17679} &17679 $\to$ 1922& \multicolumn{1}{!{\vrule width 2pt}c|}{6.72e+5} &6.72e+5 $\to$ 2896 \\
    \hline
    Hypothesis formation cost (ms) & \multicolumn{1}{!{\vrule width 2pt}c|}{0.19} & 1.11 & \multicolumn{1}{!{\vrule width 2pt}c|}{0.38} & 2.24  & \multicolumn{1}{!{\vrule width 2pt}c|}{2.51} & 14.66 & \multicolumn{1}{!{\vrule width 2pt}c|}{16.97} &99.01  & \multicolumn{1}{!{\vrule width 2pt}c|}{645.12} & 3763.2 \\
    \hline
    Hypothesis support measurement cost (ms) & \multicolumn{1}{!{\vrule width 2pt}c|}{8.99} & 1.89 & \multicolumn{1}{!{\vrule width 2pt}c|}{18.0} & 2.89 & \multicolumn{1}{!{\vrule width 2pt}c|}{118.37} & 14.19 & \multicolumn{1}{!{\vrule width 2pt}c|}{799.09} & 86.87 & \multicolumn{1}{!{\vrule width 2pt}c|}{30374} &130.89 \\
    \hline
    \hline
    \rowcolor{yellow}\textbf{Total Cost (ms) ICL-NUIM} & \multicolumn{1}{!{\vrule width 2pt}c|}{9.18} & 3.00 & \multicolumn{1}{!{\vrule width 2pt}c|}{18.38} & 5.13 & \multicolumn{1}{!{\vrule width 2pt}c|}{120.88} & 28.85 & \multicolumn{1}{!{\vrule width 2pt}c|}{816.06} & 185.88 & \multicolumn{1}{!{\vrule width 2pt}c|}{31019.52} & 3894.09\\
    \hline
    \hline
    \makecell{$\#$ of RANSAC iterations (99$\%$ success rate)} & \multicolumn{1}{!{\vrule width 2pt}c|}{91} & 91$\to$32 & \multicolumn{1}{!{\vrule width 2pt}c|}{340} & 340$\to$85 & \multicolumn{1}{!{\vrule width 2pt}c|}{1412} & 1412$\to$183 & \multicolumn{1}{!{\vrule width 2pt}c|}{9534} & 9534$\to$614 & \multicolumn{1}{!{\vrule width 2pt}c|}{1.36e+5} & 1.36e+5$\to$2305  \\
    \hline
    Hypothesis formation cost (ms) & \multicolumn{1}{!{\vrule width 2pt}c|}{0.09} & 0.51 & \multicolumn{1}{!{\vrule width 2pt}c|}{0.33} & 1.91 & \multicolumn{1}{!{\vrule width 2pt}c|}{1.36} & 7.92 & \multicolumn{1}{!{\vrule width 2pt}c|}{9.15} & 53.49 & \multicolumn{1}{!{\vrule width 2pt}c|}{130.56} & 762.96\\
    \hline
    Hypothesis support measurement cost (ms) & \multicolumn{1}{!{\vrule width 2pt}c|}{4.11} & 1.45 & \multicolumn{1}{!{\vrule width 2pt}c|}{15.37} & 3.84 & \multicolumn{1}{!{\vrule width 2pt}c|}{63.82} & 8.27 & \multicolumn{1}{!{\vrule width 2pt}c|}{430.94} & 27.75 & \multicolumn{1}{!{\vrule width 2pt}c|}{6147.2} & 104.19\\
    \hline
    \hline
    \rowcolor{yellow}\textbf{Total Cost (ms) RGBD Scene v2} & \multicolumn{1}{!{\vrule width 2pt}c|}{4.2} & 1.96 & \multicolumn{1}{!{\vrule width 2pt}c|}{15.69} & 5.75 & \multicolumn{1}{!{\vrule width 2pt}c|}{65.18} & 16.19 & \multicolumn{1}{!{\vrule width 2pt}c|}{440.09} & 81.24 & \multicolumn{1}{!{\vrule width 2pt}c|}{6277.76} & 867.82 \\
    \hline
    \end{tabular}
    \caption{Cost of unfiltered (traditional RANSAC) and filtered RANSAC (GDC constraints applied) for 99$\%$ success rate over the entire TUM-RGBD, ICL-NUIM, and RGBD Scene v2 datasets, with a grand total of 132,946, 38,085 and 39,325 image pairs, respectively. GDC-Filtered columns are the number of RANSAC iterations and the number of hypothesis passing the GDC test.} 
    \label{tab:GDC_Profiling_ICL_RGBD_Scene_v2}
}
\end{table*}

\section{Time Savings Over the Classic RANSAC}

The time savings from applying \emph{(i)} the GDC filter, \emph{(ii)} the nested RANSAC loops, and \emph{(iii)} the nested GDC filer, over the classic RANSAC which were shown for the TUM-RGBD dataset in the main paper, are now shown for the ICL-NUIM and the RGBD Scene v2 datasets, Tables~\ref{tab:GDC_Profiling_ICL_RGBD_Scene_v2},~\ref{tab:nested_RANSAC_time_profiling_ICL_RGBD_Scene_v2}, and~\ref{tab:supp_GDC_nested_RANSAC_time_profiling}. The hypotheses are selected from the top $M = 250$ from the rank-ordered list of correspondences in the GDC filter RANSAC, Table~\ref{tab:GDC_Profiling_ICL_RGBD_Scene_v2}, while $M_1 \geq 100$, $M_2 \geq 150$, and $M_3 \geq 250$ are used for the nested RANSAC and the nested GDC RANSAC, Tables~\ref{tab:nested_RANSAC_time_profiling_ICL_RGBD_Scene_v2} and~\ref{tab:supp_GDC_nested_RANSAC_time_profiling}. The trend of the time savings presented in the tables for the TUM-RGBD dataset shown in the main paper, are consistent for the other two datasets shown here. Notice how the speedup grows significantly as the outlier ratio increases, especially for 95\% to 99\%, since the required hypotheses support measurement cost is much less than the classic RANSAC.  

\begin{figure}[!htbp]
    \centering
    (a)\includegraphics[width=.2\textwidth]{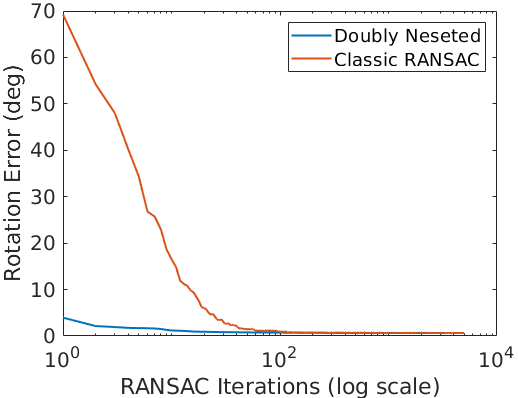}
    (b)\includegraphics[width=.2\textwidth]{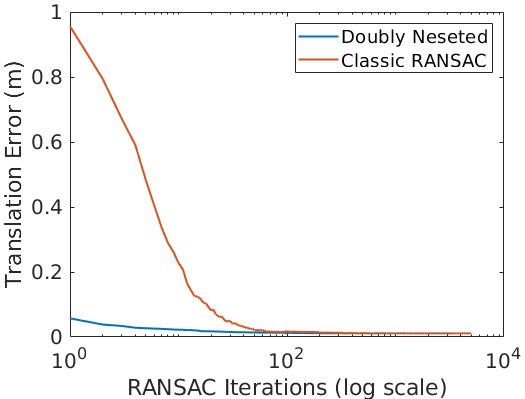}
    \caption{(a) Rotation error and (b) translation error over the RANSAC iterations in log scale using the classic and the doubly nested RANSAC on the TUM-RGBD dataset. Evidently, when fixing the number of RANSAC iterations in a typical visual odometry pipeline, our method gives better pose estimation accuracy than the classic RANSAC. }
    \label{fig:rate_of_estimation_err}
\end{figure}

Two aspects can be observed from the tables: \emph{(i)} both the GDC and the nested RANSAC reduce the total cost in all three datasets, but GDC has the extent of time savings more than nested RANSAC, \emph{e.g.}, for 95\%-95\% outlier ratio of the TUM-RGBD dataset, GDC has around 6.3$\times$ speedup while double nested gives only around 3.1$\times$ speedup; \emph{(ii)} only the nested RANSAC helps reducing the required RANSAC iterations to achieve equal success rate as the classic RANSAC does. Our approach thus gives not only significant improvement in efficiency, but the accuracy is also improved as the likelihood of picking promising hypothesis from the doubly nested RANSAC is higher than the classic RANSAC, Figure~\ref{fig:rate_of_estimation_err}. As a result, when fixing a certain RANSAC iterations in the visual odometry pipeline, \emph{e.g.}, 100 in CVO-SLAM~\cite{lin2023robust}, the proposed method provides more accurate estimations. Experiments on the accuracy are demonstrated in the next section.

\setlength\tabcolsep{1.5pt} 
\begin{table*}[t]
\centering
{\footnotesize
    \begin{tabular}{|c|c|c|c|c|c|c|c|c|c|c|c|c|c|c|c|}
    \hline
    & \multicolumn{3}{!{\vrule width 2pt}c|}{$e$ = 60-70$\%$} & \multicolumn{3}{!{\vrule width 2pt}c|}{$e$ = 70-80$\%$} & \multicolumn{3}{!{\vrule width 2pt}c|}{$e$ = 80-90$\%$} & \multicolumn{3}{!{\vrule width 2pt}c|}{$e$ = 90-95$\%$} & \multicolumn{3}{!{\vrule width 2pt}c|}{$e$ = 95-99$\%$} \\
    \cline{2-16}
    & \multicolumn{1}{!{\vrule width 2pt}c|}{{\rotatebox{90}{\textbf{Classic}}}} & {\rotatebox{90}{\textbf{Nested}}} & {\rotatebox{90}{\textbf{\makecell{Doubly\\ Nested}}\;}} & \multicolumn{1}{!{\vrule width 2pt}c|}{{\rotatebox{90}{\textbf{Classic}}}} & {\rotatebox{90}{\textbf{Nested}}} & {\rotatebox{90}{\textbf{\makecell{Doubly\\ Nested}}}} & \multicolumn{1}{!{\vrule width 2pt}c|}{{\rotatebox{90}{\textbf{Classic}}}} & {\rotatebox{90}{\textbf{Nested}}} & {\rotatebox{90}{\textbf{\makecell{Doubly\\ Nested}}}} & \multicolumn{1}{!{\vrule width 2pt}c|}{{\rotatebox{90}{\textbf{Classic}}}} & {\rotatebox{90}{\textbf{Nested}}} & {\rotatebox{90}{\textbf{\makecell{Doubly\\ Nested}}}} & \multicolumn{1}{!{\vrule width 2pt}c|}{{\rotatebox{90}{\textbf{Classic}}}} & {\rotatebox{90}{\textbf{Nested}}} & {\rotatebox{90}{\textbf{\makecell{Doubly\\ Nested}}}} \\
    % \hline
    % \makecell{$\#$ of matches from top rank-\\ordered list: $M_1$/$M_2$/$M_3$} & \multicolumn{1}{!{\vrule width 2pt}c|}{250} &  \makecell{100/\\250} & \makecell{100/\\150/\\250}& \multicolumn{1}{!{\vrule width 2pt}c|}{250} & \makecell{100/\\250}& \makecell{100/\\150/\\250} & \multicolumn{1}{!{\vrule width 2pt}c|}{250} & \makecell{100/\\250}& \makecell{100/\\150/\\250}& \multicolumn{1}{!{\vrule width 2pt}c|}{250} & \makecell{100/\\250} & \makecell{100/\\150/\\250} & \multicolumn{1}{!{\vrule width 2pt}c|}{250} &  \makecell{100/\\250} & \makecell{100/\\150/\\250} \\
    \hline
    \makecell{$\#$ of RANSAC iterations (99$\%$ success rate)} & \multicolumn{1}{!{\vrule width 2pt}c|}{169} & 146 & 137& \multicolumn{1}{!{\vrule width 2pt}c|}{420} & 371 & 227 & \multicolumn{1}{!{\vrule width 2pt}c|}{3752} & 2218 & 2126 & \multicolumn{1}{!{\vrule width 2pt}c|}{21375} & 17509 & 6872 & \multicolumn{1}{!{\vrule width 2pt}c|}{681274} & 220637 &68824  \\
    \hline
    \hline
    \rowcolor{yellow}\textbf{Total Cost (ms) TUM-RGBD} & \multicolumn{1}{!{\vrule width 2pt}c|}{7.80} & {6.73} & {6.32} & \multicolumn{1}{!{\vrule width 2pt}c|}{19.38} & {17.1} & {10.46} & \multicolumn{1}{!{\vrule width 2pt}c|}{173.19} &{102.25} & {98} & \multicolumn{1}{!{\vrule width 2pt}c|}{986.67} & {808.21} & { 317.21} & \multicolumn{1}{!{\vrule width 2pt}c|}{31447.61} & {10184.60} & {3176.92} \\
    \hline
    \hline
    \makecell{$\#$ of RANSAC iterations (99$\%$ success rate)} & \multicolumn{1}{!{\vrule width 2pt}c|}{199} & 124  &88 & \multicolumn{1}{!{\vrule width 2pt}c|}{400} & 340 & 251 & \multicolumn{1}{!{\vrule width 2pt}c|}{2619} & 1704 & 1260 & \multicolumn{1}{!{\vrule width 2pt}c|}{17679} & 8876 & 4634 & \multicolumn{1}{!{\vrule width 2pt}c|}{6.72e+5} & 2.48e+5 & 98461  \\
    \hline
    \hline
    \rowcolor{yellow}\textbf{Total Cost (ms) ICL-NUIM} & \multicolumn{1}{!{\vrule width 2pt}c|}{9.18} & 5.72 & 4.06 & \multicolumn{1}{!{\vrule width 2pt}c|}{18.38} & 15.69 &11.58  & \multicolumn{1}{!{\vrule width 2pt}c|}{120.88} & 78.65 & 58.16 & \multicolumn{1}{!{\vrule width 2pt}c|}{816.06} & 409.72 &213.91  & \multicolumn{1}{!{\vrule width 2pt}c|}{31019.52} & 11447.68 & 4544.98 \\
    \hline
    \hline
    \makecell{$\#$ of RANSAC iterations (99$\%$ success rate)} & \multicolumn{1}{!{\vrule width 2pt}c|}{91} & 69 & 42 & \multicolumn{1}{!{\vrule width 2pt}c|}{340} & 114 & 78 & \multicolumn{1}{!{\vrule width 2pt}c|}{1412} & 715 & 460 & \multicolumn{1}{!{\vrule width 2pt}c|}{9543} & 5325 & 3586 & \multicolumn{1}{!{\vrule width 2pt}c|}{1.36e+5} & 59586 & 17791 \\
    \hline
    \hline
    \rowcolor{yellow}\textbf{Total Cost (ms) RGBD Scene v2} & \multicolumn{1}{!{\vrule width 2pt}c|}{4.2} & 3.19 & 1.94 & \multicolumn{1}{!{\vrule width 2pt}c|}{15.69} & 5.26 & 3.60 & \multicolumn{1}{!{\vrule width 2pt}c|}{65.18} & 33.00  & 21.23 & \multicolumn{1}{!{\vrule width 2pt}c|}{440.09} & 245.80 & 165.53 & \multicolumn{1}{!{\vrule width 2pt}c|}{6277.76} & 2750.49 & 821.23 \\
    \hline
    \end{tabular}
    \caption{Cost of traditional, nested, and doubly nested RANSAC for 99$\%$ success rate over the TUM-RGBD, ICL-NUIM and RGBD Scene v2 datasets. The number of correspondences from the top rank-ordered list is $M_1 = 100$, $M_2 = 150$, and $M_3 = 250$.} 
    \label{tab:nested_RANSAC_time_profiling_ICL_RGBD_Scene_v2}
}
\end{table*}

\setlength\tabcolsep{1pt} 
\begin{table*}[!htbp]
\centering
{\footnotesize
    \begin{tabular}{|c|c|c|c|c|c|c|c|c|c|c|c|c|c|c|c|c|c|c|c|c|}
    \hline
    & \multicolumn{4}{!{\vrule width 2pt}c|}{$e$ = 60-70$\%$} & \multicolumn{4}{!{\vrule width 2pt}c|}{$e$ = 70-80$\%$} &\multicolumn{4}{!{\vrule width 2pt}c|}{$e$ = 80-90$\%$} &\multicolumn{4}{!{\vrule width 2pt}c|}{$e$ = 90-95$\%$} &\multicolumn{4}{!{\vrule width 2pt}c|}{$e$ = 95-99$\%$} \\
    \cline{2-21}
    & \multicolumn{1}{!{\vrule width 2pt}c|}{{\rotatebox{90}{\textbf{Classic}}}} & {\rotatebox{90}{\textbf{GDC}}} & {\rotatebox{90}{\textbf{\makecell{GDC Nested}}}} & {\rotatebox{90}{\textbf{\makecell{GDC Doubly\\ Nested}}\;}} & \multicolumn{1}{!{\vrule width 2pt}c|}{{\rotatebox{90}{\textbf{Classic}}}} & {\rotatebox{90}{\textbf{GDC}}} & {\rotatebox{90}{\textbf{\makecell{GDC Nested}}}} & {\rotatebox{90}{\textbf{\makecell{GDC Doubly\\Nested}}\;}} & \multicolumn{1}{!{\vrule width 2pt}c|}{{\rotatebox{90}{\textbf{Classic}}}} & {\rotatebox{90}{\textbf{GDC}}} & {\rotatebox{90}{\textbf{\makecell{GDC Nested}}}} & {\rotatebox{90}{\textbf{\makecell{GDC Doubly\\ Nested}}\;}} & \multicolumn{1}{!{\vrule width 2pt}c|}{{\rotatebox{90}{\textbf{Classic}}}} & {\rotatebox{90}{\textbf{GDC}}} & {\rotatebox{90}{\textbf{\makecell{GDC Nested}}}} & {\rotatebox{90}{\textbf{\makecell{GDC Doubly\\ Nested}}\;}} & \multicolumn{1}{!{\vrule width 2pt}c|}{{\rotatebox{90}{\textbf{Classic}}}} & {\rotatebox{90}{\textbf{GDC}}} & {\rotatebox{90}{\textbf{\makecell{GDC Nested}}}} & {\rotatebox{90}{\textbf{\makecell{GDC Doubly\\ Nested}}\;}} \\
    % \hline
    % \makecell{$\#$ of matches from \\ top rank-ordered \\ list: $M_1$/$M_2$/$M_3$} & \multicolumn{1}{!{\vrule width 2pt}c|}{250} & 250 & \makecell{100/\\250} & \makecell{100/\\150/\\250} & \multicolumn{1}{!{\vrule width 2pt}c|}{250} & 250 & \makecell{100/\\250} & \makecell{100/\\150/\\250} & \multicolumn{1}{!{\vrule width 2pt}c|}{250} & 250 & \makecell{100/\\250} & \makecell{100/\\150/\\250}  & \multicolumn{1}{!{\vrule width 2pt}c|}{250} & 250 & \makecell{100/\\250} & \makecell{100/\\150/\\250} & \multicolumn{1}{!{\vrule width 2pt}c|}{250} & 250 & \makecell{100/\\250} & \makecell{100/\\150/\\250} \\
    \hline
    \makecell{$\#$ of RANSAC \\ iterations \\ (99$\%$ success rate)} & \multicolumn{1}{!{\vrule width 2pt}c|}{169} & \makecell{169\\$\downarrow$\\44} & \makecell{146\\$\downarrow$\\35} & \makecell{137\\$\downarrow$\\32} & \multicolumn{1}{!{\vrule width 2pt}c|}{420} & \makecell{420\\$\downarrow$\\85} & \makecell{371\\$\downarrow$\\76} & \makecell{227\\$\downarrow$\\53} & \multicolumn{1}{!{\vrule width 2pt}c|}{3752} & \makecell{3752\\$\downarrow$\\533} & \makecell{2218\\$\downarrow$\\272} & \makecell{2126\\$\downarrow$\\240} & \multicolumn{1}{!{\vrule width 2pt}c|}{21375} & \makecell{21375\\$\downarrow$\\876}& \makecell{17509\\$\downarrow$\\916} & \makecell{6872\\$\downarrow$\\340} & \multicolumn{1}{!{\vrule width 2pt}c|}{681274} & \makecell{681274\\$\downarrow$\\14374}& \makecell{220637\\$\downarrow$\\9708} & \makecell{68824\\$\downarrow$\\1446} \\
    \hline
    \hline
    \rowcolor{yellow}{\textbf{\Gape[0pt][0pt]{\makecell{Total Cost (ms) \\ TUN-RGBD}}}} & \multicolumn{1}{!{\vrule width 2pt}c|}{7.8} & 2.9 & 2.4 & 2.2 & \multicolumn{1}{!{\vrule width 2pt}c|}{19.4} & 6.2 & 5.5 & 3.7 & \multicolumn{1}{!{\vrule width 2pt}c|}{173.2} & 45.1 & 24.7 & 22.8 & \multicolumn{1}{!{\vrule width 2pt}c|}{986.7} & 155.5 & 139.6 & 53.9 & \multicolumn{1}{!{\vrule width 2pt}c|}{31447.6} & 3822.6 & 1676.6 & 451.5 \\
    \hline
    \hline
    \makecell{$\#$ of RANSAC \\ iterations \\ (99$\%$ success rate)} & \multicolumn{1}{!{\vrule width 2pt}c|}{199} & \makecell{199\\$\downarrow$\\42} & \makecell{124\\$\downarrow$\\36} & \makecell{88\\$\downarrow$\\32} & \multicolumn{1}{!{\vrule width 2pt}c|}{400} & \makecell{400\\$\downarrow$\\64} & \makecell{340\\$\downarrow$\\51} & \makecell{251\\$\downarrow$\\39} & \multicolumn{1}{!{\vrule width 2pt}c|}{2619} & \makecell{2619\\$\downarrow$\\314} & \makecell{1704\\$\downarrow$\\168} & \makecell{1260\\$\downarrow$\\114} & \multicolumn{1}{!{\vrule width 2pt}c|}{17679} & \makecell{17679\\$\downarrow$\\1922} & \makecell{8876\\$\downarrow$\\1021} & \makecell{4632\\$\downarrow$\\688} & \multicolumn{1}{!{\vrule width 2pt}c|}{6.72e+5} & \makecell{6.72e+5\\$\downarrow$\\2897} & \makecell{2.48e+5\\$\downarrow$\\1569} & \makecell{98461\\$\downarrow$\\1014} \\
    \hline
    \hline
    \rowcolor{yellow}{\textbf{\Gape[0pt][0pt]{\makecell{Total Cost (ms) \\ ICL-NUIM}}}} & \multicolumn{1}{!{\vrule width 2pt}c|}{9.18} &3.00  &  2.32& 1.93 & \multicolumn{1}{!{\vrule width 2pt}c|}{18.18} & 5.13 & 4.21 & 3.17 & \multicolumn{1}{!{\vrule width 2pt}c|}{120.88} & 28.85 & 17.13 & 12.21 & \multicolumn{1}{!{\vrule width 2pt}c|}{816.06} & 185.88 &95.85  & 57.04 & \multicolumn{1}{!{\vrule width 2pt}c|}{31019.52} & 3894.09 & 1459.71 &597.21  \\
    \hline
    \hline
    \makecell{$\#$ of RANSAC \\ iterations \\ (99$\%$ success rate)} & \multicolumn{1}{!{\vrule width 2pt}c|}{91} & \makecell{91\\$\downarrow$\\32} & \makecell{69\\$\downarrow$\\28} & \makecell{42\\$\downarrow$\\24} & \multicolumn{1}{!{\vrule width 2pt}c|}{340} & \makecell{340\\$\downarrow$\\85} & \makecell{114\\$\downarrow$\\55} & \makecell{78\\$\downarrow$\\53} & \multicolumn{1}{!{\vrule width 2pt}c|}{1412} & \makecell{1412\\$\downarrow$\\183} & \makecell{715\\$\downarrow$\\163} & \makecell{460\\$\downarrow$\\132} & \multicolumn{1}{!{\vrule width 2pt}c|}{9543} & \makecell{9543\\$\downarrow$\\614}& \makecell{5325\\$\downarrow$\\717} & \makecell{3586\\$\downarrow$\\642} & \multicolumn{1}{!{\vrule width 2pt}c|}{1.36e+5} & \makecell{1.36e+5\\$\downarrow$\\2305}& \makecell{59586\\$\downarrow$\\7782} & \makecell{17791\\$\downarrow$\\5484} \\
    \hline
    \hline
    \rowcolor{yellow}{\textbf{\Gape[0pt][0pt]{\makecell{Total Cost (ms) \\ RGBD Scene v2}}}} & \multicolumn{1}{!{\vrule width 2pt}c|}{4.2} & 1.96 & 1.65 & 1.32 & \multicolumn{1}{!{\vrule width 2pt}c|}{15.69} & 5.75 & 3.13 & 2.83 & \multicolumn{1}{!{\vrule width 2pt}c|}{65.18} & 16.19 & 11.38 & 8.55 & \multicolumn{1}{!{\vrule width 2pt}c|}{440.09} & 81.24 & 62.28 & 49.14 & \multicolumn{1}{!{\vrule width 2pt}c|}{6277.76} & 867.82 & 686.02 & 347.68\\
    \hline
    \end{tabular}
    \caption{A comparison of timings for classic RANSAC, GDC-Filtered RANSAC, nested RANSAC, and doubly nested RANSAC for the TUM-RGBD, ICL-NUIM, and RGBD Scene v2 datasets. Note that the change in the number of RANSAC iterations indicates the number of hypothesis passing the GDC test.} 
    \label{tab:supp_GDC_nested_RANSAC_time_profiling}
}
\end{table*}

\section{Relative Pose Estimation Accuracy Against Existing Methods}

Experiments evaluate RGBD relative pose estimation accuracy in terms of relative pose error (RPE) for translation and for rotation, comparing the proposed method with the existing VO/SLAM pipelines were shown in the main paper. They are now extended in two ways. Tables~\ref{tab:RPE_trans_comparison_supp_tum}-\ref{tab:RPE_rots_comparison_supp_RGBD_Scene_v2}: \emph{(i)} four additional sequences of the TUM-RGBD dataset and all sequences of the RGBD Scene v2 dataset are included for a complete comparisons, \emph{(ii)} two very recent algorithms, CVO-SLAM~\cite{lin2023robust} and PLP-SLAM~\cite{shu2023structure} are added to the list of methods for comparisons. Evaluation values of ORB SLAM2~\cite{mur2017orb}, KinectFusion~\cite{izadi2011kinectfusion}, RGBD DVO~\cite{cai2021direct}, Canny VO~\cite{zhou2018canny}, and RGBD DSO~\cite{yuan2021rgb}, are taken from their papers or from the third party evaluations, \emph{e.g.}~\cite{yuan2021rgb}. For the rest of the methods, \emph{i.e.}, CVO~\cite{ghaffari2019continuous}, ACO~\cite{lin2019adaptive}, Edge DVO~\cite{christensen2019edge}, CVO-SLAM~\cite{lin2023robust}, and PLP-SLAM~\cite{shu2023structure}, their source code is used on the datasets with their default parameter settings. Note that for CVO-SLAM~\cite{lin2023robust} and PLP-SLAM~\cite{shu2023structure}, we turn off the loop closure detection and global bundle adjustment and leave only the visual odometry mode. Evaluation results of CVO~\cite{ghaffari2019continuous}, ACO~\cite{lin2019adaptive}, Edge DVO~\cite{christensen2019edge}, CVO-SLAM~\cite{lin2023robust}, PLP-SLAM~\cite{shu2023structure}, and our method are averaged over 10 runs, if otherwise specified.

Overall, the proposed method has competitive performances against the existing contemporary approaches which typically contain not only the RGBD relative pose estimation module, but they used additional optimization through a local bundle adjustment refinement. Our results shows that the nested GDC RANSAC is sufficient to give nearly optimal pose estimations, even without refinement. In particular, for the RGBD Scene v2 dataset which exhibits high outlier ratio scenarios, our method performs either the best or the second best among all the competing algorithms.

\setlength\tabcolsep{2pt}
\begin{table}[!htbp]
\centering
\footnotesize
\begin{tabular}
{lcccccccc}
Methods & {\rotatebox{0}{\makecell{fr1/\\desk}}} & {\rotatebox{0}{\makecell{fr1/\\room}}} & {\rotatebox{0}{\makecell{fr1/\\xyz}}} & {\rotatebox{0}{\makecell{fr2/\\desk}}} & {\rotatebox{0}{\makecell{fr3/\\struct}}} & {\rotatebox{0}{\makecell{fr3/\\office}}} \\
\hline
ORB SLAM2$^{\star}$~\cite{mur2017orb}  &2.00&-&-&-&-&0.83 \\
KinectFusion$^{\star}$~\cite{izadi2011kinectfusion} &34.43&-&-&-&-&21.32\\
RGBD DVO$^{\star}$~\cite{cai2021direct}&1.3&-&-&-&-&-\\
Canny VO$^{\star}$~\cite{zhou2018canny}&5.1 &- &- &- &- &1.9 \\
RGBD DSO$^{\star}$$^{\lozenge}$~\cite{yuan2021rgb} &{\bf 0.12} & -&- & -&- &0.56 \\
CVO~\cite{ghaffari2019continuous} & {\underline {0.43}} & {\bf 0.56} & 0.84 & {\underline{0.45}} & {\bf 1.29} & {\bf 0.42} \\
ACO~\cite{lin2019adaptive} & 1.00 & {\bf 0.56} & 0.88 & 0.49 & 1.59 & \underline{0.47} \\
Edge DVO~\cite{christensen2019edge} & 17.32 & $\times$ & 1.57 & 1.34 & 1.63 & 1.04 \\
CVO-SLAM$^\dag$~\cite{lin2023robust} & 1.09 & \underline{0.63} & {\bf 0.43} & {\bf 0.37} & {1.43} & \underline{0.47} \\
PLP-SLAM$^\dag$~\cite{shu2023structure} & 1.07 & 1.68 & 4.86 & 3.43 & 2.56 & 4.12 \\
Our Method & 1.05 & 0.78 & {\underline{0.56}}&0.91 & \underline{1.38} & 0.75 \\
\hline
\multicolumn{7}{l}{\textbf{Boldfaced:} the best. \quad \underline{Underlined:} the second best. } \\
\multicolumn{7}{l}{$\star$: Values taken from their original papers, or from~\cite{yuan2021rgb}.} \\
\multicolumn{7}{l}{$\lozenge$: No depth refinement and occlusion removal modules.} \\
% \multicolumn{7}{l}{-: Not available from the original paper or the third parties.} \\
\multicolumn{7}{l}{$\times$: Estimations diverged. \quad -: Values unavailable.} \\
\multicolumn{7}{l}{$^\dag$: Loop closure and global bundle adjustment are turned off.} \\
% \multicolumn{7}{l}{$^\ddag$: Point + Line configuration.} \\
\hline
\end{tabular}
\caption{$\text{RPE}_\text{trans}$ (cm) comparisons on the selected sequences of the TUM-RGBD dataset.}
\label{tab:RPE_trans_comparison_supp_tum}
\end{table}

\setlength\tabcolsep{2pt}
\begin{table}[!htbp]
\centering
\footnotesize
\begin{tabular}
{lcccccccc}
Methods & {\rotatebox{0}{\makecell{fr1/\\desk}}} & {\rotatebox{0}{\makecell{fr1/\\room}}} & {\rotatebox{0}{\makecell{fr1/\\xyz}}} & {\rotatebox{0}{\makecell{fr2/\\desk}}} & {\rotatebox{0}{\makecell{fr3/\\struct}}} & {\rotatebox{0}{\makecell{fr3/\\office}}} \\
\hline
ORB SLAM2$^{\star}$~\cite{mur2017orb} &0.94 &- &- & - & -& 1.25 \\
KinectFusion$^{\star}$~\cite{izadi2011kinectfusion} &3.09&-&-&-&-&8.00 \\
RGBD DVO$^{\star}$~\cite{cai2021direct} &1.75&-&-&-&-&- \\
Canny VO$^{\star}$~\cite{zhou2018canny}&2.39 &- &- &- &- &0.91 \\
RGBD DSO$^{\star}$$^{\lozenge}$~\cite{yuan2021rgb} &{\bf 0.32} & -&- & -&- &{\bf 0.23} \\
CVO~\cite{ghaffari2019continuous} & \underline{0.37} & 0.41 & \underline{0.37} & 0.85 & \underline{0.77} & 0.37 \\
ACO~\cite{lin2019adaptive} & 0.59 & {\bf 0.39} & 1.12 & 0.57 & 0.83 & 0.35 \\
Edge DVO~\cite{christensen2019edge} & 15.17 & $\times$ & 5.37 & 2.76 & 0.98 & 0.56 \\
CVO-SLAM$^\dag$~\cite{lin2023robust} & 0.75 & 0.43 & 0.41 & {\bf 0.31} & 0.83 & \underline{0.29} \\
PLP-SLAM$^\dag$~\cite{shu2023structure} & 0.84 & 2.57 & 1.32 & 3.93 & 3.66 & 1.93 \\
Our Method & 0.59 & \underline{0.40} & {\bf 0.36} &\underline{0.49} & {\bf 0.76} & 0.32 \\
\hline
\multicolumn{7}{l}{\textbf{Boldfaced:} the best. \quad \underline{Underlined:} the second best. } \\
\multicolumn{7}{l}{$\star$: Values taken from the original papers, or from~\cite{yuan2021rgb}.} \\
\multicolumn{7}{l}{$\lozenge$: No depth refinement and occlusion removal modules.} \\
% \multicolumn{7}{l}{-: Not available from the original paper or the third parties.} \\
\multicolumn{7}{l}{$\times$: Estimations diverged. \quad -: Values unavailable.} \\
\multicolumn{7}{l}{$^\dag$: Loop closure and global bundle adjustment are turned off.} \\
% \multicolumn{7}{l}{$^\ddag$: Point + Line configuration.} \\
\hline
\end{tabular}
\caption{$\text{RPE}_\text{rot}$ (degree) comparisons on the selected sequences of the TUM-RGBD dataset.}
\label{tab:RPE_rots_comparison_supp}
\end{table}

\setlength\tabcolsep{1.5pt}
\begin{table}[!htbp]
\centering
\footnotesize
\begin{tabular}
{lcccccccc}
Methods & {\rotatebox{40}{lr kt0}} & {\rotatebox{40}{lr kt1}} & {\rotatebox{40}{lr kt2}} & {\rotatebox{40}{lr kt3}} & {\rotatebox{40}{of kt0}} & {\rotatebox{40}{of kt1}} & {\rotatebox{40}{of kt2}} & {\rotatebox{40}{of kt3}} \\
% Methods & {lr kt0} & {lr kt1} & {lr kt2} & {lr kt3} & {of kt0} & {of kt1} & {of kt2} & {of kt3} \\
\hline
ORB SLAM2$^{\star}$~\cite{mur2017orb}  &4.29&-&9.68&14.35&6.00&16.53&6.40&25.42\\
KinectFusion$^{\star}$~\cite{izadi2011kinectfusion} &32.17&10.05&5.30&32.46&17.5&29.34&28.44&42.45\\
RGBD DVO$^{\star}$~\cite{cai2021direct} &-&0.78&3.28&3.30&1.27&0.77&2.65&2.07\\
Canny VO$^{\star}$~\cite{zhou2018canny}&-&0.9&1.1&0.7&-&-&-&-\\
RGBD DSO$^{\star}$$^{\lozenge}$~\cite{yuan2021rgb}&-&-&-&-&-&-&-&-\\
CVO~\cite{ghaffari2019continuous} & 2.14 & 3.36 & 3.24 & 2.65 & 1.46 & 2.26 & 3.00 & 1.82 \\
ACO~\cite{lin2019adaptive} & 2.19 & 2.46 & 3.12 & 2.79 & 1.59 & 2.13 & 3.36 & 1.76 \\
Edge DVO~\cite{christensen2019edge} & $\times$ & 1.51 & 3.68 & $\times$ & 1.95 & $\times$ & 2.46 & 1.14 \\
CVO-SLAM$^\dag$~\cite{lin2023robust} & \underline{0.55} & 1.86 & 0.64 & \underline{0.89} & \underline{0.64} & \textbf{0.39} & \textbf{0.68} & \textbf{0.33} \\
PLP-SLAM$^\dag$~\cite{shu2023structure} & 0.61 & \underline{0.97} & \underline{0.44} & 1.41 & 1.79 & 2.03 & \underline{0.71} & 1.12\\
Our Method & \textbf{0.37} & \textbf{0.39} & \textbf{0.38} & \textbf{0.35} & \textbf{0.58} & \underline{0.52} & 2.30 & \underline{0.44} \\
\hline
\multicolumn{9}{l}{\textbf{Boldfaced:} the best. \quad \underline{Underlined:} the second best. } \\
\multicolumn{9}{l}{$\star$: Values taken from their original papers, or from~\cite{yuan2021rgb}.} \\
\multicolumn{9}{l}{$\lozenge$: No depth refinement and occlusion removal modules.} \\
% \multicolumn{9}{l}{-: Not available from the original paper or the third parties.} \\
\multicolumn{9}{l}{$\times$: Estimations diverged. \quad -: Values unavailable.} \\
\multicolumn{9}{l}{$^\dag$: Loop closure and global bundle adjustment are turned off.} \\
% \multicolumn{9}{l}{$^\ddag$: Point + Line configuration.} \\
\hline
\end{tabular}
\caption{$\text{RPE}_\text{trans}$ (cm) comparisons on all sequences of the ICL-NUIM dataset.}
\label{tab:RPE_trans_comparison_supp_icl}
\end{table}

\setlength\tabcolsep{1.5pt}
\begin{table}[t]
\centering
\footnotesize
\begin{tabular}
{lcccccccc}
Methods & {\rotatebox{45}{lr kt0}} & {\rotatebox{45}{lr kt1}} & {\rotatebox{45}{lr kt2}} & {\rotatebox{45}{lr kt3}} & {\rotatebox{45}{of kt0}} & {\rotatebox{45}{of kt1}} & {\rotatebox{45}{of kt2}} & {\rotatebox{45}{of kt3}} \\
\hline
ORB SLAM2$^{\star}$~\cite{mur2017orb}  & 5.61&-&2.37&3.22&0.93&2.46&2.90&6.58\\
KinectFusion$^{\star}$~\cite{izadi2011kinectfusion} &9.12&1.20&1.37&9.98&1.16&1.23&2.93&1.16\\
RGBD DVO$^{\star}$~\cite{cai2021direct}&-& \underline{0.17} &0.91&0.56&0.24&0.26&1.03&0.34\\
Canny VO$^{\star}$~\cite{zhou2018canny}&-&0.21&0.27&\textbf{0.15} &-&-&-&-\\
RGBD DSO$^{\star}$$^{\lozenge}$~\cite{yuan2021rgb}&-&-&-&-&-&-&-&-\\
CVO~\cite{ghaffari2019continuous} & 0.55 & 0.49 & 0.57 & 0.47 & 0.54 & 0.51 & 1.54 & 0.40 \\
ACO~\cite{lin2019adaptive} & 0.55 & 0.48 & 0.56 & 0.48 & 0.51 & 0.49 & 0.55 & 0.39 \\
Edge DVO~\cite{christensen2019edge} & $\times$ & {0.18} & \underline{0.12} & $\times$ & \textbf{0.16} & $\times$ & 0.36 & $\times$ \\
CVO-SLAM$^\dag$~\cite{lin2023robust} & \textbf{0.14} & {0.18} & 0.78 & 0.35 & \underline{0.22} & \textbf{0.09} & \textbf{0.14} & \textbf{0.09} \\
PLP-SLAM$^\dag$~\cite{shu2023structure} & \underline{0.33} & 0.55 & 0.58 & 0.73 & 1.52 & 0.32 & \underline{0.25} & 1.23 \\
Our Method & \textbf{0.14} & \textbf{0.09} & \textbf{0.10} & \underline{0.16} & 0.32 & \underline{0.18} & 0.78 & \underline{0.12} \\
\hline
\multicolumn{9}{l}{\textbf{Boldfaced:} the best. \quad \underline{Underlined:} the second best. } \\
\multicolumn{9}{l}{$\star$: Values taken from their original papers, or from~\cite{yuan2021rgb}.} \\
\multicolumn{9}{l}{$\lozenge$: No depth refinement and occlusion removal modules.} \\
% \multicolumn{9}{l}{-: Not available from the original paper or the third parties.} \\
\multicolumn{9}{l}{$\times$: Estimations diverged. \quad -: Values unavailable.} \\
\multicolumn{9}{l}{$^\dag$: Loop closure and global bundle adjustment are turned off. } \\
% \multicolumn{9}{l}{$^\ddag$: Point + Line configuration.} \\
\hline
\end{tabular}
\caption{$\text{RPE}_\text{rot}$ (degree) comparisons on all sequences of ICL-NUIM dataset.}
\label{tab:RPE_rots_comparison_supp_icl}
\end{table}

\setlength\tabcolsep{1.2pt}
\begin{table}[!htbp]
\centering
\footnotesize
\begin{tabular}
{lcccccccccccccc}
Methods & {\rotatebox{45}{s01}} & {\rotatebox{45}{s02}} & {\rotatebox{45}{s03}} & {\rotatebox{45}{s04}} & {\rotatebox{45}{s05}} & {\rotatebox{45}{s06}} & {\rotatebox{45}{s07}} & {\rotatebox{45}{s08}} & {\rotatebox{45}{s09}} & {\rotatebox{45}{s10}} & {\rotatebox{45}{s11}} & {\rotatebox{45}{s12}} & {\rotatebox{45}{s13}} & {\rotatebox{45}{s14}} \\
\hline
ORB SLAM2$^{\star}$~\cite{mur2017orb} & - & - & - & - & 4.37 & 3.89 & 2.40 & 3.76 & - & - & - & - & - & - \\
KinectFusion$^{\star}$~\cite{izadi2011kinectfusion} &-&-&-&-&179&178&174&166&-&-&-&-&-&-\\
RGBD DVO$^{\star}$~\cite{cai2021direct}& - &-&-&-&11.4&15.5&12.4&11.8&-&-&-&-&-&-\\
Canny VO$^{\star}$~\cite{zhou2018canny} & - & -& -& -& -&-  & -& -& -& -&-&-&-&-  \\
RGBD DSO$^{\star}$$^{\lozenge}$~\cite{yuan2021rgb} & - & -&- &- &5.76&39.2 &2.8 & 5.56&- &- &- &-&-&- \\
CVO~\cite{ghaffari2019continuous} & 1.99 & 2.29 & 2.51 & 2.95 & 3.76 & 3.66 & 3.63 & 3.62 & 1.67 & 1.66 & 2.22 & 1.82 & 1.73 & 2.84 \\
ACO~\cite{lin2019adaptive} & 1.98 & 2.27 & 2.48 & 2.94 & 3.57 & 3.82 & 3.64 & 3.84 & 1.74 & 1.72 & 2.15 & 1.91 & 1.71 & 2.82 \\
Edge DVO~\cite{christensen2019edge} & 2.00 & 2.02 & 2.01 & 2.01  &2.98 & 2.97& 2.98&2.99 & 3.01& 3.00& 3.01 &3.00 &$\times$ &$\times$ \\
CVO-SLAM$^\dag$~\cite{lin2023robust} & \underline{0.69} & \underline{0.72} & \underline{0.77} & \underline{0.84} & \underline{0.97} & \underline{1.04} & \underline{1.02} & \underline{1.10} & \underline{0.72} & \underline{0.72} & \underline{0.78} & \textbf{0.68} & {\bf 0.50} & \underline{0.89} \\
PLP-SLAM$^\dag$~\cite{shu2023structure} & 2.09 & 2.78 & 3.69 & 2.55 & 2.81 & 3.11 & 2.32 & 3.33 & 1.82 & 1.30 & 2.49 & 1.45 & 2.89 & 3.06 \\
Our Method & {\bf 0.68} & {\bf 0.70} & \textbf{0.75} &{\bf 0.83} & \textbf{0.96}& \textbf{1.03}& \textbf{1.02}& \textbf{1.07} &{\bf0.71}& {\bf 0.70} & {\bf 0.75}& \underline{0.69}&\underline{0.61} & {\bf 0.87} \\
\hline
\multicolumn{15}{l}{\textbf{Boldfaced:} the best. \quad \underline{Underlined:} the second best. } \\
\multicolumn{15}{l}{$\star$: Values taken from their original papers, or from~\cite{yuan2021rgb}.} \\
\multicolumn{15}{l}{$\lozenge$: No depth refinement and occlusion removal modules.} \\
% \multicolumn{15}{l}{-: Not available from the original paper or the third parties.} \\
\multicolumn{15}{l}{$\times$: Estimations diverged. \quad -: Values unavailable.} \\
\multicolumn{15}{l}{$^\dag$: Loop closure and global bundle adjustment are turned off. } \\
% \multicolumn{15}{l}{$^\ddag$: Point + Line configuration.} \\
\hline
\end{tabular}
\caption{$\text{RPE}_\text{trans}$ (cm) comparisons on all sequences of the RGBD Scene v2 dataset.}
\label{tab:RPE_trans_comparison_supp_RGBD_Scene_v2}
\end{table}

\setlength\tabcolsep{1.2pt}
\begin{table}[!htbp]
\centering
\footnotesize
\begin{tabular}
{lcccccccccccccc}
Methods & {\rotatebox{45}{s01}} & {\rotatebox{45}{s02}} & {\rotatebox{45}{s03}} & {\rotatebox{45}{s04}} & {\rotatebox{45}{s05}} & {\rotatebox{45}{s06}} & {\rotatebox{45}{s07}} & {\rotatebox{45}{s08}} & {\rotatebox{45}{s09}} & {\rotatebox{45}{s10}} & {\rotatebox{45}{s11}} & {\rotatebox{45}{s12}} & {\rotatebox{45}{s13}} & {\rotatebox{45}{s14}} \\
\hline
ORB SLAM2$^{\star}$~\cite{mur2017orb} & - &-  &- &-  &1.54 &1.26 &0.96 &1.08 &- &- &- &-&-&- \\
KinectFusion$^{\star}$~\cite{izadi2011kinectfusion} &-&-&-&-&80.8&87.0&82.0&77.7&-&-&-&-&-&-\\
RGBD DVO$^{\star}$~\cite{cai2021direct} &-&-&-&-&4.21&5.96&4.83&4.44&-&-&-&-&-&-\\
Canny VO$^{\star}$~\cite{zhou2018canny} & - & -& -& -& -&-  & -& -& -& -&-&-&-&-  \\
RGBD DSO$^{\star}$$^{\lozenge}$~\cite{yuan2021rgb} &- & -&- &- &1.29 &8.95 &1.02 & 1.60&- &- &- &-&-&- \\
CVO~\cite{ghaffari2019continuous} & 1.76 & 1.17 & 1.11 & 2.40 & 1.62 & 1.54 & 1.37 & 1.55 & 1.13 & 1.18 & 1.46 & 1.42 & 1.29 & 1.47 \\
ACO~\cite{lin2019adaptive} & 1.52 & 1.41 & \underline{1.08} & \underline{1.47} & 1.59 & 1.55 & 1.48 & 2.11 & 1.16 & 1.32 & \underline{1.06} & 2.07 & 1.30 & 1.46 \\
Edge DVO~\cite{christensen2019edge} & 2.22&2.42 &2.40 & 2.26 & 1.03 & 0.99 & 1.00 &{1.15} &1.05 & 1.08 & 1.17 & 1.01 &$\times$ &$\times$ \\
CVO-SLAM$^\dag$~\cite{lin2023robust} & {\bf 0.09} & \underline{0.11} & {\bf 0.10} & {\bf 0.11} & {\bf 0.13} & {\bf 0.14} & \textbf{0.13} & \underline{0.15} & \underline{0.19} & \underline{0.12} & {\bf 0.12} & \textbf{0.07} & \textbf{0.09} & \textbf{0.09} \\
PLP-SLAM$^\dag$~\cite{shu2023structure} & 1.39 & 1.76 & 1.36 & 2.14 & 2.06 & 1.98 & 1.26 & 1.19 & 1.25 & 1.84 & 3.24 & 2.98 & 1.74 & 2.61 \\
Our Method & \underline{0.10}  & {\bf 0.09} & {\bf 0.10} & {\bf 0.11} & \underline{0.15}& \underline{0.15} & \underline{0.14} & {\bf 0.01} & {\bf 0.13} & \textbf{0.11} & {\bf 0.12} & \underline{0.11} & \underline{0.14} & \underline{0.16} \\
\hline
\multicolumn{15}{l}{\textbf{Boldfaced:} the best. \quad \underline{Underlined:} the second best. } \\
\multicolumn{15}{l}{$\star$: Values taken from their original papers, or from~\cite{yuan2021rgb}.} \\
\multicolumn{15}{l}{$\lozenge$: No depth refinement and occlusion removal modules.} \\
% \multicolumn{15}{l}{-: Not available from the original paper or the third parties.} \\
\multicolumn{15}{l}{$\times$: Estimations diverged. \quad -: Values unavailable.} \\
\multicolumn{15}{l}{$^\dag$: Loop closure and global bundle adjustment are turned off. } \\
% \multicolumn{15}{l}{$^\ddag$: Point + Line configuration.} \\
\hline
\end{tabular}
\caption{$\text{RPE}_\text{rot}$ (degree) comparisons on all sequences of the RGBD Scene v2 dataset.}
\label{tab:RPE_rots_comparison_supp_RGBD_Scene_v2}
\end{table}

\clearpage
% \newpage
% {
%     \small
%     \bibliographystyle{ieeenat_fullname}
%     \bibliography{main}
% }

\newpage
{
    \small
    \bibliographystyle{ieeenat_fullname}
    \bibliography{main}
}

% \newpage
% \input{sec/Nested_RANSAC_Old.tex}
% \input{sec/to_be_integrated}\input{CVPR_Authors/sec/To_Be_Reconsidered}

\end{document}